\documentclass[numbers]{article}

\usepackage[preprint]{neurips_2025}

\usepackage{natbib}

\usepackage[utf8]{inputenc} % allow utf-8 input
\usepackage[T1]{fontenc}    % use 8-bit T1 fonts
\usepackage{hyperref}       % hyperlinks
\usepackage{url}            % simple URL typesetting
\usepackage{booktabs}       % professional-quality tables
\usepackage{amsfonts}       % blackboard math symbols
\usepackage{nicefrac}       % compact symbols for 1/2, etc.
\usepackage{microtype}      % microtypography
\usepackage{xcolor}         % colors

\usepackage{makecell}
\usepackage{tabularx}
\usepackage{booktabs}
\usepackage{array}
\usepackage{subcaption}
\usepackage{multirow}
\usepackage{algorithm}
\usepackage{algorithmic}
\usepackage{amssymb}
\usepackage{tcolorbox}

\usepackage{pifont}
\usepackage{color}

\usepackage{amsmath}
\usepackage{makecell}
\usepackage{newfloat}
\usepackage{listings}
\usepackage{afterpage}
\usepackage{colortbl} 
\usepackage{wrapfig}

\usepackage{longtable}

\newcommand{\etc}{\textit{etc}.}

\newcommand{\eg}{\textit{e}.\textit{g}.}

\usepackage{mathtools}
\usepackage[linesnumbered,ruled,vlined,algo2e]{algorithm2e}
\SetCommentSty{mycommfont}

\usepackage{rotating}

\definecolor{c1}{HTML}{003371}
\definecolor{c2}{HTML}{057748}

\tcbuselibrary{breakable}
\NewDocumentCommand{\cdp}
{ mO{} }{\textcolor{blue}{\textsuperscript{\textit{Dong ping}}\textsf{\textbf{\small[#1]}}}}
\NewDocumentCommand{\yinuo}
{ mO{} }{\textcolor{blue}{\textsuperscript{\textit{Yi Nuo}}\textsf{\textbf{\small[#1]}}}}

\title{FIOVA: A Multi-Annotator Benchmark for Human-Aligned Video Captioning}

\author{\textbf{Shiyu Hu}$^{1*}$\hspace{9pt} 
\textbf{Xuchen Li}$^{2,3,4*}$ \hspace{9pt}
\textbf{Xuzhao Li}$^{5}$ \hspace{9pt}
\textbf{Jing Zhang}$^{2}$ \hspace{9pt}
\textbf{Yipei Wang}$^{6}$ \hspace{9pt}\\
\textbf{Xin Zhao}$^{7\dagger}$ \hspace{9pt}
\textbf{Kang Hao Cheong}$^{1,8\dagger}$ \hspace{9pt} \\
$*$ Equal Contributions \quad $\dagger$ Corresponding Authors \\
\textsuperscript{1}School of Physical and Mathematical Sciences, Nanyang Technological University\\
\textsuperscript{2}Institute of Automation, Chinese Academy of Sciences
\textsuperscript{3}Zhongguancun Academy\\
\textsuperscript{4}School of Artificial Intelligence, University of Chinese Academy of Sciences\\
\textsuperscript{5}School of Automation, Beijing Institute of Technology\\
\textsuperscript{6}Institute of Automation, Southeast University\\
\textsuperscript{7}School of Computer and Communication Engineering, University of Science and Technology Beijing \\
\textsuperscript{8}College of Computing and Data Science, Nanyang Technological University\\
\tt\small \{shiyu.hu, kanghao.cheong\}@ntu.edu.sg \quad s-lxc24@bjzgca.edu.cn \quad jing\_zhang@ia.ac.cn \\
\tt\small xuzhaoli2024@bit.edu.cn  \quad 230248984@seu.edu.cn \quad xinzhao@ustb.edu.cn\\
\textbf{Project Page}: \href{https://huuuuusy.github.io/fiova/}{https://huuuuusy.github.io/fiova/}
}

\begin{document}

\maketitle

\begin{abstract}
Despite rapid progress in large vision-language models (LVLMs), existing video caption benchmarks remain limited in evaluating their alignment with human understanding. Most rely on a single annotation per video and lexical similarity-based metrics, failing to capture the variability in human perception and the cognitive importance of events. These limitations hinder accurate diagnosis of model capabilities in producing coherent, complete, and human-aligned descriptions.
To address this, we introduce \textbf{FIOVA} (\textbf{F}ive-\textbf{I}n-\textbf{O}ne \textbf{V}ideo \textbf{A}nnotations), a human-centric benchmark tailored for evaluation. It comprises 3,002 real-world videos ($\sim$33.6s each), each annotated independently by five annotators. This design enables modeling of semantic diversity and inter-subjective agreement, offering a richer foundation for measuring human–machine alignment.
We further propose \textbf{FIOVA-DQ}, an event-level evaluation metric that incorporates cognitive weights derived from annotator consensus, providing fine-grained assessment of event relevance and semantic coverage. Leveraging FIOVA, we conduct a comprehensive evaluation of nine representative LVLMs and introduce a complexity-aware analysis framework based on inter-annotator variation (CV). This reveals consistency gaps across difficulty levels and identifies structural issues such as event under-description and template convergence.
Our results highlight FIOVA’s diagnostic value for understanding LVLM behavior under varying complexity, setting a new standard for cognitively aligned evaluation in long-video captioning. The benchmark, annotations, metric, and model outputs are publicly released to support future evaluation-driven research in video understanding.
\end{abstract}

\begin{figure*}[t!]
% \vspace{-12pt}
\centering
  \includegraphics[width=\textwidth]{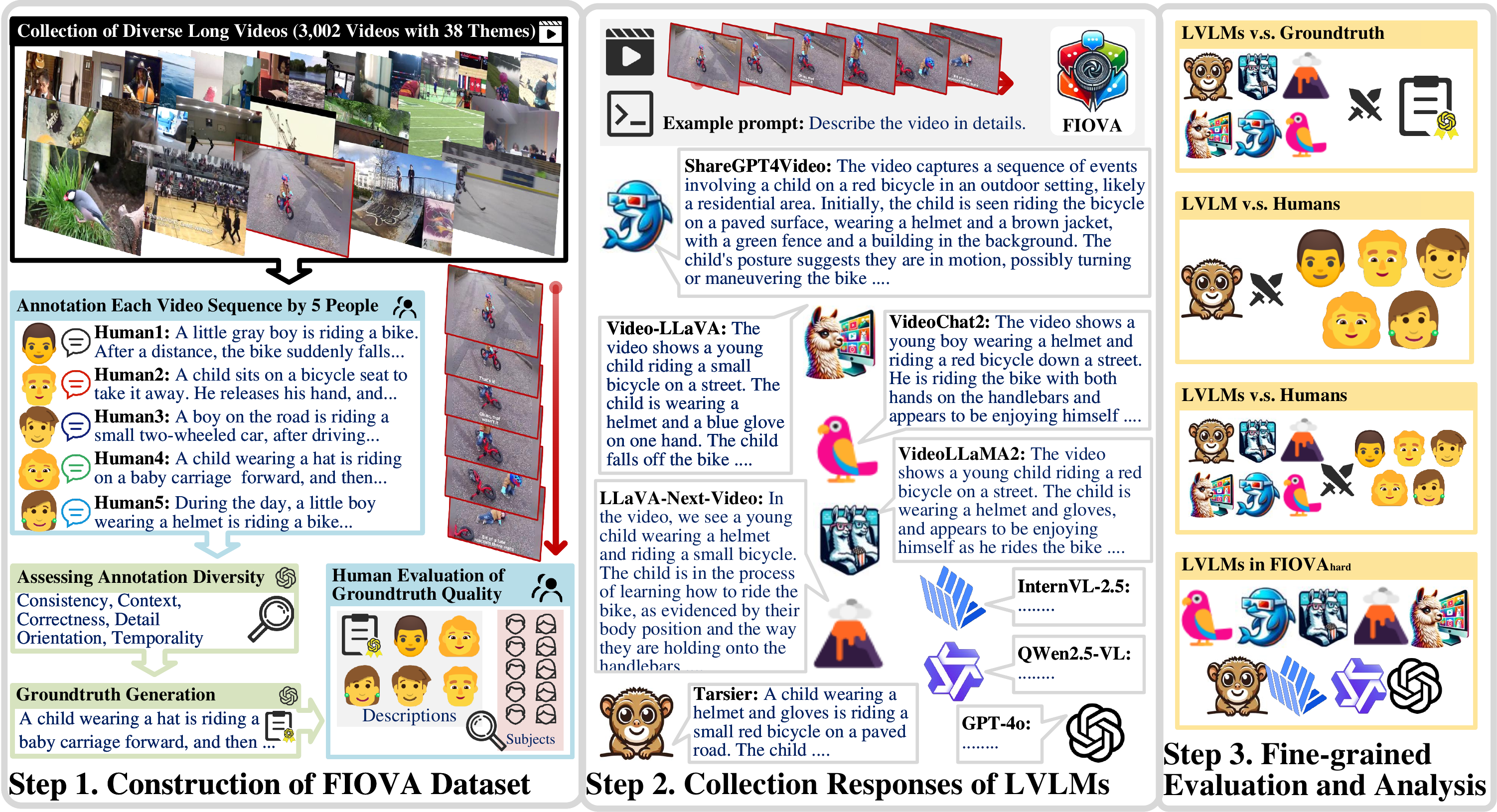}
  \vspace{-6pt}
  \caption{Overview of FIOVA. The workflow comprises three steps: (i) dataset construction (Sec.~\ref{step1}), (ii) LVLM response collection (Sec.~\ref{step2}), and (iii) fine-grained evaluation and analysis (Sec.~\ref{step3}). Together, these steps form FIOVA for systematically comparing human and LVLM video understanding.}
  \label{fig:overview}
  % \vspace{-20pt}
\end{figure*}

\section{Introduction}

Large language models (LLMs) have shown strong performance in natural language tasks such as text generation~\cite{dtllm,li2024visual,mahapatra2024impact} and question answering~\cite{toolqa,saito2024unsupervised}. Building upon these advances, large vision-language models (LVLMs), including GPT-4o~\cite{gpt4} and LLaVA~\cite{llava}, integrate visual and textual modalities to support complex applications like text-to-video generation~\cite{huang2024free} and video captioning~\cite{huang2024vtimellm}. 
However, existing evaluation metrics—typically based on lexical overlap or embedding similarity—fail to capture the cognitive alignment necessary for human-level video understanding~\cite{sotverse,hu2024multi,hu2023global}. This limitation raises a fundamental research question: \textit{Can LVLMs achieve a comprehensive and coherent understanding of videos comparable to humans?}

\begin{wraptable}{r}{0.6\linewidth}
\vspace{-10pt}
\caption{Comparison of FIOVA and existing video caption datasets. FIOVA uniquely offers \textbf{multi-annotator (5×)} captions for each video. We split datasets into two groups: automatic caption by ASR or LVLM~\cite{howto100m,acav,merlot,hdvila,panda70m}, and manual caption~\cite{msvd,msrvtt,youcook2,activitynet,didemo,lsmdc,vatex,tarsier,chai2024auroracap}.}
\label{tab:dataset}
\centering
\small
\fontsize{8pt}{9.5pt}\selectfont
\setlength{\tabcolsep}{2pt}
\begin{tabular}{llcccc}
\toprule
\textbf{Dataset} & \textbf{Type} & \textbf{Domain} & \textbf{\#Videos} & 
\begin{tabular}[c]{@{}c@{}}\textbf{Avg}\\\textbf{Video Len}\end{tabular} & 
\begin{tabular}[c]{@{}c@{}}\textbf{Avg}\\\textbf{Text Len}\end{tabular} \\
\midrule
\multicolumn{6}{l}{\textit{Automatic Captions (ASR or LVLM)}} \\
HowTo100M       & ASR      & Open     & 136M   & 3.6s      & 4.0 w \\
ACAV            & ASR      & Open     & 100M   & 10.0s     & --    \\
YT-Temporal-180M& ASR      & Open     & 180M   & --        & --    \\
HD-VILA-100M    & ASR      & Open     & 103M   & 13.4s     & 32.5 w \\
Panda-70M       & LVLM     & Open     & 70.8M  & 8.5s      & 13.2 w \\
\midrule
\multicolumn{6}{l}{\textit{Manual Captions (Single Annotator)}} \\
MSVD            & 1-annot. & Open     & 1.97K  & 9.7s      & 8.7 w \\
LSMDC           & 1-annot. & Movie    & 118K   & 4.8s      & 7.0 w \\
MSR-VTT         & 1-annot. & Open     & 10K    & 15.0s     & 9.3 w \\
DiDeMo          & 1-annot. & Flickr   & 27K    & 6.9s      & 8.0 w \\
ActivityNet     & 1-annot. & Action   & 100K   & 36.0s     & 13.5 w \\
YouCook2        & 1-annot. & Cooking  & 14K    & 19.6s     & 8.8 w \\
VATEX           & 1-annot. & Open     & 41K    & $\sim$10s & 15.2 w \\
DREAM-1K        & 1-annot. & Open     & 1K     & 8.9s      & 59.3 w \\
AuroraCap (VDC) & 1-annot. & Open     & 2.3K   & $\sim$60s & 72.4 w \\
\midrule
FIOVA (Ours) & 5-annot. & Open & 3K & 33.6s & 63.3 w \\
\bottomrule
\end{tabular}
\vspace{2pt}
\caption*{\footnotesize
\textbf{Note:} “ASR” denotes automatic speech recognition; “LVLM” indicates captions generated by LVLMs; “1-annot.”/“5-annot.” indicates the number of human annotators per video. “w” = words.
}
\vspace{-15pt}
\end{wraptable}

Video captioning tasks~\cite{aafaq2019video,ramanishka2016multimodal} offer a suitable testbed for this question, as they require integrated spatial–temporal reasoning and nuanced event interpretation. Compared to structured tasks such as object recognition~\cite{logothetis1996visual} or visual question answering~\cite{antol2015vqa}, video captioning demands holistic perception and narrative coherence—posing substantial challenges for both humans and machines (see App.~\ref{sec:related-work}).

Existing video captioning benchmarks~\cite{howto100m,acav,msvd,activitynet,msrvtt,panda70m,youcook2} suffer from notable limitations: they primarily include short clips ($\sim$10 seconds), provide brief annotations ($\sim$15 words on average), and rely on single-annotator inputs (see Tab.~\ref{tab:dataset}). These constraints hinder their effectiveness in evaluating LVLM performance under realistic and cognitively diverse conditions.

To address these gaps, we present \textbf{FIOVA} (\textbf{F}ive \textbf{I}n \textbf{O}ne \textbf{V}ideo \textbf{A}nnotations), an evaluation-oriented benchmark designed to measure the semantic and perceptual alignment between LVLM-generated captions and human understanding. As shown in Fig.~\ref{fig:overview}, FIOVA offers three key contributions:
(1) \textbf{Comprehensive Dataset Construction}: FIOVA comprises 3,002 real-world videos (average duration 33.6s) spanning diverse scenarios with complex spatiotemporal dynamics. Each video is annotated independently by five human annotators, yielding captions 4–15 times longer than those in most existing datasets. This multi-perspective design supports robust modeling of human perceptual diversity and establishes a solid foundation for evaluating human-level comprehension (Sec.~\ref{step1}).
(2) \textbf{Rigorous LVLM Evaluation}: We systematically evaluated nine representative LVLMs—VideoLLaMA2~\cite{videollama2}, Video-LLaVA~\cite{videollava}, LLaVA-NEXT-Video~\cite{llavanextvideo}, Tarsier~\cite{tarsier}, VideoChat2~\cite{videochat2}, ShareGPT4Video~\cite{sharegpt4video}, InternVL-2.5~\cite{InternVL}, Qwen2.5-VL~\cite{Qwen}, and GPT-4o~\cite{gpt4}—across varying input complexities, analyzing their outputs to reveal model strengths, limitations, and sensitivity (Sec.~\ref{step2}).
(3) \textbf{Fine-grained Human–Machine Comparative Analysis}: We conducted in-depth comparisons on semantic coherence, temporal ordering, and contextual consistency between LVLM outputs and human annotations. In addition, we propose \textbf{FIOVA-DQ}, a cognitively weighted event-level metric derived from annotator consensus, which enables more nuanced and human-aligned evaluation of semantic relevance and caption fluency (Sec.~\ref{step3}).

Through the integration of diverse human annotations, cognitively grounded evaluation, and systematic analysis, FIOVA offers a robust framework for diagnosing LVLM behavior and benchmarking next-generation models aligned with human perception.

\section{Construction of FIOVA Dataset}
\label{step1}

Fig.~\ref{fig:overview} provides an overview of our workflow. This section focuses on the first stage: constructing the FIOVA dataset. We curated 3,002 video–annotation pairs, denoted as \( D = \{(V_1, C_{1}), \ldots, (V_n, C_{n})\} \), where $C_{i} = \{c_{i1}, c_{i2}, c_{i3}, c_{i4}, c_{i5}\}$ comprises five independent human-written descriptions for video $V_i$ (see Sec.~\ref{subsec:video-collection-and-annotation}). These annotations were further synthesized into a unified groundtruth $g_i$, serving as a reliable reference for human-level video understanding (see Sec.~\ref{subsec:gt-generation}). In total, FIOVA consists of 3,002 triplets $(V_i, C_{i}, g_i)$, corresponding to 3,002 videos, 15,010 human annotations, and 3,002 consolidated groundtruths.

\begin{figure*}[t!]
\centering
\includegraphics[width=\textwidth]{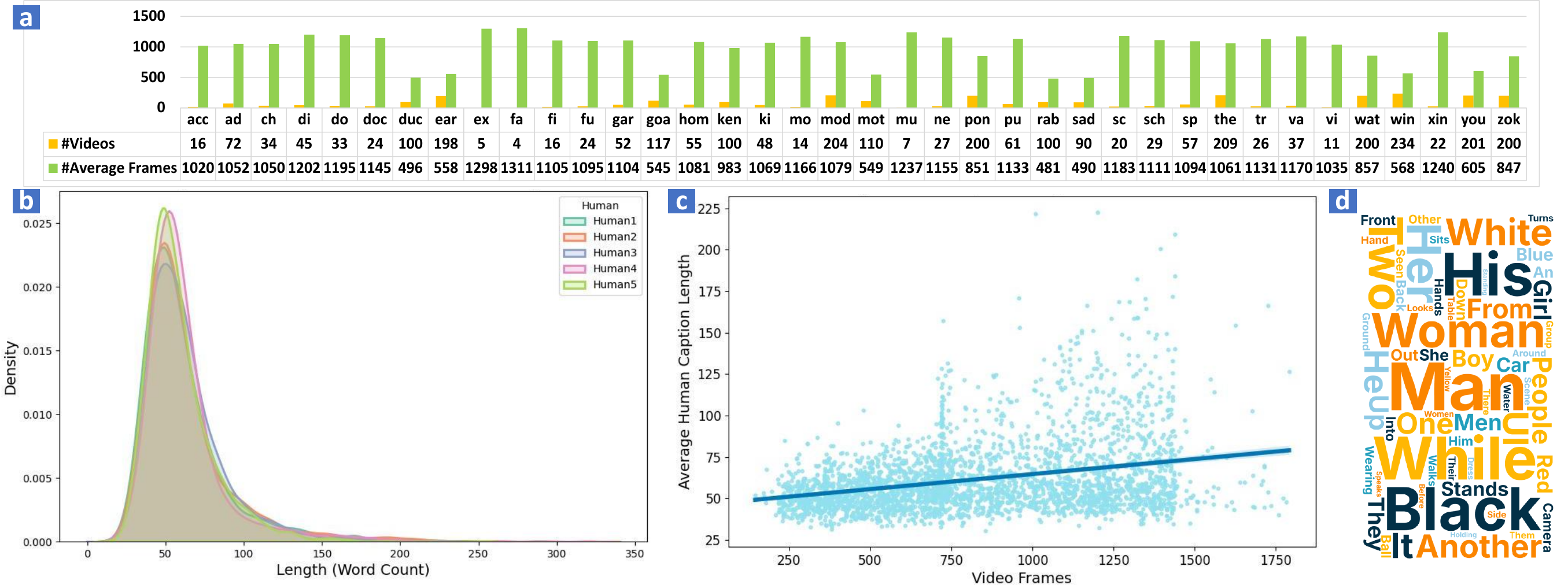}
\caption{Statistical overview of FIOVA:
(a) Average frame count per video across thematic categories (see Tab.~\ref{tab:fiova-theme});
(b) Distribution of annotation lengths among five annotators;
(c) Correlation between annotation length and video duration;
(d) Word cloud based on GPT-synthesized groundtruth captions.}
\label{fig:distribution}
% \vspace{-12pt}
\end{figure*}

\subsection{Video Collection and Annotation}
\label{subsec:video-collection-and-annotation}

FIOVA is specifically curated to evaluate the video comprehension capabilities of LVLMs. Spanning 38 diverse themes (see App.~\ref{subsec:fiova-theme}), the dataset covers a broad range of real-world scenarios, enhancing its applicability for robust model evaluation.
Each video was independently annotated by five individuals based solely on visual content, with audio and subtitles excluded (except for naturally embedded text). To ensure quality and consistency, annotators followed standardized guidelines (see App.~\ref{subsec:annotation-rules}) that emphasize factual accuracy, neutrality, and temporal coherence. Descriptions focus on observable visual elements—locations, actions, and objects—presented in chronological order using clear and objective language.

FIOVA introduces unique challenges compared to existing benchmarks, making it a more rigorous testbed for video understanding. As illustrated in Fig.~\ref{fig:example}, the videos exhibit diverse resolutions, aspect ratios, and frequent scene transitions, demanding contextual adaptability from models. Additional complexities such as lens distortions (\eg, fisheye effects) further challenge spatial reasoning.

Each video is accompanied by five distinct English descriptions composed as coherent paragraphs. The number of sentences varies with video complexity, offering rich and structured event narratives. With an average duration of 33.6 seconds, FIOVA captures intricate temporal interactions and long-range dependencies—making it particularly suitable for evaluating deep video comprehension.
By combining thematic diversity, multi-annotator richness, and rigorous annotation standards, FIOVA supports nuanced and comprehensive evaluation. As shown in Tab.~\ref{tab:dataset}, it surpasses existing datasets by providing multiple independent references per video. Fig.~\ref{fig:distribution} further illustrates key dataset statistics, highlighting its value for assessing LVLM performance across varying complexities.

\begin{wrapfigure}{r}{0.65\columnwidth}
% \vspace{-30pt}
\centering
\includegraphics[width=0.65\columnwidth]{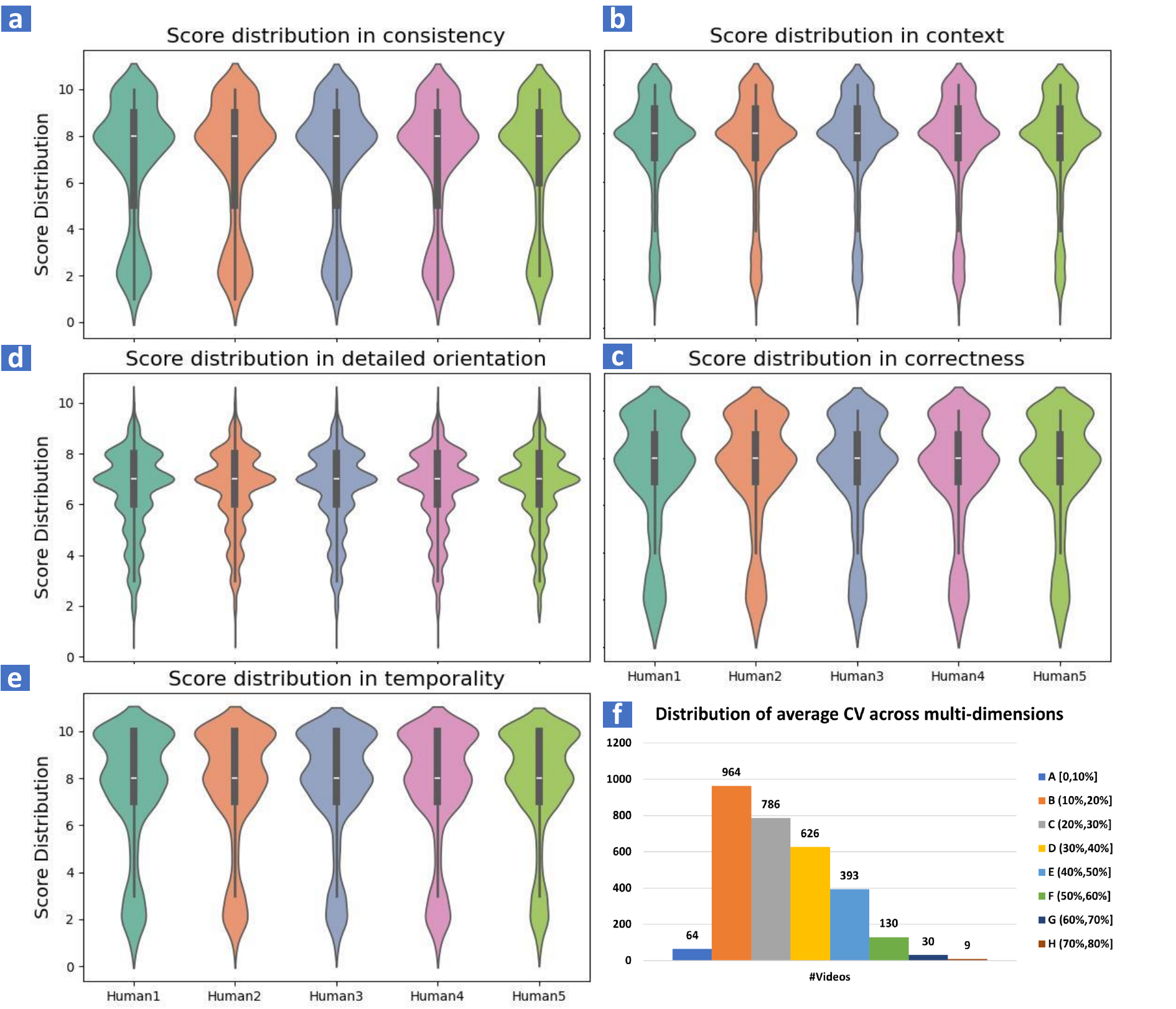}

\caption{Distribution of human annotation scores across five dimensions (a-e) and variation among annotators (f), measured using the coefficient of variation (CV).}
\label{fig:6D-CV}
% \vspace{-10pt}
\end{wrapfigure}

\subsection{Assessing Annotation Diversity}
\label{subsec:accessing-annotation-diversity}

To construct a reliable human baseline, each video in FIOVA is annotated independently by five individuals, capturing diverse human perspectives. Before synthesizing these annotations into groundtruth (see Sec.~\ref{subsec:gt-generation}), we assess their consistency and variability to examine inter-annotator agreement in video descriptions.
Following prior studies such as Video-ChatGPT~\cite{maaz2023video} and Tarsier~\cite{tarsier}, we adopt a GPT-based evaluation framework to score annotations across five dimensions inspired by VideoLLaMA2~\cite{videollama2}:
(1) \textbf{Consistency:} logical coherence and alignment with video content;
(2) \textbf{Context:} accurate modeling of scene transitions and event dependencies;
(3) \textbf{Correctness:} factual accuracy without misleading content;
(4) \textbf{Detail Orientation:} coverage of key people, objects, and actions;
(5) \textbf{Temporality:} adherence to chronological order without omission or overgeneralization.
Each annotation receives a score from 1 to 10 for each dimension (see App.~\ref{subsubsec:prompt-human-annotation}). Unlike traditional methods that compare against a fixed reference, our approach directly quantifies inter-annotator differences across these criteria.
Fig.~\ref{fig:6D-CV} (a–e) shows the score distribution across dimensions. While human annotations exhibit high overall consistency, variation in \textit{Detail Orientation} suggests occasional omission of finer elements.

To further quantify diversity, we compute the coefficient of variation (CV), a normalized dispersion metric based on standard deviation and mean. Higher CV values indicate greater disagreement and subjectivity. In addition to the five semantic dimensions, we include \textit{Annotation Length} as a sixth measure (see Fig.~\ref{fig:distribution} (b)). Averaging CVs across all six metrics (see Alg.~\ref{alg:cv_human}), we group videos into eight subsets (Groups A–H) based on inter-annotator agreement (see Fig.~\ref{fig:6D-CV} (f) and App.~\ref{subsec:8groups}). Group A videos show strong agreement, while Group H exhibits the highest variance.

\textbf{FIOVA$_{\text{hard}}$ Subset.}
Based on this grouping, we define \textbf{FIOVA$_{\text{hard}}$} as the subset containing all videos in Groups F, G, and H—those with the highest annotation divergence and semantic complexity. This subset provides a challenging evaluation setting for probing model robustness under cognitively demanding conditions. In later sections, we use FIOVA$_{\text{hard}}$ for targeted stress-testing, including capability analysis and frame-length ablation.
Such “hard” subset paradigms are well established in benchmark design. For example, BIG-Bench Hard (BBH)\cite{srivastava2022beyond} focuses on tasks where models underperform relative to humans. GLGE\cite{liu2020glge} and DL-HARD~\cite{mackie2021deep} similarly define difficult subsets to enable diagnostic evaluation. Inspired by these precedents, FIOVA$_{\text{hard}}$ offers a principled setting for analyzing LVLM performance under high semantic difficulty.

\begin{figure*}[t!]
\centering
  \includegraphics[width=\textwidth]{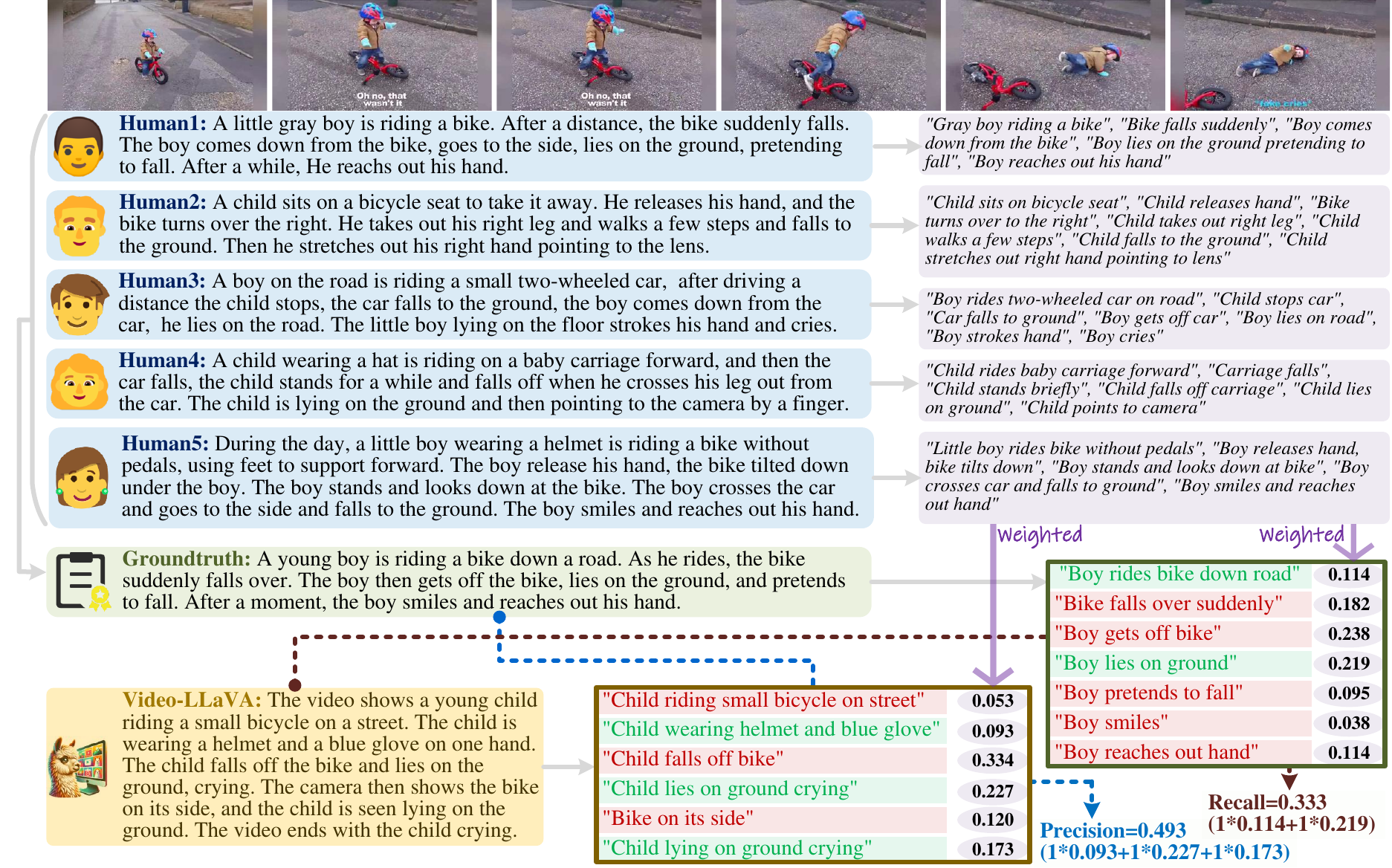}
  % \vspace{-6pt}
  \caption{An example of FIOVA and the calculation process of FIOVA-DQ (see Fig.~\ref{fig:acc15}).}
  \label{fig:acc15-eg}
  % \vspace{-12pt}
\end{figure*}

\subsection{Unified Groundtruth Generation}
\label{subsec:gt-generation}

To establish a robust reference for evaluating video understanding, we synthesize the five independent human annotations for each video into a unified groundtruth using GPT (see App.~\ref{subsubsec:prompt-groundtruth-generation}). This procedure integrates multiple perspectives into a coherent and comprehensive narrative that captures the key content of each video (see Fig.~\ref{fig:acc15-eg}).

GPT-based synthesis offers a structured approach for aggregating diverse viewpoints while minimizing subjective bias and ensuring semantic coverage. Each resulting groundtruth reflects a balanced understanding of the video—preserving detail orientation, contextual relevance, and temporal order.
Unlike individual annotations that may emphasize different aspects, the synthesized groundtruth provides a more consistent and neutral reference. By consolidating complementary human perspectives, it serves as a reliable baseline for model evaluation under cognitively aligned criteria.

\subsection{Human Evaluation of Groundtruth Quality}
\label{subsec:gt-quality}

\begin{wraptable}{r}{0.25\textwidth}
  \centering
  \vspace{-40pt}
  \caption{Human evaluation (ranking; lower is better).}
  
  \label{tab:human_eval_expanded}
  \setlength{\tabcolsep}{2pt}
  \small
  \begin{tabular}{lcc}
    \toprule
    & \textbf{Mode A} & \textbf{Mode B} \\
     % & \textbf{(Text-only)} & \textbf{(Video+Text)} \\
    \midrule
    Human1 & 3.65 & 3.62 \\
    Human2 & 4.33 & 4.15 \\
    Human3 & 4.25 & 4.19 \\
    Human4 & 3.90 & 3.91 \\
    Human5 & 2.81 & 2.95 \\
    \midrule
    GPT-GT & 2.06 & 2.20 \\
    \bottomrule
  \end{tabular}
  \vspace{-10pt}
\end{wraptable}

To assess the reliability of GPT-synthesized groundtruths in FIOVA, we conducted a human evaluation comparing them with individual human annotations in terms of informativeness, consistency, and semantic alignment (see App.~\ref{sec:human-gt-evaluation}).
Ten computer vision experts evaluated 20 videos spanning diverse topics, annotation complexity, and semantic structures. Each video was accompanied by six captions—five from individual annotators and one GPT-generated via multi-annotation fusion. Evaluations were performed under two settings: (1) \textbf{Text-Only} (Mode A), without video context; and (2) \textbf{Video+Text} (Mode B), with video provided for semantic grounding. Captions were ranked from 1 (best) to 6 (worst).

As shown in Tab.~\ref{tab:human_eval_expanded}, GPT-summarized captions consistently received higher average rankings—2.06 in Mode A and 2.20 in Mode B—compared to human-written captions (ranging from 2.81–4.33 and 2.95–4.19, respectively). Evaluators noted three main strengths of the GPT outputs: (i) clear and coherent structure, (ii) concise yet comprehensive event coverage, and (iii) strong semantic alignment with video content.
These results demonstrate that GPT-based synthesis enhances caption stability and quality, offering a reliable and human-aligned reference for evaluation.

\section{LVLMs Response Collection}
\label{step2}

As illustrated in Step 2 of Fig.~\ref{fig:overview}, this section describes how LVLM responses were collected to construct the benchmark. Each video \( V_i \) in the dataset was processed by multiple LVLMs, resulting in a benchmark set denoted as \( B = \{(V_i, C_{i}, R_i) \ | \ (V_i, C_{i}) \in D\} \), where $R_{i} = \{r_{i1}, r_{i2}, \ldots, r_{in}\}$ represents model-generated captions for video $V_i$.

\begin{wrapfigure}{r}{0.5\columnwidth}
\vspace{-10pt}
\centering
\includegraphics[width=0.5\columnwidth]{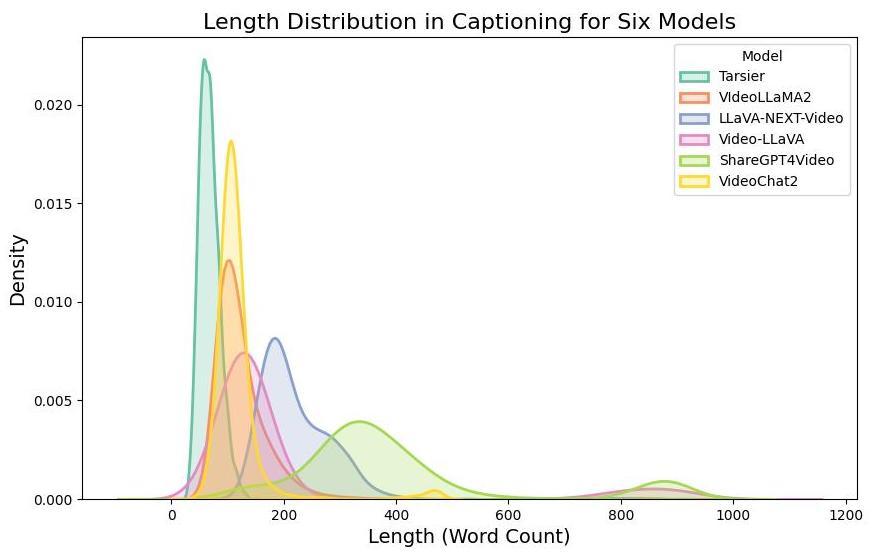}
\vspace{-5pt}
\caption{The distribution of response length.}
\label{fig:lvlm-length}
% \vspace{-10pt}
\end{wrapfigure}

\subsection{Baseline Models Selection}
\label{subsec:baseline-models}

We selected six state-of-the-art open-source LVLMs for this study: VideoLLaMA2~\cite{videollama2}, Video-LLaVA~\cite{videollava}, LLaVA-NEXT-Video~\cite{llavanextvideo}, Tarsier~\cite{tarsier}, VideoChat2~\cite{videochat2}, and ShareGPT4Video~\cite{sharegpt4video} (see App.~\ref{subsec:related-work-lvlms} for model details).
Each model was prompted with a video description task, generating a total of 18,012 responses (see App.~\ref{subsec:lvlms-prompt}). Fig.~\ref{fig:lvlm-length} shows the distribution of response lengths, revealing inter-model variability in output verbosity.
Model configurations were as follows:
VideoLLaMA2 used default settings with a temperature of 0.2 and a maximum token limit of 1,024. VideoChat2 and ShareGPT4Video were configured with default settings, a temperature of 1.0, top\_p of 0.9, and a maximum token limit of 1,024. Video-LLaVA had a temperature of 0.1 and the same token limit. Tarsier and LLaVA-NEXT-Video were set with a temperature of 0, top\_p of 1, and a maximum token limit of 1,024. 
All models were run on 8-frame inputs using four RTX 3090 GPUs.
Additionally, we included three high-capacity LVLMs—InternVL-2.5~\cite{InternVL}, Qwen2.5-VL~\cite{Qwen}, and GPT-4o~\cite{gpt4}—for evaluation on constrained subsets (see Sec.~\ref{subsec:hard-sota}).

\subsection{Event Generation}
\label{subsec:event-generation}

Video descriptions generated by LVLMs (Sec.~\ref{subsec:baseline-models}) are typically evaluated using text-based metrics. However, recent work such as AutoDQ~\cite{tarsier} adopts event-based metrics by extracting structured events from both model outputs and reference captions, enabling more detailed assessments. While AutoDQ aligns model outputs with human annotations effectively, it treats all events equally and ignores their varying cognitive salience.
Thus, we propose \textbf{FIOVA-DQ}, an enhanced event-based metric that integrates human cognitive weighting into event alignment (see Sec.~\ref{subsec:evaluation-methods}). Unlike conventional approaches, FIOVA-DQ accounts for both the correctness and perceived importance of events, offering a more human-aligned evaluation framework.
For consistency and generalizability, we apply GPT-based event parsing to both the groundtruth annotations $g_i$ and LVLM outputs $r_{ij}$,  producing event sets $E^{gt}_{i}$ and $E^{r}_{ij}$ (see App.~\ref{subsubsec:event-extraction}). Human-derived importance scores are then assigned to events in $E^{gt}_{i}$, normalized to sum to one (see Fig.~\ref{fig:acc15-eg}). 
By incorporating these weights into event matching, FIOVA-DQ produces scores that reflect not only alignment accuracy but also human-perceived event significance. This cognitively grounded approach bridges the gap between surface-level matching and human understanding, enabling finer-grained and more interpretable model evaluation.

\section{Fine-grained Evaluation and Analysis}
\label{step3}

As shown in Step 3 of Fig.~\ref{fig:overview}, this section presents a multi-perspective evaluation of LVLMs on FIOVA. We assess model-generated captions against two human references: (i) the GPT-synthesized groundtruth (from five annotations), and (ii) the original individual annotations. This allows for systematic analysis of model–human alignment across linguistic, structural, and cognitive dimensions.

\begin{table*}[t!]
\centering
\small
\caption{Comparison of LVLMs across different evaluation metrics.
The background color represents the performance of the metric. The darker the green, the better the performance.}
\label{tab:metric-result}
\setlength{\tabcolsep}{2pt}
\begin{tabular}{lccc c ccc c ccc}
\toprule
 & \multicolumn{3}{c}{\textbf{Traditional Metrics}} & & \multicolumn{3}{c}{\textbf{AutoCQ-based Metrics}} & & \multicolumn{3}{c}{\textbf{FIOVA-DQ-based Metrics}} \\
\cmidrule{2-4} \cmidrule{6-8} \cmidrule{10-12}
 & BLEU & METEOR & GLEU & & F1 & Recall & Precision & & F1 & Recall & Precision \\
\midrule
\textbf{Tarsier} & \cellcolor[HTML]{63BE7B}0.043 & \cellcolor[HTML]{8ACE9D}0.265 & \cellcolor[HTML]{63BE7B}0.119 &  & \cellcolor[HTML]{63BE7B}0.351 & \cellcolor[HTML]{63BE7B}0.283 & \cellcolor[HTML]{FCFCFF}0.628 &  & \cellcolor[HTML]{63BE7B}0.320 & \cellcolor[HTML]{63BE7B}0.584 & \cellcolor[HTML]{FCFCFF}0.584 \\
\textbf{VIdeoLLaMA2} & \cellcolor[HTML]{A0D7B0}0.030 & \cellcolor[HTML]{83CB97}0.268 & \cellcolor[HTML]{9CD5AC}0.088 &  & \cellcolor[HTML]{9CD6AD}0.325 & \cellcolor[HTML]{ABDBB9}0.245 & \cellcolor[HTML]{AFDDBD}0.680 &  & \cellcolor[HTML]{8DCFA0}0.304 & \cellcolor[HTML]{EAF5EF}0.250 & \cellcolor[HTML]{B4DFC1}0.645 \\
\textbf{LLaVA-NEXT-Video} & \cellcolor[HTML]{D0EAD9}0.020 & \cellcolor[HTML]{7FCA93}0.270 & \cellcolor[HTML]{CFEAD8}0.060 &  & \cellcolor[HTML]{D1EBDA}0.301 & \cellcolor[HTML]{D7EDDF}0.221 & \cellcolor[HTML]{B8E1C4}0.674 &  & \cellcolor[HTML]{BEE3C9}0.286 & \cellcolor[HTML]{F2F8F7}0.229 & \cellcolor[HTML]{B6E0C2}0.644 \\
\textbf{Video-LLaVA} & \cellcolor[HTML]{AFDDBD}0.027 & \cellcolor[HTML]{9ED6AE}0.257 & \cellcolor[HTML]{B0DEBE}0.077 &  & \cellcolor[HTML]{F4F9F8}0.285 & \cellcolor[HTML]{EFF7F4}0.208 & \cellcolor[HTML]{84CC98}0.709 &  & \cellcolor[HTML]{EDF6F2}0.269 & \cellcolor[HTML]{F7FAFB}0.216 & \cellcolor[HTML]{8BCE9D}0.680 \\
\textbf{ShareGPT4Video} & \cellcolor[HTML]{FCFCFF}0.010 & \cellcolor[HTML]{FCFCFF}0.218 & \cellcolor[HTML]{FCFCFF}0.034 &  & \cellcolor[HTML]{FCFCFF}0.281 & \cellcolor[HTML]{FCFCFF}0.201 & \cellcolor[HTML]{63BE7B}0.731 &  & \cellcolor[HTML]{FCFCFF}0.263 & \cellcolor[HTML]{FCFCFF}0.203 & \cellcolor[HTML]{63BE7B}0.714 \\
\textbf{VideoChat2} & \cellcolor[HTML]{82CB95}0.037 & \cellcolor[HTML]{63BE7B}0.281 & \cellcolor[HTML]{89CE9C}0.098 &  & \cellcolor[HTML]{C0E4CC}0.309 & \cellcolor[HTML]{BAE2C6}0.237 & \cellcolor[HTML]{D3ECDB}0.656 &  & \cellcolor[HTML]{BDE3C9}0.287 & \cellcolor[HTML]{ECF6F1}0.243 & \cellcolor[HTML]{D1EBDA}0.621 \\ 
\bottomrule
\end{tabular}
% \vspace{-8pt}
\end{table*}

\subsection{Evaluation Framework}
\label{subsec:evaluation-methods}

We adopt a three-tier evaluation framework to comprehensively assess alignment between LVLMs and humans:
(i) \textbf{Traditional metrics} (BLEU, GLEU, METEOR),
(ii) \textbf{AutoDQ}, which evaluates event-level coverage, and
(iii) \textbf{FIOVA-DQ}, our cognitively weighted extension.
Traditional metrics capture surface-level lexical similarity but fail to reflect semantic completeness or temporal structure, especially in complex videos. Recent LLM-based methods (\eg, ChatGPT ratings~\cite{maaz2023video, gpt4}) offer more flexible assessments but suffer from limited interpretability and reproducibility (see App.~\ref{subsuec:related-work-evaluation}).
To address these limitations, AutoDQ~\cite{tarsier} extracts events from both the groundtruth ($E^{gt}_{i}$, synthesized from five human annotations) and model output ($E^{r}_{ij}$), and computes recall (event coverage) and precision (event accuracy). FIOVA-DQ extends this by incorporating human-derived cognitive weights over $E^{gt}_{i}$, prioritizing semantically salient events in the computation of weighted F1, Precision, and Recall.
This unified framework supports two complementary modes:
(1) \textbf{Overall evaluation}—measuring how closely each model output aligns with human references;
(2) \textbf{Batch evaluation}—comparing model behaviors across samples to identify systematic patterns in human-likeness.

\begin{wraptable}{r}{0.35\textwidth}
\centering
\vspace{-20pt}
\caption{Spearman correlations between average human ranking and automatic metrics.}

\label{tab:spearman-summary}
\setlength{\tabcolsep}{2pt}
\small
\begin{tabular}{lccc}
\toprule
\textbf{Metric} & \textbf{F1} & \textbf{Precision} & \textbf{Recall} \\
\midrule
FIOVA-DQ & 0.579 & 0.239 & 0.359 \\
AutoDQ   & 0.125 & 0.268 & 0.048 \\
\bottomrule
\end{tabular}
\vspace{-10pt}
\end{wraptable}

\subsection{Overall Evaluation for Baseline LVLMs}

We systematically evaluate baseline LVLMs by comparing model-generated captions against the synthesized groundtruth using three metric families: traditional linguistic metrics, AutoDQ, and FIOVA-DQ.

\textbf{Traditional metrics.}
As shown in Tab.~\ref{tab:metric-result}, Tarsier achieves the highest BLEU and GLEU scores, indicating strong lexical overlap. However, its lower METEOR score suggests limited morphological flexibility. In contrast, ShareGPT4Video—despite strong performance elsewhere—performs poorly here due to repetitive outputs induced by its sliding-window strategy, which traditional metrics penalize.
These results underscore the need to balance lexical similarity with expressive diversity.

\textbf{AutoDQ-based metrics.}
Tarsier again leads in Recall and F1, demonstrating strong coverage but low Precision due to overgeneration. ShareGPT4Video, conversely, yields the highest Precision but lowest Recall, indicating a conservative style that misses contextual events. Other models fall in between, revealing different trade-offs between coverage and selectivity.

\textbf{FIOVA-DQ-based metrics.}
FIOVA-DQ incorporates human-derived weights, prioritizing cognitively salient events. Under this metric, Tarsier retains top F1 and Recall, but its Precision declines further, highlighting misalignment on key content. Other models show slight shifts, confirming that cognitive weighting meaningfully alters evaluations.

\textbf{Human alignment validation.}
We validated metric reliability via a human judgment experiment (see App.~\ref{app:human-validation}). Spearman correlations (Tab.~\ref{tab:spearman-summary}) show that FIOVA-DQ aligns better with human preferences (F1: 0.579) than AutoDQ, supporting its interpretability and trustworthiness.

\begin{wrapfigure}{r}{0.5\columnwidth}
% \vspace{-10pt}
\centering
\includegraphics[width=0.5\columnwidth]{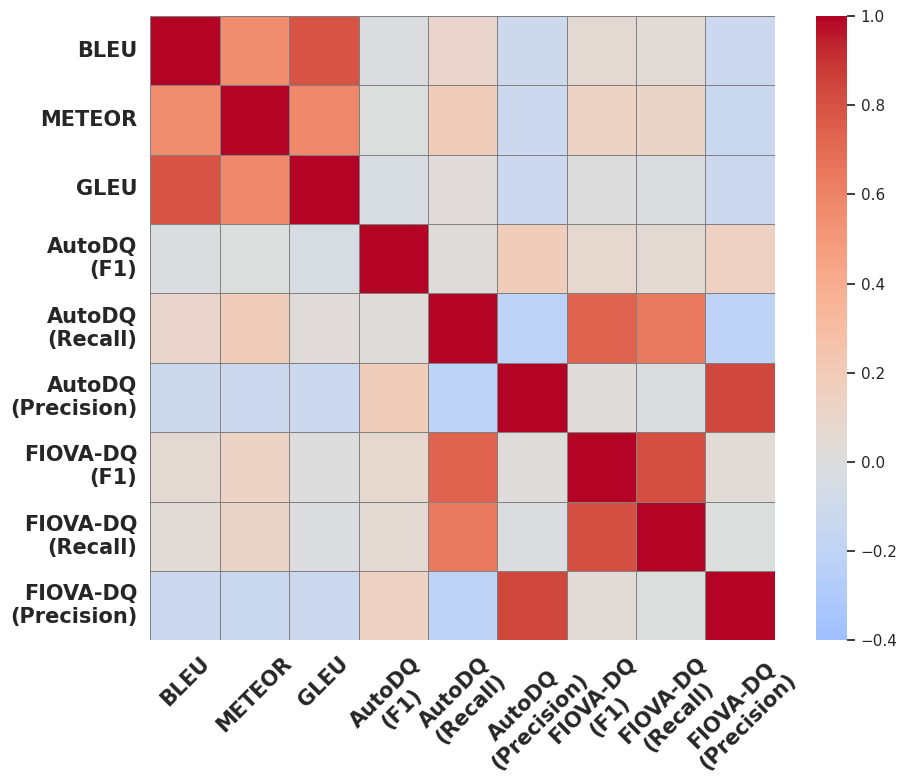}
\vspace{-5pt}
\caption{Spearman correlation matrix of evaluation metrics.}
\label{fig:spearman-heatmap}
% \vspace{-10pt}
\end{wrapfigure}

\textbf{Metric correlation analysis.}
To explore metric consistency, we computed Spearman correlations across nine metrics (BLEU, METEOR, GLEU, AutoDQ-F1/Recall/Precision, and FIOVA-DQ-F1/Recall/Precision) for six baseline models. As shown in Fig.\ref{fig:spearman-heatmap}, traditional metrics poorly correlate with content-aware scores, while FIOVA-DQ shows strong internal consistency and aligns better with semantic evaluation. Details are provided in App.~\ref{app:metric-correlation-analysis}.

\begin{figure*}[t!]
% \vspace{-10pt}
\centering
\includegraphics[width=\columnwidth]{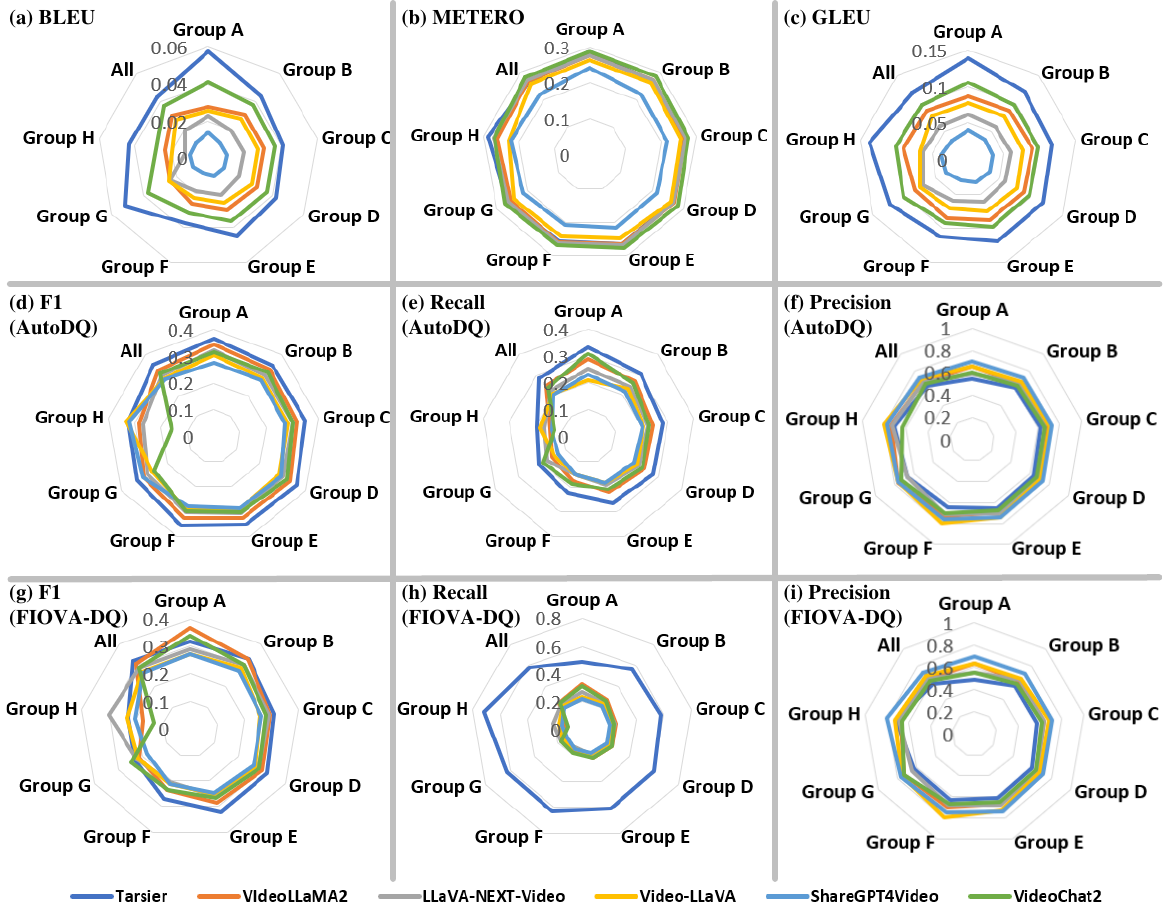}
\vspace{-5pt}
\caption{Radar plot of LVLMs on FIOVA and 8 sub-groups. See App.~\ref{subsec:results-on-8groups} for details.}
\label{fig:radar-metric}
\vspace{-10pt}
\end{figure*}

\textbf{Summary.}
FIOVA integrates:
(i) traditional metrics for lexical overlap and syntactic fluency,
(ii) AutoDQ for interpretable event coverage, and
(iii) FIOVA-DQ for cognitively grounded evaluation of semantic relevance.
While FIOVA-DQ aligns better with human judgment and reduces hallucinations, it is order-agnostic. In a controlled case (Fig.~\ref{fig:example-right}), we construct a model output $M$ that covers three events (A→B→C) in reverse (C→B→A). AutoDQ and FIOVA-DQ assign perfect F1 scores (1.0), but traditional metrics detect structural violations (BLEU: 0.9409, GLEU: 0.9429, METEOR: 0.9054).
This confirms that traditional metrics complement semantic metrics by capturing narrative fluency. Together, the three metric types offer a comprehensive framework for evaluating LVLM-generated captions.

\subsection{Batch Evaluation for Baseline LVLMs}

To evaluate model robustness under varying content complexities, we conducted batch evaluations across eight subgroups defined by inter-annotator variation (CV), as detailed in Fig.~\ref{fig:radar-metric} and App.~\ref{subsec:results-on-8groups}.

\textbf{Batch score evaluation.} 
Performance trends under AutoDQ and FIOVA-DQ remain consistent with overall findings. Tarsier consistently achieves the highest Recall, while most LVLMs exhibit sharp performance drops in GroupH, which contains videos with frequent scene transitions and multi-perspective shifts (CV > 70\%). These videos demand stronger semantic modeling. Tarsier’s relative advantage in GroupH may reflect superior temporal handling. 
In contrast, Precision scores remain stable across groups, indicating reliable identification of salient content. The divergence between BLEU and GLEU in Group~H suggests that traditional metrics are less aligned with semantic variation, especially when lexical phrasing differs.
These results further highlight the limitations of lexical metrics in capturing temporal coherence and semantic completeness, particularly in open-ended captioning scenarios.

\begin{figure*}[t!]
% \vspace{-10pt}
\centering
\includegraphics[width=\columnwidth]{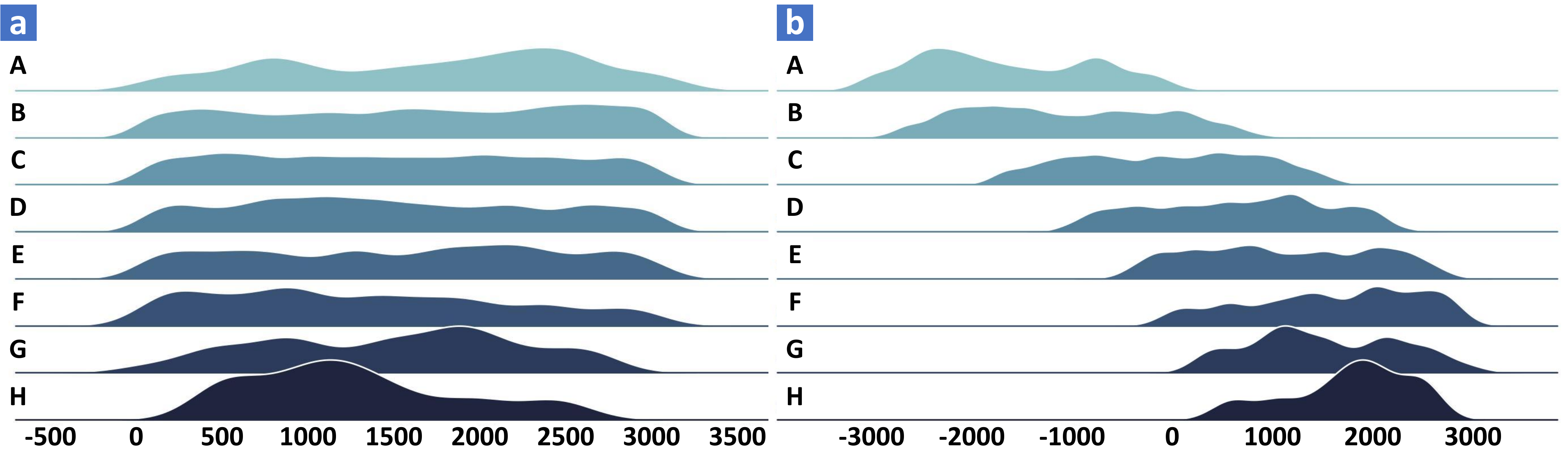}
\vspace{-5pt}
\caption{Comparison between humans and LVLMs based on the ranking of CV (coefficient of variation). (a) Ranking of CV for six LVLMs. (b) Difference between the ranking of CV for six LVLMs and humans.}
\label{fig:CV-rank}
% \vspace{-10pt}
\end{figure*}

\textbf{Batch ranking analysis.} 
To quantify human–model consistency gaps, we conducted a three-step analysis:
(1) computing human annotation consistency across six dimensions (Sec.~\ref{subsec:accessing-annotation-diversity});
(2) measuring LVLM consistency using traditional metrics, AutoDQ, and FIOVA-DQ (Alg.~\ref{alg:cv_lvlm});
(3) comparing CV-based rankings between humans and models (Alg.~\ref{alg:cv_ranking_diff}).
As shown in Fig.~\ref{fig:CV-rank}(a), model output variability decreases from Group~A to Group~H—indicating convergence toward safer, template-like generations under increasing uncertainty. In contrast, human annotations become more diverse as complexity rises, but more stable under low-context conditions.

\textbf{Model-human comparison.} 
Fig.\ref{fig:CV-rank} (b) shows that in Groups A–B, human annotations are more consistent than LVLM outputs, suggesting model instability when inputs are simple and under-constrained (see Fig.~\ref{fig:example--}). In Group H, the trend reverses: LVLMs produce highly consistent outputs, while human annotations diverge (Fig.~\ref{fig:example-+}). Mid-range groups (C–E) exhibit closer alignment, suggesting a complexity-aware equilibrium (Fig.~\ref{fig:example-0}).

\textbf{Summary.}  
Baseline LVLMs tend to prioritize Precision over Recall, especially under high complexity, resulting in semantically accurate yet incomplete descriptions. Their behavior is complexity-sensitive: more variable when the input is simple, more rigid when content is ambiguous. These observations reinforce the need for evaluation metrics that jointly capture lexical similarity, semantic coverage, and structural coherence.

\subsection{Evaluation of SOTA LVLMs on FIOVA$_{\text{hard}}$}
\label{subsec:hard-sota}

\begin{table*}[t!]
\centering
\caption{Performance of 9 LVLMs on FIOVA$_{\text{hard}}$ (based on 8-frame as input)}
\label{tab:sota-hard-metric}

\small
\setlength{\tabcolsep}{2pt}
\begin{tabular}{lccc c ccc c ccc}
\toprule
 & \multicolumn{3}{c}{\textbf{Traditional Metrics}} & & \multicolumn{3}{c}{\textbf{AutoDQ-based Metrics}} & & \multicolumn{3}{c}{\textbf{FIOVA-DQ-based Metrics}} \\
\cmidrule{2-4} \cmidrule{6-8} \cmidrule{10-12}
 & BLEU & METEOR & GLEU & & F1 & Recall & Precision & & F1 & Recall & Precision \\
\midrule
\textbf{Tarsier} & \cellcolor[HTML]{63BE7B}0.040 & \cellcolor[HTML]{7CC991}0.258 & \cellcolor[HTML]{63BE7B}0.116 & & \cellcolor[HTML]{96D3A7}0.349 & \cellcolor[HTML]{98D4A9}0.222 & \cellcolor[HTML]{FCFCFF}0.652 & & \cellcolor[HTML]{C9E8D3}0.262 & \cellcolor[HTML]{63BE7B}0.636 & \cellcolor[HTML]{FCFCFF}0.636 \\
\textbf{VideoLLaMA2} & \cellcolor[HTML]{ABDBB9}0.026 & \cellcolor[HTML]{7BC890}0.259 & \cellcolor[HTML]{9BD5AB}0.085 & & \cellcolor[HTML]{C5E6D0}0.319 & \cellcolor[HTML]{DBEFE2}0.171 & \cellcolor[HTML]{A4D9B3}0.735 & & \cellcolor[HTML]{E8F4EE}0.230 & \cellcolor[HTML]{F8FBFC}0.164 & \cellcolor[HTML]{B4DFC1}0.706 \\
\textbf{LLaVA-NEXT-Video} & \cellcolor[HTML]{C8E7D2}0.020 & \cellcolor[HTML]{6FC385}0.264 & \cellcolor[HTML]{C4E6CF}0.062 & & \cellcolor[HTML]{E6F3EC}0.297 & \cellcolor[HTML]{EEF7F3}0.156 & \cellcolor[HTML]{C4E5CE}0.706 & & \cellcolor[HTML]{F7FAFA}0.214 & \cellcolor[HTML]{FCFCFF}0.152 & \cellcolor[HTML]{D2EBDB}0.677 \\
\textbf{Video-LLaVA} & \cellcolor[HTML]{B9E1C5}0.023 & \cellcolor[HTML]{A9DBB7}0.242 & \cellcolor[HTML]{B3DFC0}0.072 & & \cellcolor[HTML]{F1F8F6}0.290 & \cellcolor[HTML]{F5F9F9}0.151 & \cellcolor[HTML]{63BE7B}0.796 & & \cellcolor[HTML]{F5FAF9}0.216 & \cellcolor[HTML]{FBFCFE}0.156 & \cellcolor[HTML]{63BE7B}0.784 \\
\textbf{ShareGPT4Video} & \cellcolor[HTML]{FCFCFF}0.009 & \cellcolor[HTML]{FCFCFF}0.210 & \cellcolor[HTML]{FCFCFF}0.031 & & \cellcolor[HTML]{F7FAFB}0.286 & \cellcolor[HTML]{FCFCFF}0.146 & \cellcolor[HTML]{8ACE9D}0.759 & & \cellcolor[HTML]{FCFCFF}0.208 & \cellcolor[HTML]{FCFCFF}0.150 & \cellcolor[HTML]{81CA95}0.756 \\
\textbf{VideoChat2} & \cellcolor[HTML]{8BCE9D}0.032 & \cellcolor[HTML]{63BE7B}0.268 & \cellcolor[HTML]{89CE9C}0.095 & & \cellcolor[HTML]{FCFCFF}0.282 & \cellcolor[HTML]{C9E8D3}0.185 & \cellcolor[HTML]{C2E5CD}0.707 & & \cellcolor[HTML]{E4F3EA}0.234 & \cellcolor[HTML]{F3F9F7}0.180 & \cellcolor[HTML]{CBE9D5}0.683 \\ \midrule
\textbf{InternVL-2.5} & \cellcolor[HTML]{DCEFE3}0.015 & \cellcolor[HTML]{A1D7B0}0.245 & \cellcolor[HTML]{B0DEBE}0.073 & & \cellcolor[HTML]{B3DFC0}0.330 & \cellcolor[HTML]{A6D9B5}0.212 & \cellcolor[HTML]{97D3A8}0.748 & & \cellcolor[HTML]{8CCF9F}0.325 & \cellcolor[HTML]{EAF5F0}0.208 & \cellcolor[HTML]{91D1A3}0.740 \\
\textbf{QWen2.5-VL} & \cellcolor[HTML]{D2EBDB}0.017 & \cellcolor[HTML]{91D1A3}0.250 & \cellcolor[HTML]{ABDBB9}0.076 & & \cellcolor[HTML]{D9EEE1}0.305 & \cellcolor[HTML]{BCE2C8}0.195 & \cellcolor[HTML]{BCE2C8}0.713 & & \cellcolor[HTML]{B4DFC1}0.284 & \cellcolor[HTML]{F4F9F8}0.176 & \cellcolor[HTML]{A2D8B1}0.724 \\
\textbf{GPT-4o} & \cellcolor[HTML]{B8E1C4}0.023 & \cellcolor[HTML]{72C488}0.262 & \cellcolor[HTML]{98D4A9}0.087 & & \cellcolor[HTML]{63BE7B}0.382 & \cellcolor[HTML]{63BE7B}0.262 & \cellcolor[HTML]{C9E8D3}0.700 & & \cellcolor[HTML]{63BE7B}0.367 & \cellcolor[HTML]{DDF0E4}0.251 & \cellcolor[HTML]{C4E6CF}0.690 \\
\bottomrule
\end{tabular}
% \vspace{-15pt}
\end{table*}

We evaluate nine LVLMs on FIOVA$_{\text{hard}}$, a high-complexity subset (Groups F–H) characterized by dense events, high annotation variance, and multi-perspective ambiguity. To assess its diagnostic value, we compare model performance on FIOVA$_{\text{hard}}$ (Tab.~\ref{tab:sota-hard-metric}) with full-dataset results (Tab.~\ref{tab:metric-result}).

Tarsier stands out as the most recall-oriented model, achieving a FIOVA-DQ Recall of 0.636 on FIOVA$_{\text{hard}}$—an improvement over its full-dataset score (0.584). Although its F1 drops from 0.320 to 0.262 due to lower Precision, the gain in Recall demonstrates robustness under semantic complexity.
Most baselines show performance drops on FIOVA$_{\text{hard}}$. VideoChat2’s F1 declines from 0.287 to 0.234, with Recall dropping from 0.243 to 0.180. Video-LLaVA follows a similar pattern. ShareGPT4Video, while achieving the highest Precision (0.756), suffers the lowest Recall (0.150) and a reduced F1 of 0.208. These results confirm that FIOVA$_{\text{hard}}$ effectively stresses models on multi-event, causally entangled content.

Among newer models, GPT-4o achieves the highest FIOVA-DQ F1 (0.367), indicating strong overall alignment. However, InternVL-2.5 and Qwen2.5-VL attain higher Precision (0.740 and 0.724), suggesting better coverage of key events. This implies GPT-4o maintains a good balance between recall and conciseness but may overlook highly salient moments.
Frame-length ablation (Tab.~\ref{tab:frame-ablation}) reveals temporal sensitivity across models. GPT-4o’s F1 improves from 0.367 to 0.418 when increasing input frames from 8 to 32. Qwen2.5-VL follows a similar trend (F1: 0.284 → 0.337). InternVL-2.5 attains peak Precision (0.743) at 32 frames, while Tarsier consistently maintains high Recall (~0.638), reflecting distinct modeling strategies.

\textbf{Summary.}
\textsc{FIOVA}$_\text{hard}$ serves as a robust stress-testing benchmark for semantic and narrative complexity. It exposes divergence in model behaviors and strategy trade-offs between coverage and precision. FIOVA-DQ proves particularly effective in capturing these nuances, offering actionable insights for advancing LVLM robustness and alignment.

\begin{table*}[t!]
\centering
\caption{Frame-length ablation results on \textsc{FIOVA}$_{\text{hard}}$. }

\label{tab:frame-ablation}
\small
\setlength{\tabcolsep}{3pt}
\begin{tabular}{lccc c ccc c ccc}
\toprule
\textbf{LVLMs} & \multicolumn{3}{c}{\textbf{Traditional Metrics}} & & \multicolumn{3}{c}{\textbf{AutoDQ-based Metrics}} & & \multicolumn{3}{c}{\textbf{FIOVA-DQ-based Metrics}} \\
\cmidrule{2-4} \cmidrule{6-8} \cmidrule{10-12}
 & BLEU & METEOR & GLEU & & F1 & Recall & Precision & & F1 & Recall & Precision \\
\midrule
% \textbf{Tarsier (8)}       & \cellcolor[HTML]{7EC993}0.040 & \cellcolor[HTML]{C1E4CC}0.258 & \cellcolor[HTML]{66C07E}0.116 &&
%                                \cellcolor[HTML]{C2E5CD}0.349 & \cellcolor[HTML]{D5EDDE}0.222 & \cellcolor[HTML]{D4ECDD}0.652 &&
%                                \cellcolor[HTML]{FCFCFF}0.262 & \cellcolor[HTML]{6AC181}0.636 & \cellcolor[HTML]{F1F8F6}0.636 \\
\textbf{Tarsier (16)}      & \cellcolor[HTML]{63BE7B}0.046 & \cellcolor[HTML]{63BE7B}0.280 & \cellcolor[HTML]{63BE7B}0.117 &&
                               \cellcolor[HTML]{C2E5CD}0.349 & \cellcolor[HTML]{C9E8D3}0.235 & \cellcolor[HTML]{B8E1C5}0.679 &&
                               \cellcolor[HTML]{EAF5EF}0.281 & \cellcolor[HTML]{63BE7B}0.657 & \cellcolor[HTML]{E1F1E8}0.648 \\
\textbf{Tarsier (32)}      & \cellcolor[HTML]{6DC284}0.044 & \cellcolor[HTML]{63BE7B}0.280 & \cellcolor[HTML]{69C180}0.115 &&
                               \cellcolor[HTML]{CDE9D6}0.338 & \cellcolor[HTML]{D4ECDD}0.223 & \cellcolor[HTML]{A2D8B2}0.700 &&
                               \cellcolor[HTML]{F0F8F5}0.274 & \cellcolor[HTML]{69C181}0.638 & \cellcolor[HTML]{FCFCFF}0.627 \\
% \addlinespace
% \textbf{InternVL-2.5 (8)}  & \cellcolor[HTML]{F5FAF9}0.015 & \cellcolor[HTML]{FCFCFF}0.245 & \cellcolor[HTML]{D3ECDC}0.073 &&
%                                \cellcolor[HTML]{D4ECDD}0.330 & \cellcolor[HTML]{DFF0E6}0.212 & \cellcolor[HTML]{71C487}0.748 &&
%                                \cellcolor[HTML]{BFE4CA}0.325 & \cellcolor[HTML]{EFF7F4}0.208 & \cellcolor[HTML]{67C07F}0.740 \\
\textbf{InternVL-2.5 (16)} & \cellcolor[HTML]{EEF7F3}0.017 & \cellcolor[HTML]{DDF0E4}0.252 & \cellcolor[HTML]{CFEAD8}0.075 &&
                               \cellcolor[HTML]{FCFCFF}0.288 & \cellcolor[HTML]{FCFCFF}0.179 & \cellcolor[HTML]{75C68B}0.743 &&
                               \cellcolor[HTML]{CBE9D5}0.312 & \cellcolor[HTML]{EDF6F2}0.214 & \cellcolor[HTML]{79C78E}0.727 \\
\textbf{InternVL-2.5 (32)} & \cellcolor[HTML]{FCFCFF}0.014 & \cellcolor[HTML]{EAF5EF}0.249 & \cellcolor[HTML]{D6EDDF}0.072 &&
                               \cellcolor[HTML]{EDF6F2}0.304 & \cellcolor[HTML]{F2F8F7}0.190 & \cellcolor[HTML]{63BE7B}0.760 &&
                               \cellcolor[HTML]{B9E1C5}0.331 & \cellcolor[HTML]{F4F9F8}0.193 & \cellcolor[HTML]{63BE7B}0.743 \\
% \addlinespace
% \textbf{Qwen2.5-VL (8)}    & \cellcolor[HTML]{ECF6F1}0.017 & \cellcolor[HTML]{E3F2EA}0.250 & \cellcolor[HTML]{CBE9D5}0.076 &&
%                                \cellcolor[HTML]{ECF6F1}0.305 & \cellcolor[HTML]{EEF7F3}0.195 & \cellcolor[HTML]{95D3A6}0.713 &&
%                                \cellcolor[HTML]{E7F4ED}0.284 & \cellcolor[HTML]{F9FBFC}0.176 & \cellcolor[HTML]{7DC992}0.724 \\
\textbf{Qwen2.5-VL (16)}   & \cellcolor[HTML]{D6EDDF}0.022 & \cellcolor[HTML]{ACDCBA}0.263 & \cellcolor[HTML]{BCE2C8}0.082 &&
                               \cellcolor[HTML]{E4F3EB}0.313 & \cellcolor[HTML]{E8F4EE}0.201 & \cellcolor[HTML]{98D4A9}0.710 &&
                               \cellcolor[HTML]{D0EAD9}0.308 & \cellcolor[HTML]{FCFCFF}0.164 & \cellcolor[HTML]{98D4A8}0.704 \\
\textbf{Qwen2.5-VL (32)}   & \cellcolor[HTML]{D6EDDE}0.022 & \cellcolor[HTML]{7BC890}0.275 & \cellcolor[HTML]{C5E6D0}0.079 &&
                               \cellcolor[HTML]{D0EAD9}0.335 & \cellcolor[HTML]{D6EDDE}0.221 & \cellcolor[HTML]{AADBB9}0.692 &&
                               \cellcolor[HTML]{B3DFC0}0.337 & \cellcolor[HTML]{FAFBFD}0.173 & \cellcolor[HTML]{8DCF9F}0.712 \\
% \addlinespace
% \textbf{GPT-4o (8)}        & \cellcolor[HTML]{D2EBDB}0.023 & \cellcolor[HTML]{B0DDBD}0.262 & \cellcolor[HTML]{B1DEBE}0.087 &&
%                                \cellcolor[HTML]{A2D8B2}0.382 & \cellcolor[HTML]{B1DEBE}0.262 & \cellcolor[HTML]{A2D8B1}0.700 &&
%                                \cellcolor[HTML]{95D3A7}0.367 & \cellcolor[HTML]{E2F2E8}0.251 & \cellcolor[HTML]{A9DBB7}0.690 \\
\textbf{GPT-4o (16)}       & \cellcolor[HTML]{E5F3EB}0.019 & \cellcolor[HTML]{86CD99}0.272 & \cellcolor[HTML]{F7FAFB}0.059 &&
                               \cellcolor[HTML]{6DC284}0.438 & \cellcolor[HTML]{6AC181}0.340 & \cellcolor[HTML]{FCFCFF}0.614 &&
                               \cellcolor[HTML]{73C589}0.402 & \cellcolor[HTML]{DBEFE3}0.271 & \cellcolor[HTML]{BFE3CA}0.674 \\
\textbf{GPT-4o (32)}       & \cellcolor[HTML]{F3F9F7}0.016 & \cellcolor[HTML]{93D2A5}0.269 & \cellcolor[HTML]{FCFCFF}0.057 &&
                               \cellcolor[HTML]{63BE7B}0.448 & \cellcolor[HTML]{63BE7B}0.347 & \cellcolor[HTML]{EBF5F0}0.631 &&
                               \cellcolor[HTML]{63BE7B}0.418 & \cellcolor[HTML]{DEF0E5}0.264 & \cellcolor[HTML]{DDF0E4}0.651 \\
\bottomrule
\end{tabular}
% \vspace{-15pt}
\end{table*}

\subsection{Summary}

We conducted a comprehensive evaluation of nine representative LVLMs on the FIOVA benchmark using a tri-layer framework that integrates traditional lexical metrics, event-level semantic metrics (AutoDQ), and cognitively weighted evaluations (FIOVA-DQ). The results reveal several key insights: 
(1) \textbf{Performance trade-offs:} Model behaviors diverge along the precision–recall axis. Tarsier achieves the highest Recall (0.636) on FIOVA$_\text{hard}$, reflecting strong coverage in complex settings. GPT-4o yields the highest overall F1 score (0.367), balancing coverage and conciseness. InternVL-2.5 and Qwen2.5-VL lead in Precision (0.740 and 0.724), effectively avoiding redundancy while capturing salient events.
(2) \textbf{Human alignment:} A user study confirms that FIOVA-DQ aligns better with human preferences than AutoDQ (Spearman F1: 0.579 vs. 0.125), validating the cognitive weighting derived from multi-annotator consensus.
(3) \textbf{Complexity sensitivity:} CV-based analysis shows an inverse trend—human annotations become more diverse as content complexity increases, whereas LVLM outputs become more uniform. This suggests a tendency toward safe, template-like generation under uncertainty.
(4) \textbf{Failure patterns:} Models still exhibit hallucinations and omissions. ShareGPT4Video often generates irrelevant details, while Tarsier occasionally skips key content. Although models perform reliably on structured videos (\eg, sports practice), their performance degrades significantly when facing temporal inconsistencies or scene transitions.
(5) \textbf{Evaluation implications:} Traditional metrics remain essential for detecting syntactic and narrative disruptions. While AutoDQ and FIOVA-DQ effectively assess semantic coverage and cognitive salience, they are agnostic to event order. The tri-layer framework of FIOVA provides a methodologically complementary foundation: lexical metrics evaluate fluency and structure, AutoDQ captures event coverage, and FIOVA-DQ emphasizes human-centered relevance.

In conclusion, FIOVA establishes a cognitively aligned evaluation paradigm that jointly considers structural, semantic, and human-centric dimensions. We argue that future LVLMs must move beyond event identification to model temporal logic, narrative coherence, and contextual awareness—while evaluation frameworks should evolve toward greater semantic depth and interpretability.

\section{Conclusions}

We present FIOVA, a cognitively aligned benchmark designed to evaluate the video captioning capabilities of LVLMs from structural, semantic, and human-preference perspectives. By integrating diverse human annotations and introducing FIOVA-DQ—an event-based metric with cognitive weighting—FIOVA enables fine-grained and interpretable comparisons between machine- and human-generated video descriptions. Our evaluation reveals that while LVLMs demonstrate basic perceptual and temporal reasoning abilities, they still face persistent challenges such as incomplete event coverage, hallucinated content, and stylistic rigidity, particularly in complex, multi-event scenarios. Furthermore, we show that traditional, event-driven, and cognitively weighted metrics offer complementary insights, and should be jointly considered in future evaluation. We hope FIOVA will serve as a foundation for advancing both the modeling and assessment of long-video understanding, guiding the development of LVLMs that are more complete, coherent, and aligned with human judgment.

\clearpage
% \section*{References}

%%%%%%%%%%%%%%%%%%%%%%%%%%%%%%%%%%%%%%%%%%%%%%%%%%%%%%%%%%%%
\clearpage

\renewcommand{\thetable}{A\arabic{table}}
\renewcommand{\thefigure}{A\arabic{figure}}
\renewcommand{\thealgorithm}{A\arabic{algorithm}}
\setcounter{figure}{0}  
\setcounter{table}{0}  
\setcounter{algorithm}{0}  

\appendix

\section{Background Supplement and Appendix Overview}
\label{sec:app-overview}

This work is grounded in our team’s extensive research on video understanding, multimodal analysis, and human-machine evaluation. Over the past few years, we have developed a suite of high-impact benchmark platforms—published in \textbf{TPAMI}, \textbf{IJCV}, and \textbf{NeurIPS}—to support rigorous evaluation in real-world, cognitively challenging scenarios. Our prior contributions include \textbf{GOT-10k}~\cite{huang2019got} (TPAMI'21), \textbf{VideoCube}~\cite{hu2023global} (TPAMI'23), and \textbf{MGIT}~\cite{hu2024multi} (NeurIPS'23) for video object tracking, multimodal alignment, and hierarchical annotation; \textbf{SOTVerse}~\cite{sotverse} (IJCV'24) for open-world task space construction and fine-grained challenge modeling; and \textbf{BioDrone}~\cite{zhao2024biodrone} (IJCV'24) for evaluation under embodied platform conditions. We also introduced the \textbf{Visual Turing Test framework}~\cite{zhao2025visual}, enabling principled human-machine comparisons in dynamic vision tasks.

Building upon these foundations, FIOVA advances the evaluation paradigm from object-level perception to event-level narrative alignment, transitioning from static, single-reference lexical metrics to a cognitively grounded, multi-annotator evaluation framework. The proposed \textbf{FIOVA-DQ} metric draws directly from our previous work on \textbf{human agreement modeling}, \textbf{semantic diversity quantification}, and \textbf{trustworthy evaluation design}, ensuring methodological continuity and reinforcing scientific reliability across task, data, and metric design.

This appendix begins with Appendix~\ref{sec:related-work}, which presents a comprehensive review of related work and situates our contribution within the broader research landscape. The subsequent sections provide in-depth documentation of dataset design, annotation protocol, model configuration, experimental analysis, and evaluation procedures. Our goal is to offer a transparent and complete account of the system design, implementation, and scientific foundation that support this benchmark.

\vspace{0.5em}
\noindent\textbf{Appendix Structure:}
\begin{itemize}
    \item Appendix~\ref{sec:related-work}: Detailed introduction and analysis of related work;
    \item Appendix~\ref{sec:detailed-fiova}: Dataset construction and annotation protocol;
    \item Appendix~\ref{sec:human-gt-evaluation}: Human preference experiments and alignment validation;
    \item Appendix~\ref{sec:calculation-cv}: Benchmark models and evaluation settings;
    \item Appendix~\ref{sec:prompts}: Unified caption synthesis and prompt design;
    \item Appendix~\ref{sec:detailed-exp}: Supplementary experiments and performance breakdown;
    \item Appendix~\ref{sec:limitations}: Limitations and societal considerations.
\end{itemize}

\clearpage

\section{Comprehensive Related Works}
\label{sec:related-work}

\subsection{LVLMs for Video Caption}
\label{subsec:related-work-lvlms}

Recent advancements in large vision-language models (LVLMs) have significantly advanced video captioning, improving models' abilities to bridge visual perception and linguistic representation. State-of-the-art LVLMs, including Koala~\cite{tan2024koala}, VTimeLLM~\cite{huang2024vtimellm}, VideoAgent~\cite{fan2025videoagent}, X-LLM~\cite{panagopoulou2025x}, Vamos~\cite{wang2025vamos}, and Text2Video~\cite{korbar2025text}, have achieved performance comparable to GPT-4o~\cite{gpt4} in tasks such as visual question answering, video description, and multimodal reasoning. However, these models still face considerable challenges when dealing with temporal reasoning, long-range coherence, and complex narrative structures in long videos.

A critical research direction in LVLMs involves frame encoding and temporal relationship modeling to enhance video comprehension by capturing spatiotemporal dependencies. For example, Tarsier~\cite{tarsier} employs CLIP-ViT for efficient single-frame encoding and utilizes a large language model (LLM) for modeling semantic relations across frames, offering a structured, two-stage training approach for video captioning. VideoLLaMA2~\cite{videollama2} further advances this direction by proposing a Spatio-Temporal Convolution (STC) connector, capturing interactions between spatial features and temporal dynamics to improve context-aware captioning capabilities.

Cross-modal alignment between visual and linguistic representations has also emerged as a crucial area in LVLM development. Video-LLaVA~\cite{videollava} explicitly addresses this by aligning visual features with linguistic embedding spaces, ensuring the semantic accuracy and faithfulness of generated captions. ShareGPT4Video~\cite{sharegpt4video} emphasizes fine-grained video understanding, promoting more detailed, event-level caption generation.

Moreover, spatiotemporal reasoning and causal inference have become increasingly significant in video understanding research. VideoChat2~\cite{videochat2} proposes a learnable neural interface for enhanced event localization and causal analysis. Meanwhile, ST-LLM~\cite{st-llm} introduces a dynamic spatiotemporal tokenization mechanism, enabling efficient processing of long video sequences while maintaining output stability.

Recent studies have also emphasized modeling narrative coherence in long-form videos. LongVLM~\cite{longvlm} designs a hierarchical token-merging strategy that encodes local features while maintaining global semantic consistency, effectively improving narrative coherence across extended video content. VideoAgent~\cite{videoagent} further explores an agent-based reasoning framework, iteratively selecting and processing video segments to generate more structured, contextually coherent captions.

In addition, a series of recently developed LVLMs have further enriched the research landscape. InternVL-2.5~\cite{InternVL} has demonstrated advanced multimodal fusion capabilities, significantly improving performance through refined visual feature encoding and cross-modal representation learning. Qwen2.5-VL~\cite{Qwen}, introduced by Alibaba, employs efficient visual-language alignment techniques to enhance cross-modal understanding in video tasks. Other prominent models include Gemini~\cite{team2023gemini} and CogVLM2~\cite{hong2024cogvlm2}, each showcasing exceptional performance and innovative designs. Specifically, Gemini is reported to achieve or even surpass GPT-4o-level performance in multimodal reasoning, and CogVLM2 has gained attention for its open-source availability and high performance.

These rapid developments highlight a critical yet unresolved research question: “\textit{Can video-based LVLMs achieve the depth, accuracy, and coherence of human annotators in generating video descriptions?}”
Motivated by these recent advancements, we introduce the FIOVA benchmark and systematically evaluate representative LVLMs to reveal their strengths and limitations in complex long-video scenarios. Our goal is to provide a robust evaluation framework and insights to guide future research and development in LVLMs.

\subsection{Video Caption Dataset}
\label{subsec:video-caption-dataset}

The rapid progress in video captioning heavily relies on large-scale datasets that accurately reflect the complexity and diversity of real-world videos. These datasets provide essential benchmarks for model training and evaluation, significantly advancing video understanding technologies. Over the past years, researchers have proposed various datasets, differing in scale, domain coverage, and annotation richness, shaping the trajectory of research in video captioning.

YouCook-II~\cite{youcook2} is a widely adopted dataset consisting of 2,000 cooking-related video clips spanning 89 distinct recipes. Sourced from YouTube, these videos feature diverse camera angles, significant variations in motion, lighting, and scenes, making YouCook-II an important benchmark for assessing model robustness in dynamic and unstructured scenarios. The MSVD dataset~\cite{msvd} contains 1,970 short videos paired with human-annotated descriptions, widely utilized in general video captioning tasks due to its diversity and open-domain nature. MSR-VTT~\cite{msrvtt} extends this further, comprising 7,180 videos divided into 10,000 clips across 20 real-world categories such as sports and news, serving as a critical standard for evaluating models' cross-domain generalization. Panda-70M~\cite{panda70m}, the largest dataset to date, provides 70 million video clips annotated with high-quality captions, greatly enhancing resources for training models capable of understanding complex, open-world scenarios.

Recently introduced datasets such as LVBench~\cite{wang2024lvbench} and Video-MME~\cite{fu2024video} have further advanced research into long-form video understanding. LVBench focuses explicitly on very long videos with complex multimodal reasoning and extreme temporal understanding tasks. Similarly, Video-MME emphasizes multimodal fusion and the modeling of long video sequences. However, these benchmarks typically utilize automatically generated captions or single-source annotations, thus lacking comprehensive multi-annotator semantic assessments. In contrast, the FIOVA benchmark targets medium-length videos, uniquely emphasizing semantic diversity and consistency through multiple human annotations, thereby addressing a critical gap left by existing long-video datasets.

The FIOVA benchmark provides detailed annotations from multiple human annotators, offering a more authentic reflection of human variability and consensus in video understanding. Furthermore, its captions are notably longer and richer in detail compared to existing datasets. This makes FIOVA a crucial tool for evaluating whether models can generate nuanced, semantically accurate, and human-aligned video descriptions.

\subsection{Video Caption Evaluation}
\label{subsuec:related-work-evaluation}

Early research in video captioning primarily focused on pretraining video-language models, which were then fine-tuned on task-specific datasets. Traditional evaluation metrics such as BLEU~\cite{bleu}, GLEU~\cite{gleu}, METEOR~\cite{meteor}, and CIDEr~\cite{cider}, widely employed for their effectiveness in assessing lexical and syntactic similarity, often resulted in models overfitting to specific datasets. Consequently, these models had limited generalization capability when applied to diverse or open-world video scenarios~\cite{tarsier}.

To address these limitations, recent research has shifted toward zero-shot video captioning tasks~\cite{tewel2022zero, wang2019learning, zhou2024actionhub}, enabling models to describe unseen videos without task-specific fine-tuning. However, most existing benchmarks feature short and structurally simple videos, inadequately assessing models' ability to handle complex narratives and multi-event sequences.

As videos grow more complex—with longer durations, richer visual diversity, and intricate temporal dependencies—traditional metrics struggle to fully capture semantic relevance and contextual coherence. To tackle this challenge, recent studies have explored using LLMs such as ChatGPT for automatic evaluation~\cite{sottana2023evaluation}. While this approach provides flexible and rich assessment capabilities, enhancing the objectivity and interpretability of scoring frameworks remains an open issue~\cite{maaz2023video}.

The AutoDQ metric~\cite{tarsier}, proposed by researchers at ByteDance, marked a significant advancement in evaluation methods for video captioning. Unlike traditional metrics, AutoDQ extracts event-level representations from both generated captions and reference annotations, comparing them through precision, recall, and F1 scores. This event-based approach enhances the evaluation accuracy regarding the completeness and correctness of video descriptions, establishing a valuable foundation for subsequent research.

AuroraCap~\cite{chai2024auroracap} further contributed novel ideas in the evaluation of long-video captions. AuroraCap introduced a structured dataset of long captions (VDC) along with its corresponding evaluation metric, VDCscore, which utilizes large language models in a divide-and-conquer, question-answering approach. This refined evaluation strategy significantly improves semantic sensitivity and reliability, demonstrating an alternative and innovative direction in evaluating long-video captions.

Building upon these pioneering studies, we introduce the FIOVA-DQ metric. Consistent with AutoDQ's emphasis on event extraction and structured comparisons, FIOVA-DQ innovatively incorporates cognitive weighting derived from multi-annotator consensus to explicitly address the varying importance of events within a video narrative. Unlike AuroraCap’s indirect evaluation via LLM-based question-answering, FIOVA-DQ directly integrates human annotation consensus to assign differentiated event importance weights, aligning evaluation processes more intuitively with human preferences and significantly enhancing consistency with human cognition.

In this study, based on the FIOVA benchmark, we comprehensively evaluate representative LVLMs using traditional lexical metrics (BLEU, GLEU, METEOR), the event-driven metric AutoDQ, and our cognitively weighted metric FIOVA-DQ. Our goal is to systematically reveal the strengths and limitations of current LVLMs in complex video scenarios, facilitating the development of video understanding techniques that align more closely with human cognitive preferences.

\clearpage

\section{Detailed Information of FIOVA Dataset}
\label{sec:detailed-fiova}

\subsection{Theme Abbreviations and Corresponding Meanings}
\label{subsec:fiova-theme}

\begin{table}[h!]
\caption{The video themes of the FIOVA dataset.}
\label{tab:fiova-theme}
\small 
\setlength{\tabcolsep}{3pt}
\begin{tabularx}{\textwidth}{>{\bfseries}p{2.2em} p{4.7cm} X}
\toprule
\textbf{Prefix} & \textbf{Video Theme} & \textbf{Description} \\
\midrule
acc & Accident & Sudden events such as traffic accidents and unexpected collisions. \\
ad & Advertisement & Commercial advertisements and product promotions for marketing communication. \\
ch & Children & Scenes of children’s daily activities, play, and interactions. \\
di & Dialogue & Video content featuring conversations, discussions, and communicative interactions. \\
do & Daily Observations & Observations and events from everyday life. \\
doc & Documentary & Documentaries with educational, informational, or historical content. \\
duc & Daily Unique Content & Unique or unusual events in daily life. \\
ear & Event Action Record & Actions and behaviors during specific activities or events. \\
ex & Examination & Records of exams, tests, or other assessment activities. \\
fa & Family Activities & Family activities, parent-child interactions, and family life. \\
fi & Film Industry & Film production, actor performances, and behind-the-scenes content. \\
fu & Fun & Entertaining, fun, or humorous content. \\
gar & Gathering Activities & Social activities, gatherings, and collective events. \\
goa & Games of Action & Action games, sports competitions, and outdoor activities. \\
hom & Home & Home environments, domestic life, and family relationships. \\
ken & Kinetic Engaging Narratives & Dynamic participation, physical activities, and interactive narratives. \\
ki & Kids Interaction & Interactions and social activities among children. \\
mo & Motion & Physical movement, action displays, and dynamic expressions. \\
mod & Movement Onsite Display & On-site activities, movements, and mobility. \\
mot & Motor & Mechanical motion, vehicle operation, and engine functionality. \\
mu & Music & Music performances, music creation, and musical activities. \\
ne & News Event & News reports, news events, and news interviews. \\
pon & People's Ordinary Narratives & Ordinary people’s daily lives and personal stories. \\
pu & Public Utility & Public services, public utilities, and municipal engineering. \\
rab & Recreational Activities & Recreational activities, leisure behaviors, and entertainment venues. \\
sad & Sports and Daily Activities & Sports activities, daily exercises, and outdoor activities. \\
sc & Scholarly Contexts & Scholarly research, educational contexts, and academic discussions. \\
sch & Social and Cultural Happenings & Social events, cultural activities, and community life. \\
sp & Sports & Sports, physical exercises, and competitive activities. \\
the & Typical Human Experiences & Universal emotions and everyday challenges. \\
tr & Thematic Representation & Thematic presentations, topic discussions, and events. \\
va & Vacation and Activities & Leisure travel, vacation activities, and holiday experiences. \\
vi & Various Interactions & Social interactions and interpersonal relationships. \\
wat & Wildlife and Adventure Themes & Wildlife, adventure activities, and nature exploration. \\
win & Warm Interactive Narratives & Warm interactions, touching stories, and positive communications. \\
xin & Experience Interaction Narratives & Experiential interactions, event narratives, and personal experiences. \\
you & Youthful Unison Observed & Collective activities among young people, teamwork, and youthful vitality. \\
zok & Zoom Occurrences Kinetics & Fast-paced actions, dynamic events, and high-energy activities. \\
\bottomrule
\end{tabularx}
\end{table}

\clearpage

\begin{figure}[t!]
\centering
\includegraphics[width=\textwidth]{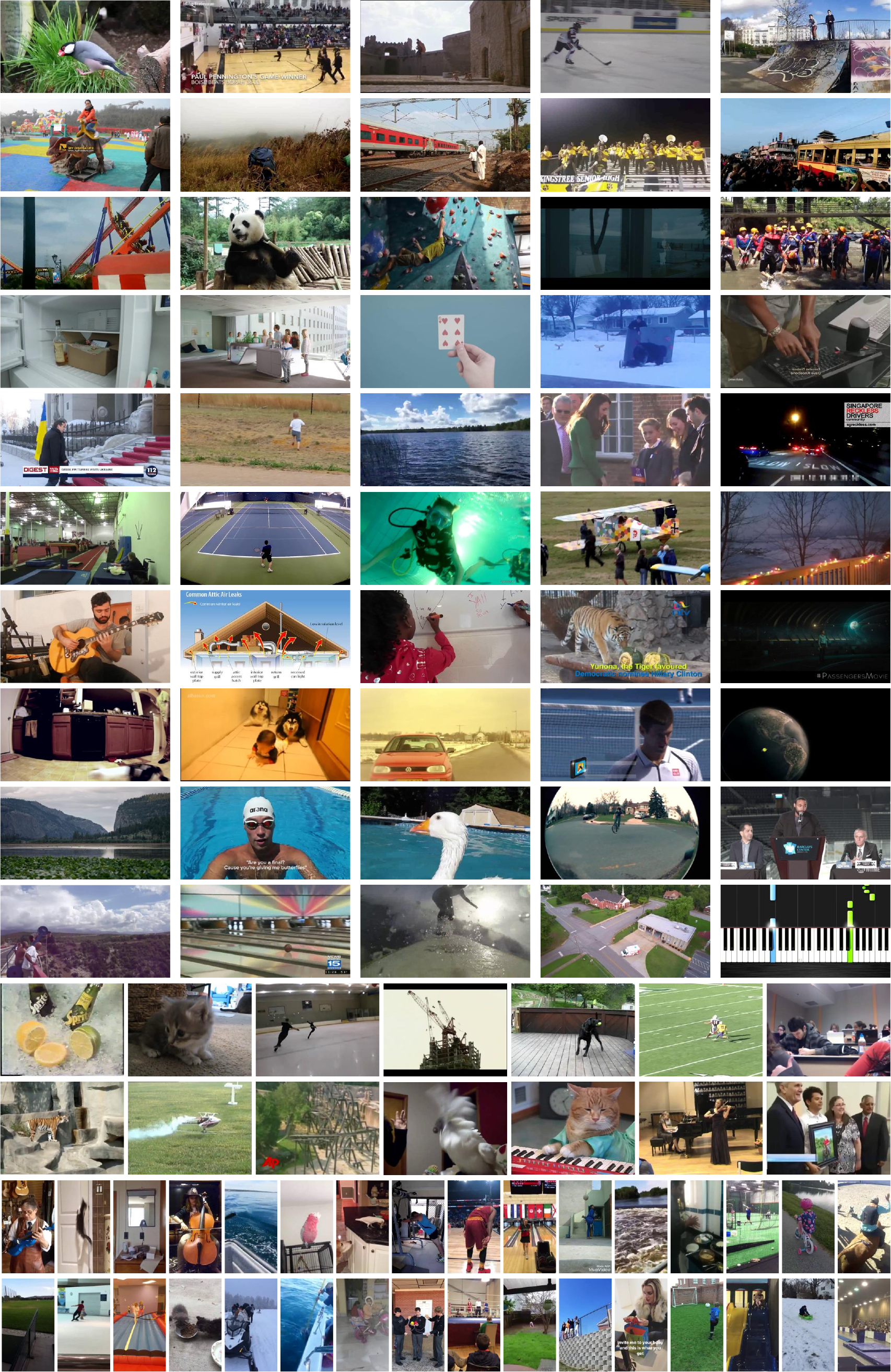}
\caption{The representative data of FIOVA. Each video is strictly selected based on themes.}
\label{fig:example}
\end{figure}

\clearpage

The FIOVA dataset is designed with legal compliance, thematic diversity, and high-quality video content to serve as a robust benchmark for evaluating LVLMs in complex video understanding tasks. It encompasses a broad spectrum of real-world scenarios, structured into 38 distinct themes (Tab.~\ref{tab:fiova-theme}), ranging from everyday activities and social interactions to dynamic events and specialized domains. This thematic diversity ensures a balanced representation of temporal complexity, semantic diversity, and visual richness, providing LVLMs with a comprehensive evaluation environment.

All videos in FIOVA are sourced from copyright-compliant public platforms, strictly adhering to licensing agreements and ethical guidelines. The dataset selection process ensures that only videos explicitly permitted for non-commercial research are included. Furthermore, platform-specific regulations and copyright policies are carefully reviewed to prevent the inclusion of proprietary or restricted content.

To construct a dataset that captures the real-world challenges of video comprehension, FIOVA integrates a wide range of videos, spanning different spatiotemporal dynamics and contextual complexities. The dataset features diverse motion patterns, frequent scene transitions, multi-agent interactions, and varied perspectives, replicating the intricate nature of human video perception.

To maintain high quality and fairness in evaluation, a multi-stage screening process was implemented. Initially, videos were filtered based on copyright compliance, thematic relevance, and visual clarity. A manual verification phase followed, where annotators ensured the completeness of key events and consistency in narrative structure. Finally, dataset composition was carefully balanced in terms of video length, event density, and thematic representation, preventing any single category from dominating the benchmark.

Through this systematic selection process, FIOVA ensures a legally compliant, diverse, and high-quality dataset, establishing a rigorous foundation for benchmarking LVLMs in long-video understanding.

\clearpage

\subsection{Human Annotation Rules}
\label{subsec:annotation-rules}

\begin{center}
\begin{tcolorbox}[colback=gray!00,%gray background
                  colframe=c1,% black frame colour
                %   width=17.2cm,% Use 8cm total width,
                  arc=1.5mm, auto outer arc,
                  breakable,
                  left=0.9mm, right=0.9mm,
                  boxrule=0.9pt,
                  title = {Annotation Scheme and Standards.}
                 ]
1. \textbf{Annotators should label based solely on the visual content of the video, without referring to the audio content or any subtitles in the video}, except for the text that appears naturally in the scene (such as store signs, road signs, \etc). Annotators can choose to use this information or not, based on their judgment.\\
- \textbf{Example:} If a news image appears with the title “Earthquake Report,” this text can be referenced. However, if text appears in the form of movie subtitles at the bottom of the video, it should not be used.\\

2. \textbf{Annotators should describe each video using a few simple declarative sentences to form a paragraph.} The number of sentences depends on the changes in events and scenes in the video, and the content can be appropriately enriched.\\

3. \textbf{Introduce simple and observable scene information}, such as time (morning, noon, evening, late night), location (\eg, on a basketball court, beside a highway, in a bar), and the main objects and their positions in the scene (\eg, a truck overturned in the middle of the road, spectators filling the stands around the stadium). Avoid using overly literary descriptions.\\

4. \textbf{Do not include the names of public figures} in the video, such as Obama, Clinton, Sun Yang, Yao Ming, Yang Mi, \etc Use third-person references such as “a man,” “a woman,” “a boy,” “a girl,” “he,” “she,” \etc, instead.\\

5. \textbf{Optionally include observable details of characters}, such as clothing, hairstyle and color, age, \etc, \eg, “A basketball player wearing a white jersey dribbled past another player wearing a black jersey.”\\

6. \textbf{Describe the behaviors and actions of individual characters as well as interactions between them.} For interactions between multiple people, use references such as “this person, that person,” “one person, another person,” “the one on the left, in the middle, on the right,” or “this group, that group” to refer to different entities. There are no strict requirements for the specific language used, but the relationships and actions must be clearly and concisely described.\\

7. \textbf{Do not use emotionally biased words} (mostly adjectives or adverbs), such as “pitiful,” “disgusting,” “joyfully,” \etc\\

8. \textbf{Do not use idioms} (\eg, “a dime a dozen”), \textbf{proverbs} (\eg, “No pain no gain”), or \textbf{internet slang} (\eg, “imho (in my humble opinion)”).\\

9. \textbf{Do not use overly literary descriptions} or speculate on the psychological state of characters. For example: “As she thought of her youth slipping away, a faint sorrow appeared on her face.”\\

10. \textbf{Do not use subjective inference terms}, such as “obviously.” Sentences should be concise; use shorter phrases where possible, \eg, replace “at the same time” with “meantime.”\\

11. \textbf{Do not use unnecessary conjunctions} if there is no causal relationship between events in the video.\\

12. \textbf{Avoid redundant or conversational language.} For example: instead of “Just after Andy rode his bike home, he immediately ran out again,” simplify to “After a boy rode home, he ran out again” or “A boy rode home and then ran out again.”\\

13. \textbf{The events in the video must be described in the order in which they occur}, without skipping ahead or using summarizing language.
\end{tcolorbox}
\end{center}

Ensuring the quality and robustness of annotations in the FIOVA dataset required a systematic annotator assignment strategy, designed to optimize annotation diversity while maintaining consistency and accuracy. This section details the approach taken to structure annotation assignments and the measures implemented to ensure high-quality annotations.

\paragraph{Annotator Assignment}

Unlike datasets annotated videos by a fixed individual, FIOVA employs a dynamic annotator assignment strategy. Each video is independently annotated by five different individuals, but the assigned annotators vary across different videos. This approach prevents annotator-specific biases and introduces a wider range of descriptive perspectives, enhancing linguistic diversity and semantic coverage.

To maintain annotation consistency, all annotators underwent rigorous training before contributing to the dataset. Standardized guidelines were provided to ensure uniform annotation criteria, covering aspects such as event segmentation, temporal coherence, and neutral descriptive tone. This training phase was crucial in maintaining annotation quality across annotators with varying backgrounds.

\paragraph{Ensuring Annotation Diversity and Robustness}

The adoption of multiple annotator groups serves a dual purpose: increasing annotation diversity while improving the dataset’s adaptability to different evaluation scenarios. This approach introduces heterogeneous descriptive styles, reducing the risk of bias that arises when relying on a fixed group of annotators. It also enhances semantic completeness, as different annotators capture distinct details and perspectives, leading to richer and more informative video descriptions.

Moreover, the variability introduced by different annotators strengthens the dataset’s generalizability, ensuring its effectiveness across diverse real-world applications. By incorporating multiple viewpoints, FIOVA provides a robust groundtruth (GT) that better aligns with human cognition and facilitates a more comprehensive evaluation of LVLMs.

\paragraph{Quality Control and Standardization}

While diversity enriches the dataset, it also introduces potential variability in annotation styles. To mitigate inconsistencies and maintain reliability, a multi-stage quality control process was implemented. Annotators adhered to a unified annotation framework, ensuring that descriptions remained factually accurate and neutral. After annotation, a systematic review process was conducted to verify alignment with video content and correct any inconsistencies.

To further enhance reliability, annotations from five annotators were synthesized into a cohesive groundtruth (GT) using GPT, balancing consistency while retaining diverse perspectives. This integration process ensured that the final GT captured the most critical video elements while preserving linguistic and semantic richness.

Through these measures, FIOVA establishes a high-quality, diverse, and well-structured dataset, serving as a benchmark for evaluating LVLMs in long-video captioning tasks. By addressing both annotation diversity and standardization, it ensures a fair and rigorous assessment of model performance in complex video understanding.

\clearpage

\subsection{Word Cloud for Five People Annotation and LVLMs Caption}

\begin{figure}[ht!]
\centering
\includegraphics[width=\textwidth]{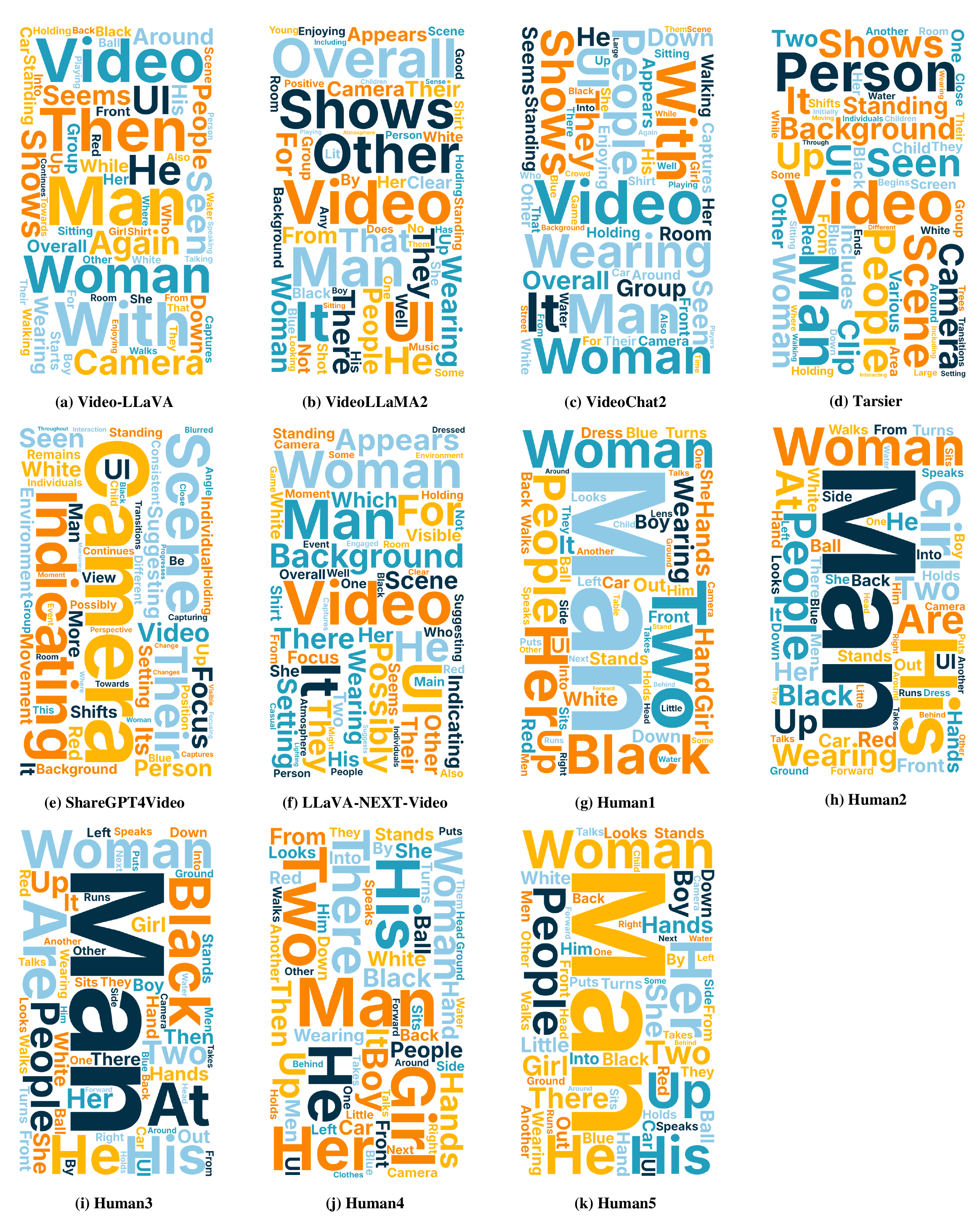}
\caption{Word cloud for five people annotation and LVLMs caption.}
\label{fig:word-cloud}
\end{figure}

\clearpage

\subsection{Distribution of Disagreement Among Human Annotators (Based on Multiple Dimensions)}
\label{subsec:8groups}

\begin{figure}[ht!]
\centering
\includegraphics[width=\textwidth]{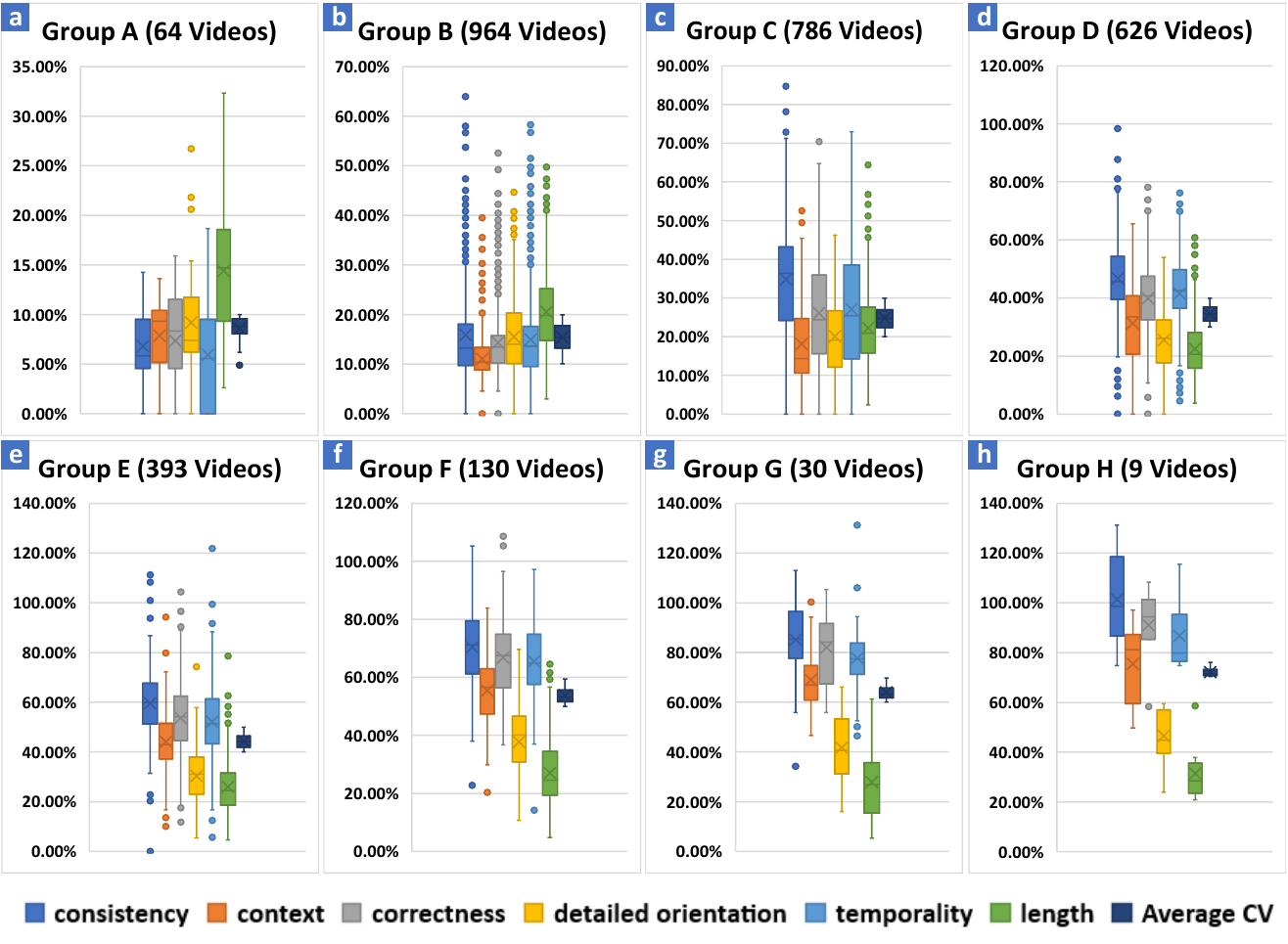}
\caption{The distribution of the multi-dimensional coefficient of variation for 8 groups. Please refer to Sec.~\ref{subsec:accessing-annotation-diversity} for more details. The dataset is divided based on the coefficient of variation (CV) of human annotators across multiple dimensions, resulting in 8 groups. Each group represents a different degree of disagreement among the 5 annotators, ranging from the smallest (Group A) to the largest (Group H).}
\label{fig:8groups}
\end{figure}

\begin{figure}[ht!]
\centering
\includegraphics[width=\textwidth]{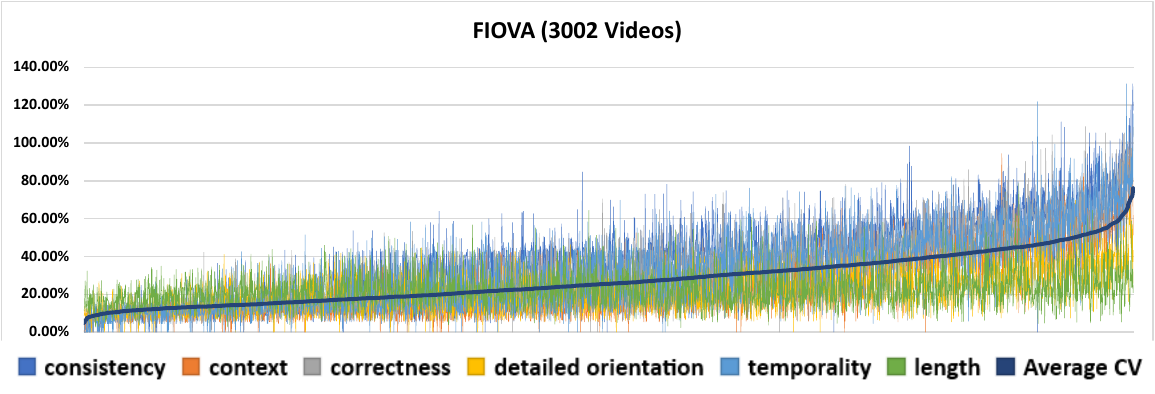}
\caption{Distribution of multiple dimensions coefficient of variation (CV) and their mean value on FIOVA.}
\label{fig:8groups-trend}
\end{figure}

\clearpage

\subsection{Examples for Various Groups}
\label{subsec:8groups-examples}

To illustrate the evaluation capabilities of GPT more intuitively, we present representative examples from different groups.

\subsubsection{High Consistency Group Sample (Group A)}

\begin{figure}[ht!]
\centering
\includegraphics[width=\textwidth]{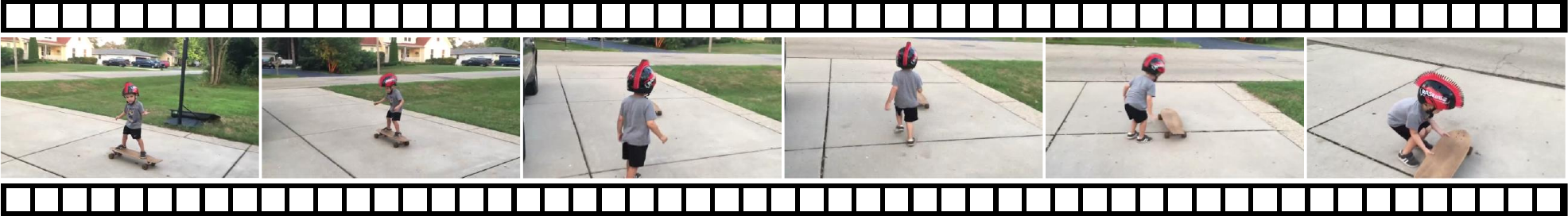}
\caption{An example of videos in Group A.}
\label{fig:example-A}
\end{figure}

The descriptions provided by annotators in Group A exhibit remarkable consistency regarding linguistic structure, semantic coverage, and contextual accuracy. The annotators consistently describe the core actions and key visual elements, with minimal phrase choice and sentence structure variations. The following are the annotators' descriptions:

\begin{itemize}
    \item \textbf{Human 1: }\textit{A little boy wearing a red hat stands on the scooter and slips forward. He waves his hand. He steps off the scooter. He looks at the scooter sliding forward and runs forward. The boy presses the scooter, then the scooter stops. The boy grabs the scooter with both hands.}
    \item \textbf{Human 2: }\textit{A little boy wearing a helmet is playing skateboard in the open air. He jumps down and pauses. He looks at the skateboard slipping away. He runs following the skateboard and squats down, and stops the skateboard with his hand. He picks up the skateboard and holds it in the arms.}
    \item \textbf{Human 3: }\textit{A child wearing a hat stands on a skateboard. There is a car next to the skateboard. The skateboard slides forward. The child goes down from the skateboard, and the skateboard continues sliding forward. The child catches up the skateboard and stops the skateboard by his hands. The child picks up the skateboard.}
    \item \textbf{Human 4: }\textit{A boy wears a skateboard helmet. The skateboard slides forward. The boy waves hands to the camera len. The boy goes down from the skateboard. The skateboard still moves forward. The boy chases the skateboard and stops it. The boy picks up the skateboard.}
    \item \textbf{Human 5: }\textit{A boy standing on a skateboard is doing skateboarding. The boy lands on one foot, while the other foot also takes off from the skateboard. The skateboard continues moving forward. The boy catches up the skateboard and stops it.}
\end{itemize}

GPT-based evaluation confirms this high agreement, with near-zero variation in contextual consistency (CV: 0.00\%) and overall context (CV: 0.00\%). Minor differences emerge in correctness (CV: 4.56\%) and temporality (CV: 4.56\%), while detail orientation (CV: 8.84\%) and description length (CV: 11.40\%) show slightly higher but still relatively low variability. The average CV across all dimensions is 4.89\%, reinforcing the strong consensus among annotators in describing these videos.

\subsubsection{High Variability Group Sample (Group H)}

\begin{figure}[ht!]
\centering
\includegraphics[width=\textwidth]{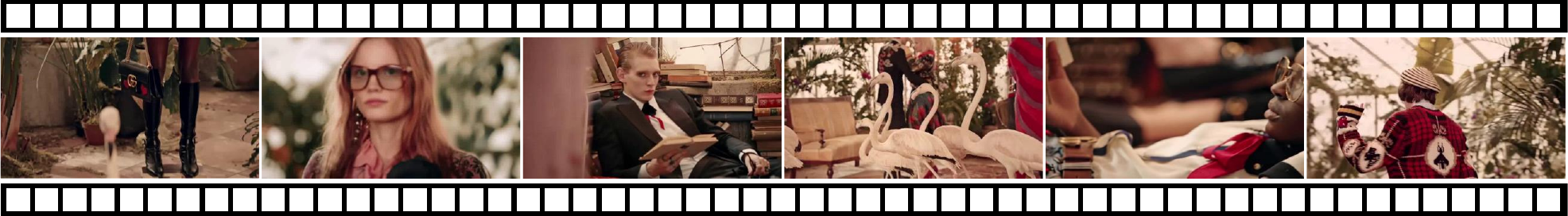}
\caption{An example of videos in Group H.}
\label{fig:example-H}
\end{figure}

Unlike Group A, the descriptions for Group H show substantial differences in wording, event focus, and narrative style. Annotators emphasize different aspects of the video, resulting in significant semantic divergence:

\begin{itemize}
    \item \textbf{Human 1: }\textit{A woman wearing a small glasses is reading books. A woman wearing a big glasses is looking forward. A man sitting beside a lot of books and holding a book looks at the front.  The woman wearing big glasses lies on the ground. A group of cranes walk by,  a man and a woman dancing behind. A woman in pink walks,  a man and a woman dancing behind. A black woman lies down and reads, a red dress woman sitting in a chair looks at the right. The woman with big glasses waves around the crane.  A man wearing glasses is reading. The pink dress woman is walking through, the man wearing glasses is reading, the black woman is lying on a black and white shirt and reading. A man wearing a hat  dances and walks through the black man upside down. A woman is lying next to a group of cranes. A woman steps on the book and walks. The woman in pink is dancing and walking through, a crane also comes.}
    \item \textbf{Human 2: }\textit{The lens sweeps a lady from top to bottom, and then there appears a woman with curly  hair. A man is wearing a suit, the man lying down is looking at her. Lens switch, the lady is lying on the floor, a group of white flamingos walk by, someone  next to them is dancing. A man and a woman push around, the first lady appears lying down and reading, the man in suit also wears glasses reading, the curly hair women and flamingos are dancing, someone next to them stretches his leg doing exercise.}
    \item \textbf{Human 3: }\textit{In a yard, a black-skinned woman is carrying a bag in the hands and reading a book, another long-haired woman is staring at the camera. A woman wearing a suit is lying on the stool, holding A book and looks at the lens, the long hair woman is lying on the carpet. A group of birds walk through the hall, a red dress man pushes a blonde woman away, the black skin woman next to him sitting to the side reads, another woman with black skin is lying down and reading. A woman wearing a red hat is sitting to the side, the long hair woman shakes hands, a woman in suit wears glasses, another woman wearing a striped shirt lies next to the carpet. The man in red keeps beating, A woman lying on the table raises her legs, the long hair woman is lying on the carpet, a pink dress woman is shaking the body and walking through.}
    \item \textbf{Human 4: }\textit{A woman standing next to some leaves. A woman is lying on the ground. Some geese are walking. A man and a woman are talking. A man is reading a book. A woman is sitting in a chair. A woman is waving her hands. A man is wearing glasses. Several people are lying on the ground. A man is leaning up and a man is walking by his side.}
    \item \textbf{Human 5: }\textit{A woman carrying a bag is standing and reading. A woman wearing glasses looks at the camera. A person holding a book looks at the woman. The woman wearing glasses is lying on the ground. Several people are dancing, a person is lying down and reading, a person is sitting on a chair. A man is waving his hands. The reading people wears the glasses. A man jumps forward and looks at another person who stands on the stool. The women with glasses is lying on the ground. A person steps on the book. Everyone does their own thing.}
\end{itemize}

GPT-based evaluation highlights notable variability across multiple dimensions. The highest inconsistency is observed in contextual consistency (CV: 98.54\%), followed by correctness (CV: 105.34\%), temporality (CV: 76.70\%), and context (CV: 49.79\%). Differences are also evident in detail orientation (CV: 53.93\%) and description length (CV: 37.87\%). The average CV across all dimensions is 70.36\%, indicating substantial variation in annotators' interpretations of the video content.

This variability suggests that some videos present inherent ambiguity or multiple valid interpretations, making them particularly challenging for standardized evaluation. The findings emphasize the importance of handling diverse linguistic styles and semantic perspectives when designing LVLM evaluation methodologies.

\clearpage

\subsection{Example for Correspondence between Videos, Human Descriptions, and Generated Groundtruth}
\label{subsec:human-gt-example}

\begin{figure}[t!]
\centering
\includegraphics[width=\textwidth]{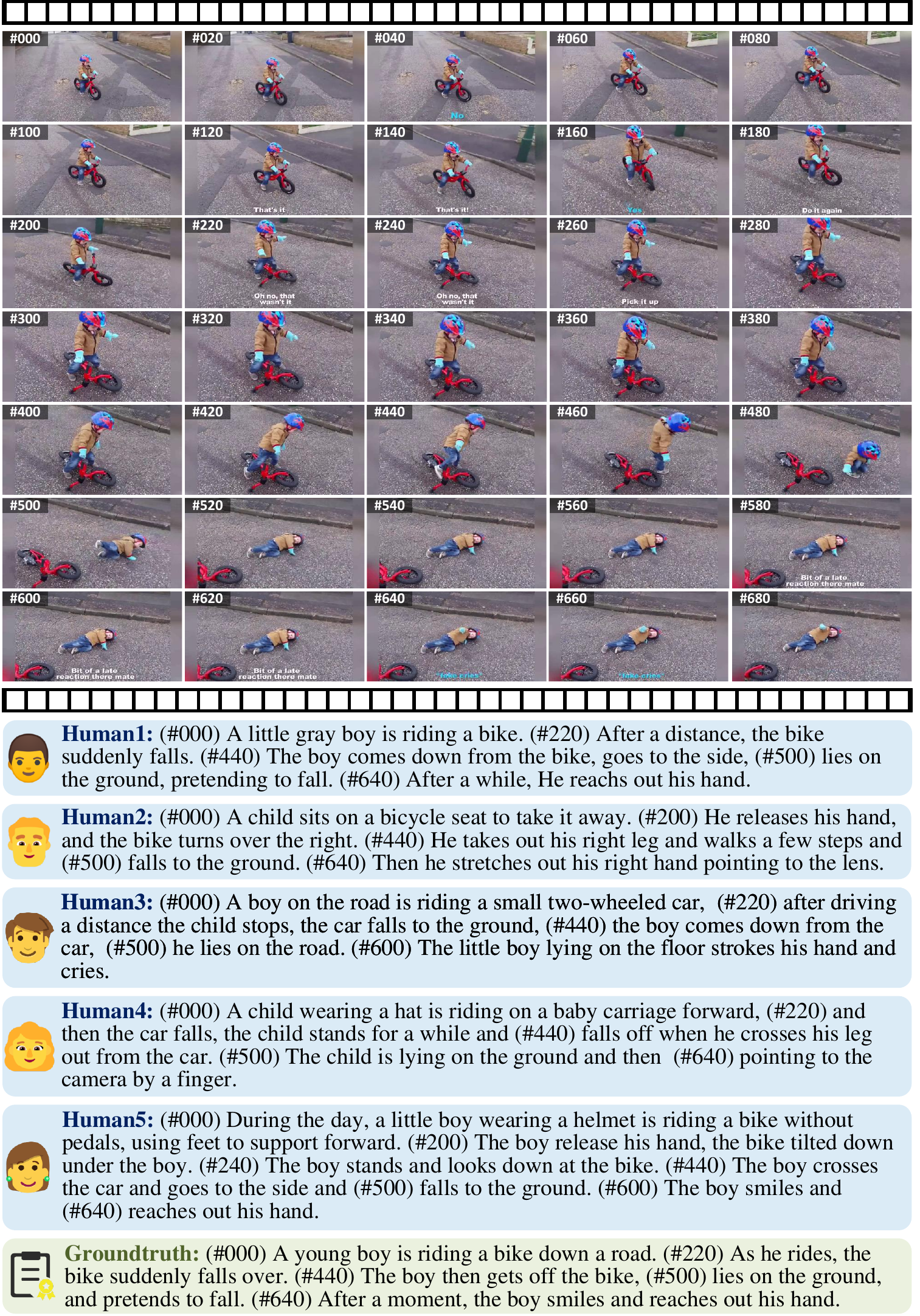}
\caption{A detailed example illustrating the correspondence between videos, human descriptions, and synthesized unified groundtruth.}
\label{fig:acc15}
\end{figure}

Fig.~\ref{fig:acc15} provides an illustrative case from the FIOVA dataset, demonstrating the multi-perspective annotation process and the synthesis of groundtruth (GT). The upper panel presents sampled frames at 20-frame intervals, highlighting key moments in the video. The lower panel displays individual annotations from five human annotators (Human1-Human5), followed by the final GT synthesized by GPT.

The video captures a sequence where a young boy rides a bicycle, stops, falls off, and ultimately lies on the ground while gesturing toward the camera. Although all annotators describe the core events consistently, notable variations exist in their focus and interpretation.

\paragraph{Diversity in Human Annotations}

While annotators agree on key moments—riding (\#000), stopping (\#200), falling off (\#440), lying on the ground (\#500), and pointing at the camera (\#640)—their descriptions vary in granularity and emphasis.

Some annotators highlight intentional actions, such as Human1 explicitly noting that the boy is pretending to fall. Others emphasize emotional expressions, with Human3 describing the boy as stroking his hand and crying, while Human5 instead describes him smiling after falling. Additionally, variations arise in the level of contextual details, such as Human5 mentioning that the boy rides without pedals before falling.

These differences reflect both objective event sequences and subjective interpretations, demonstrating the value of multi-annotator annotations in capturing the full spectrum of human perception.

\paragraph{Groundtruth Synthesis via GPT}

To construct a unified GT, GPT integrates multiple human perspectives while resolving discrepancies and ensuring coherence. Two key aspects are addressed:

\begin{itemize}
    \item \textbf{Resolving Annotation Discrepancies:} The notion of ``pretending to fall" appears in multiple annotations, justifying its inclusion in the GT. Besides, conflicting emotional expressions (crying vs. smiling) are reconciled by examining surrounding descriptions. Since multiple annotators describe playful behavior, the GT adopts "the boy smiles after falling" as the most representative interpretation.
    \item \textbf{Maintaining Event Completeness:} The GT ensures full coverage of key events while eliminating redundant or inconsistent descriptions. It preserves the critical transitions—riding, stopping, falling, lying down, and pointing—while balancing comprehensiveness and clarity.
\end{itemize}

\paragraph{Implications for LVLM Benchmarking}

This example underscores the importance of multi-perspective annotation synthesis in LVLM evaluation. The FIOVA dataset leverages diverse human descriptions to construct a high-quality, semantically rich groundtruth, enabling a more realistic and human-aligned evaluation of long-video comprehension. By integrating varied perspectives, FIOVA ensures a more robust and interpretable benchmark compared to single-annotator datasets, setting a new standard for evaluating LVLMs in complex video understanding tasks.

\clearpage

\clearpage

\section{Detailed Human Evaluation of GPT-Generated Groundtruth Quality}
\label{sec:human-gt-evaluation}

This section serves as a detailed supplement to Tab.~\ref{tab:human_eval_expanded} in the main paper, presenting the complete human evaluation results for GPT-fused captions compared with those from five individual annotators across 20 videos. All rankings were provided by 10 expert evaluators with backgrounds in computer vision. The videos were selected from the FIOVA dataset to cover 20 distinct themes and exhibit diverse semantics, narrative structures, and event density.

\subsection{Evaluation Setup and Procedure}

Each video was associated with six captions: five human-generated descriptions (Human1–Human5) and one GPT-synthesized groundtruth (GPT-generated GT) obtained by fusing the human annotations. Two complementary evaluation modes were adopted:

\begin{itemize}
\item \textbf{Mode A (Text-Only):} Evaluators ranked the captions by reading only the text without access to the video.
\item \textbf{Mode B (Video+Text):} Evaluators watched the video and then ranked the captions based on how well they aligned with the video content.
\end{itemize}

Each evaluator independently ranked the six captions for each video from 1 (best) to 6 (worst). The rankings were then averaged across annotators to produce final per-caption rankings for each video.

\subsection{Table Format Explanation}

\begin{table}[htbp]
\centering
\caption{Combined human evaluation results for 20 videos. Each cell shows average rank with breakdown: \textit{Mode A} (Text-Only), \underline{Mode B} (Video+Text). Lower is better.}
\label{tab:merged_modeA_B_full}
\renewcommand{\arraystretch}{1.1}
\setlength{\tabcolsep}{2pt}
\fontsize{9.2pt}{11pt}\selectfont
\resizebox{\textwidth}{!}{
\begin{tabular}{c c c c c c c c c}
\toprule
\textbf{No.} & \textbf{Group} & \textbf{CV (\%)} & \textbf{Human1} & \textbf{Human2} & \textbf{Human3} & \textbf{Human4} & \textbf{Human5} & \makecell{\textbf{GPT-generated} \\ \textbf{GT}} \\
\midrule
1  & A & 4.89  & 3.5 (\textit{3.6}/\underline{3.4}) & 4.3 (\textit{5.6}/\underline{3.0}) & 4.2 (\textit{3.2}/\underline{5.2}) & 4.5 (\textit{4.6}/\underline{4.4}) & 2.9 (\textit{2.6}/\underline{3.2}) & \textbf{1.6} (\textit{1.4}/\underline{1.8}) \\
2  & A & 9.55  & 3.8 (\textit{4.6}/\underline{3.0}) & 3.5 (\textit{3.2}/\underline{3.8}) & 4.9 (\textit{5.0}/\underline{4.8}) & 4.0 (\textit{4.0}/\underline{4.0}) & 3.1 (\textit{2.4}/\underline{3.8}) & \textbf{1.7} (\textit{1.8}/\underline{1.6}) \\
3  & B & 14.00 & 3.3 (\textit{3.8}/\underline{2.8}) & 4.5 (\textit{4.4}/\underline{4.6}) & 3.7 (\textit{4.0}/\underline{3.4}) & 4.3 (\textit{4.2}/\underline{4.4}) & 3.3 (\textit{3.0}/\underline{3.6}) & \textbf{2.1} (\textit{1.6}/\underline{2.6}) \\
4  & B & 16.28 & 3.1 (\textit{3.2}/\underline{3.0}) & 4.5 (\textit{4.0}/\underline{5.0}) & 4.9 (\textit{4.6}/\underline{5.2}) & 3.5 (\textit{4.8}/\underline{2.2}) & 2.8 (\textit{1.8}/\underline{3.8}) & \textbf{2.2} (\textit{2.6}/\underline{1.8}) \\
5  & B & 18.83 & 4.8 (\textit{4.0}/\underline{5.6}) & 4.4 (\textit{4.8}/\underline{4.0}) & 4.7 (\textit{5.4}/\underline{4.0}) & 3.1 (\textit{2.6}/\underline{3.6}) & 2.3 (\textit{2.4}/\underline{2.2}) & \textbf{1.7} (\textit{1.8}/\underline{1.6}) \\
6  & C & 23.30 & 3.4 (\textit{4.2}/\underline{2.6}) & 4.9 (\textit{5.4}/\underline{4.4}) & 3.7 (\textit{3.4}/\underline{4.0}) & 4.0 (\textit{3.0}/\underline{5.0}) & 2.1 (\textit{1.8}/\underline{2.4}) & \textbf{2.9} (\textit{3.2}/\underline{2.6}) \\
7  & C & 24.99 & 3.7 (\textit{3.4}/\underline{4.0}) & 3.0 (\textit{2.8}/\underline{3.2}) & 3.6 (\textit{4.0}/\underline{3.2}) & 3.5 (\textit{4.0}/\underline{3.0}) & 3.8 (\textit{3.8}/\underline{3.8}) & \textbf{3.4} (\textit{3.0}/\underline{3.8}) \\
8  & C & 28.97 & 3.5 (\textit{3.2}/\underline{3.8}) & 5.2 (\textit{5.4}/\underline{5.0}) & 4.5 (\textit{4.2}/\underline{4.8}) & 3.6 (\textit{3.8}/\underline{3.4}) & 2.3 (\textit{2.4}/\underline{2.2}) & \textbf{1.9} (\textit{2.0}/\underline{1.8}) \\
9  & D & 32.35 & 4.9 (\textit{4.6}/\underline{5.2}) & 4.7 (\textit{4.8}/\underline{4.6}) & 3.3 (\textit{3.6}/\underline{3.0}) & 3.8 (\textit{3.6}/\underline{4.0}) & 2.4 (\textit{2.4}/\underline{2.4}) & \textbf{1.9} (\textit{2.0}/\underline{1.8}) \\
10 & D & 35.18 & 3.7 (\textit{4.0}/\underline{3.4}) & 4.5 (\textit{4.2}/\underline{4.8}) & 3.8 (\textit{4.0}/\underline{3.6}) & 3.5 (\textit{3.8}/\underline{3.2}) & 3.5 (\textit{3.6}/\underline{3.4}) & \textbf{2.0} (\textit{1.4}/\underline{2.6}) \\
11 & D & 38.54 & 3.8 (\textit{4.0}/\underline{3.6}) & 3.7 (\textit{4.2}/\underline{3.2}) & 3.7 (\textit{3.4}/\underline{4.0}) & 3.3 (\textit{3.2}/\underline{3.4}) & 3.5 (\textit{3.8}/\underline{3.2}) & \textbf{3.0} (\textit{3.0}/\underline{3.0}) \\
12 & E & 42.19 & 4.6 (\textit{4.4}/\underline{4.8}) & 3.0 (\textit{4.0}/\underline{2.0}) & 4.1 (\textit{3.8}/\underline{4.4}) & 3.6 (\textit{3.4}/\underline{3.8}) & 3.5 (\textit{3.4}/\underline{3.6}) & \textbf{2.2} (\textit{1.4}/\underline{3.0}) \\
13 & E & 44.84 & 3.6 (\textit{3.2}/\underline{4.0}) & 4.8 (\textit{5.0}/\underline{4.6}) & 3.5 (\textit{4.2}/\underline{2.8}) & 4.8 (\textit{4.8}/\underline{4.8}) & 2.3 (\textit{2.2}/\underline{2.4}) & \textbf{2.0} (\textit{1.6}/\underline{2.4}) \\
14 & E & 47.07 & 3.8 (\textit{4.0}/\underline{3.6}) & 4.5 (\textit{4.8}/\underline{4.2}) & 5.0 (\textit{4.8}/\underline{5.2}) & 2.9 (\textit{2.4}/\underline{3.4}) & 2.7 (\textit{2.6}/\underline{2.8}) & \textbf{2.1} (\textit{2.4}/\underline{1.8}) \\
15 & F & 52.65 & 3.1 (\textit{3.0}/\underline{3.2}) & 4.1 (\textit{3.8}/\underline{4.4}) & 5.4 (\textit{5.4}/\underline{5.4}) & 4.5 (\textit{4.8}/\underline{4.2}) & 2.1 (\textit{2.0}/\underline{2.2}) & \textbf{1.8} (\textit{2.0}/\underline{1.6}) \\
16 & F & 58.23 & 4.0 (\textit{3.8}/\underline{4.2}) & 4.3 (\textit{4.2}/\underline{4.4}) & 3.8 (\textit{2.8}/\underline{4.8}) & 4.0 (\textit{4.8}/\underline{3.2}) & 2.9 (\textit{3.2}/\underline{2.6}) & \textbf{2.0} (\textit{1.6}/\underline{2.4}) \\
17 & G & 62.60 & 3.2 (\textit{4.0}/\underline{2.4}) & 4.0 (\textit{4.2}/\underline{3.8}) & 5.6 (\textit{5.2}/\underline{6.0}) & 3.5 (\textit{2.8}/\underline{4.2}) & 2.9 (\textit{2.8}/\underline{3.0}) & \textbf{1.8} (\textit{2.0}/\underline{1.6}) \\
18 & G & 68.58 & 3.1 (\textit{3.0}/\underline{3.2}) & 5.2 (\textit{5.0}/\underline{5.4}) & 4.3 (\textit{4.0}/\underline{4.6}) & 4.4 (\textit{4.6}/\underline{4.2}) & 2.5 (\textit{2.8}/\underline{2.2}) & \textbf{1.5} (\textit{1.6}/\underline{1.4}) \\
19 & H & 70.36 & 2.6 (\textit{2.2}/\underline{3.0}) & 3.1 (\textit{3.2}/\underline{3.0}) & 2.6 (\textit{2.8}/\underline{2.4}) & 5.5 (\textit{5.4}/\underline{5.6}) & 4.1 (\textit{4.6}/\underline{3.6}) & \textbf{3.1} (\textit{3.0}/\underline{3.2}) \\
20 & H & 76.09 & 3.2 (\textit{3.2}/\underline{3.2}) & 4.6 (\textit{4.8}/\underline{4.4}) & 5.1 (\textit{5.2}/\underline{5.0}) & 3.8 (\textit{3.4}/\underline{4.2}) & 2.6 (\textit{2.4}/\underline{2.8}) & \textbf{1.7} (\textit{1.6}/\underline{1.8}) \\
\midrule
\textbf{Avg.} & – & – & 
\makecell{3.64\\(\textit{3.65}/\underline{3.62})} & 
\makecell{4.24\\(\textit{4.33}/\underline{4.15})} & 
\makecell{4.22\\(\textit{4.25}/\underline{4.19})} & 
\makecell{3.91\\(\textit{3.90}/\underline{3.91})} & 
\makecell{2.88\\(\textit{2.81}/\underline{2.95})} & 
\makecell{\textbf{2.13}\\(\textit{2.06}/\underline{2.20})} \\
\bottomrule
\end{tabular}
}
\end{table}

Tab.~\ref{tab:merged_modeA_B_full} reports the average rankings for each caption under both evaluation modes. Rankings are presented in the format “overall average (\textit{Mode A}/\underline{Mode B})”. For example, 3.5 (\textit{3.6}/\underline{3.4}) indicates a Mode A ranking of 3.6, a Mode B ranking of 3.4, and an overall average of 3.5.

At the bottom of the table, we include an average row showing the overall trend across all 20 videos. The GPT-fused captions achieved average rankings of \textbf{2.06} in Mode A and \textbf{2.20} in Mode B, with a combined average of \textbf{2.13}, significantly outperforming all individual annotators (whose averages range from 2.88 to 4.24).

\subsection{Key Observations}

The table reveals that GPT-fused captions consistently performed among the top two across most videos under both evaluation modes. In contrast, rankings for individual human annotators showed considerable variation, particularly under Mode B. These findings support the conclusion that the fusion approach not only preserves linguistic fluency but also improves coverage of critical events and semantic consistency across different perspectives.

Moreover, the difference between Mode A and Mode B results illustrates the added value of video-conditioned evaluation. While text-only assessments may capture surface-level language quality, incorporating visual context enables a more comprehensive evaluation of alignment and relevance—especially crucial for multimodal tasks.

\subsection{Summary of Evaluator Feedback}

At the end of the experiment, we collected qualitative feedback from evaluators regarding the criteria they used to rank the captions. The following six dimensions were commonly emphasized:

\begin{itemize}
\item \textbf{Detail and Accuracy:} Captions should faithfully describe core events, character actions, and scene elements without factual errors.
\item \textbf{Completeness and Coverage:} Descriptions that capture full event sequences or multiple actions are preferred over oversimplified summaries.
\item \textbf{Coherence and Language Quality:} Grammatically correct, logically structured, and clearly phrased captions are easier to follow.
\item \textbf{Objectivity and Logical Consistency:} Captions should avoid subjective interpretation and remain factually and logically consistent.
\item \textbf{Video-Text Alignment:} Especially important in Mode B, captions should accurately reflect what is shown in the video.
\item \textbf{Conciseness and Non-Redundancy:} Evaluators favored captions that are informative yet succinct, avoiding repetition.
\end{itemize}

This 20-video, two-mode evaluation provides additional empirical support for the main claims regarding the superiority of GPT-fused captions. Compared to individual annotations, the fusion approach delivers more consistent, semantically rich, and human-aligned descriptions. 

\clearpage

\section{Calculation Process of Coefficient of Variation (CV)}
\label{sec:calculation-cv}

\begin{figure}[ht!]
\centering
\includegraphics[width=\textwidth]{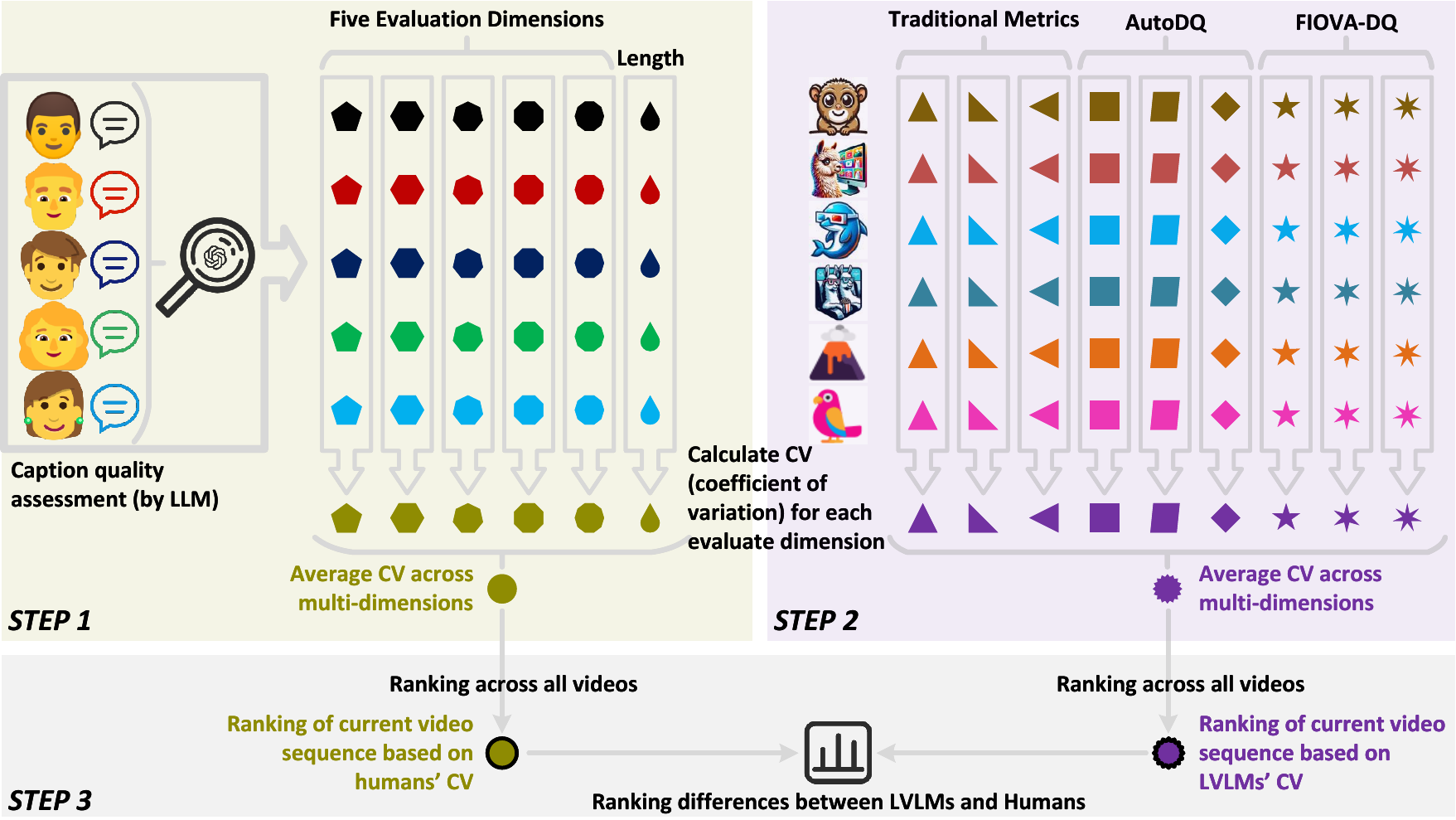}
\caption{Batch ranking for multi-dimensional consistency and human-machine comparison.}
\label{fig:batch-ranking}
\end{figure}

To systematically assess the consistency of video descriptions between human annotators and LVLMs, we adopt a structured evaluation framework based on coefficient of variation (CV), as illustrated in Fig.~\ref{fig:batch-ranking}. This process consists of three main stages:

\begin{itemize}
    \item \textbf{Step 1. Human Caption Consistency Evaluation (see Alg.~\ref{alg:cv_human}):} First, we measure the variability in human descriptions by evaluating five independent annotations for each video across six dimensions: Consistency, Context, Correctness, Detail Orientation, Temporality, and Length. Using an LLM-based analysis, we compute the CV for each dimension, where lower values indicate higher agreement among annotators. The average CV across all dimensions serves as a global consistency score, which is then used to categorize videos into different complexity levels (Groups A-H).

    \item \textbf{Step 2. LVLM Consistency Evaluation (see Alg.~\ref{alg:cv_lvlm}):} Next, we assess LVLM-generated captions following a similar methodology. Model-generated descriptions are evaluated using traditional lexical metrics (\eg, BLEU, GLEU, METEOR), event-level semantic consistency metrics (AutoDQ), and the proposed FIOVA-DQ metric, which integrates human cognitive weighting. The CV is calculated for each metric across all six LVLMs to quantify their variability. The resulting consistency score reflects how stable each model's descriptions are across different videos.

    \item \textbf{Step 3. Human-Machine Comparison (see Alg.~\ref{alg:cv_ranking_diff}):} Finally, to compare human and LVLM consistency, we rank videos based on their CV scores for both groups. The ranking difference quantifies the extent of alignment between human and LVLM-generated descriptions, offering a direct measure of their similarity in descriptive strategies. This analysis reveals key discrepancies, particularly in complex scenarios where LVLMs tend to exhibit greater consistency than humans due to their reliance on fixed generation patterns.
\end{itemize}

By integrating these evaluation stages, our framework enables a fine-grained comparison between human and machine-generated descriptions, highlighting the trade-offs between model stability and human-like variability. Notably, the incorporation of FIOVA-DQ allows for a human-centric evaluation paradigm, capturing nuances that traditional CV-based assessments fail to address. This methodological advancement provides a more interpretable, context-aware benchmark for evaluating LVLM performance in long-video description tasks.

\begin{algorithm}[htbp!]
    \caption{Framework for CV calculation between humans}
    \label{alg:cv_human}
    \SetKwInput{KwInput}{Input}                % Set the Input
    \SetKwInput{KwOutput}{Output}              % Set the Output
    \DontPrintSemicolon
    \KwInput{
    \( D = \{(V_1, C_{1}), \ldots, (V_n, C_{n})\} \): FIOVA dataset;\\
    $C_{i} = \{c_{i1}, c_{i2}, c_{i3}, c_{i4}, c_{i5}\}$: human annotations for video $V_i$;\\
    $E = \{\text{Consistency}, \text{Context}, \text{Correctness}, \text{Detail Orientation}, \text{Temporality}, \text{Length}\}$: evaluation dimensions;
    }
    \KwOutput{
    $CV^{human}_{dimension}$: Dictionary of coefficient of variation between humans for each evaluation dimension;\\
    $CV^{human}_{video}$: Dictionary of mean coefficient of variation between humans for each video;\\
    $Intervals$: Dictionary of intervals dividing $CV^{human}_{video}$
    }
    
    \tcc{Step 1: Calculate CV for each dimension}
    Initialize $CV^{human}_{dimension} \leftarrow \{\}$ \tcp*{Dictionary to store CV for each dimension}
    \For{$d \leftarrow 1$ \KwTo $\lvert E \rvert$}
    {
        Initialize $CV_{E[d]} \leftarrow \{\}$ \tcp*{Dictionary to store CV for each video in dimension $E[d]$}
        \For{$i \leftarrow 1$ \KwTo $\lvert D \rvert$}
        {
            Initialize scores list $S_i \leftarrow []$\;\\
            \For{$j \leftarrow 1$ \KwTo $\lvert C_i \rvert$}
            {
                $s_{ij} \leftarrow$ score of $c_{ij}$ in $E[d]$\;\\
                Append $s_{ij}$ to $S_i$\;
            }
            Calculate mean $\mu_i$ of $S_i$\;\\
            Calculate standard deviation $\sigma_i$ of $S_i$\;\\
            Calculate coefficient of variation $cv_i \leftarrow \frac{\sigma_i}{\mu_i}$\;\\
            $CV_{E[d]}[i] \leftarrow cv_i$\; \tcp*{Store CV for video $V_i$}
        }
        $CV^{human}_{dimension}[E[d]] \leftarrow CV_{E[d]}$\;
    }
    
    \tcc{Step 2: Calculate mean CV for each video}
    Initialize $CV^{human}_{video} \leftarrow \{\}$ \tcp*{Dictionary to store mean CV for each video}
    \For{$i \leftarrow 1$ \KwTo $\lvert D \rvert$}
    {
        Initialize sum of CVs $sum_{CV} \leftarrow 0$\;\\
        \For{$d \leftarrow 1$ \KwTo $\lvert E \rvert$}
        {
            $sum_{CV} \leftarrow sum_{CV} + CV^{human}_{dimension}[E[d]][i]$\;
        }
        Calculate mean $mean_{CV} \leftarrow \frac{sum_{CV}}{\lvert E \rvert}$\;\\
        $CV^{human}_{video}[i] \leftarrow mean_{CV}$\; \tcp*{Store mean CV for video $V_i$}
    }

    \tcc{Step 3: Divide $CV^{human}_{video}$ into intervals based on the maximum value}
    Sort $CV^{human}_{video}$ in ascending order by value and store sorted keys as $sorted\_keys$\;\\
    Calculate $max\_CV \leftarrow \max(CV^{human}_{video}.values())$\;\\
    Calculate number of intervals $N \leftarrow \lceil max\_CV \times 10 \rceil$ \tcp*{Each interval represents 10\%}
    Initialize $Intervals \leftarrow \{\}$ \tcp*{Dictionary to store interval information for each video}
    \For{$i \leftarrow 1$ \KwTo $\lvert sorted\_keys \rvert$}
    {
        $video\_id \leftarrow sorted\_keys[i]$\;\\
        $cv \leftarrow CV^{human}_{video}[video\_id]$\;\\
        Calculate interval index $index \leftarrow \lfloor cv \times 10 \rfloor$\;\\
        \If{$index \geq N$}
        {
            $index \leftarrow N - 1$\;
        }
        $Intervals[video\_id] \leftarrow index$\; \tcp*{Store interval for video $V_i$}
    }

    \Return $CV^{human}_{dimension}$, $CV^{human}_{video}$, $Intervals$
\end{algorithm}

\begin{algorithm}[t!]
    \caption{Framework for CV calculation between LVLMs}
    \label{alg:cv_lvlm}
    \SetKwInput{KwInput}{Input}                % Set the Input
    \SetKwInput{KwOutput}{Output}              % Set the Output
    \DontPrintSemicolon
    \KwInput{
    \( D = \{(V_1, R_{1}), \ldots, (V_m, R_{m})\} \): FIOVA dataset;\\
    $R_{i} = \{r_{i1}, r_{i2}, \ldots, r_{in}\}$: LVLMs' responses for video $V_i$;\\
    $E = \{\text{F1}, \text{Recall}, \text{Precision}, \text{BLEU}, \text{METEOR}, \text{GLEU}\}$: evaluation dimensions;
    }
    \KwOutput{
    $CV^{lvlm}_{dimension}$: Dictionary of coefficient of variation for each evaluation dimension;\\
    $CV^{lvlm}_{video}$: Dictionary of mean coefficient of variation between LVLMs for each video;
    }
    
    \tcc{Step 1: Calculate CV for each dimension}
    Initialize $CV^{lvlm}_{dimension} \leftarrow \{\}$ \tcp*{Dictionary to store CV for each dimension}
    \For{$d \leftarrow 1$ \KwTo $\lvert E \rvert$}
    {
        Initialize $CV_{E[d]} \leftarrow \{\}$ \tcp*{Dictionary to store CV for each video in dimension $E[d]$}
        \For{$i \leftarrow 1$ \KwTo $\lvert D \rvert$}
        {
            Initialize scores list $S_i \leftarrow []$\;\\
            \For{$j \leftarrow 1$ \KwTo $\lvert R_i \rvert$}
            {
                $s_{ij} \leftarrow$ score of $r_{ij}$ in $E[d]$\;\\
                Append $s_{ij}$ to $S_i$\;
            }
            Calculate mean $\mu_i$ of $S_i$\;\\
            Calculate standard deviation $\sigma_i$ of $S_i$\;\\
            Calculate coefficient of variation $cv_i \leftarrow \frac{\sigma_i}{\mu_i}$\;\\
            $CV_{E[d]}[i] \leftarrow cv_i$\; \tcp*{Store CV for video $V_i$}
        }
        $CV^{lvlm}_{dimension}[E[d]] \leftarrow CV_{E[d]}$\;
    }
    
    \tcc{Step 2: Calculate mean CV for each video}
    Initialize $CV^{lvlm}_{video} \leftarrow \{\}$ \tcp*{Dictionary to store mean CV for each video}
    \For{$i \leftarrow 1$ \KwTo $\lvert D \rvert$}
    {
        Initialize sum of CVs $sum_{CV} \leftarrow 0$\;\\
        \For{$d \leftarrow 1$ \KwTo $\lvert E \rvert$}
        {
            $sum_{CV} \leftarrow sum_{CV} + CV^{lvlm}_{dimension}[E[d]][i]$\;
        }
        Calculate mean $mean_{CV} \leftarrow \frac{sum_{CV}}{\lvert E \rvert}$\;\\
        $CV^{lvlm}_{video}[i] \leftarrow mean_{CV}$\; \tcp*{Store mean CV for video $V_i$}
    }

    \Return $CV^{lvlm}_{dimension}$, $CV^{lvlm}_{video}$
\end{algorithm}

\begin{algorithm}[htbp!]
    \caption{Calculate ranking differences}
    \label{alg:cv_ranking_diff}
    \SetKwInput{KwInput}{Input}                % Set the Input
    \SetKwInput{KwOutput}{Output}              % Set the Output
    \DontPrintSemicolon
    \KwInput{
    $CV^{lvlm}_{video}$: Dictionary of mean coefficient of variation between LVLMs for each video;\\
    $CV^{human}_{video}$: Dictionary of mean coefficient of variation between humans for each video;
    }
    \KwOutput{
    $Rankings^{human}$: Dictionary of rankings based on humans' CV;\\
    $Rankings^{lvlm}$: Dictionary of rankings based on LVLMs' CV;\\
    $Rankings^{diff}$: Dictionary of difference between $Rankings^{human}$ and $Rankings^{lvlm}$;
    }
    
    \tcc{Step 1: Rank videos based on $CV^{human}_{video}$ and $CV^{lvlm}_{video}$}
    Sort $CV^{human}_{video}$ by value in ascending order and store the sorted video IDs as $sorted\_ids^{human}$ \tcp*{Ranking by CV values from smallest to largest}
    Sort $CV^{lvlm}_{video}$ by value in ascending order and store the sorted video IDs as $sorted\_ids^{lvlm}$ \tcp*{Ranking by CV values from smallest to largest}
    
    Initialize $Rankings^{human} \leftarrow \{\}$ \tcp*{Dictionary to store human rankings}
    Initialize $Rankings^{lvlm} \leftarrow \{\}$ \tcp*{Dictionary to store LVLM rankings}
    \For{$rank \leftarrow 1$ \KwTo $\lvert sorted\_ids^{human} \rvert$}
    {
        $video\_id \leftarrow sorted\_ids^{human}[rank]$\;\\
        $Rankings^{human}[video\_id] \leftarrow rank$\;
    }
    
    \For{$rank \leftarrow 1$ \KwTo $\lvert sorted\_ids^{lvlm} \rvert$}
    {
        $video\_id \leftarrow sorted\_ids^{lvlm}[rank]$\;\\
        $Rankings^{lvlm}[video\_id] \leftarrow rank$\;
    }

    \tcc{Step 2: Calculate difference between rankings}
    Initialize $Rankings^{diff} \leftarrow \{\}$ \tcp*{Dictionary to store ranking differences}
    \ForEach{$video\_id \in CV^{human}_{video}.keys()$}
    {
        $Rankings^{diff}[video\_id] \leftarrow |Rankings^{human}[video\_id] - Rankings^{lvlm}[video\_id]|$\;
    }

    \Return $Rankings^{human}$, $Rankings^{lvlm}$, $Rankings^{diff}$
\end{algorithm}

\clearpage

\section{Prompts}
\label{sec:prompts}

\subsection{GPT-aided Evaluation Prompts}

\subsubsection{Prompt for Evaluation of Human Annotations}
\label{subsubsec:prompt-human-annotation}

\begin{center}
\begin{tcolorbox}[colback=gray!00,%gray background
                  colframe=c2,% black frame colour
                %   width=17.2cm,% Use 8cm total width,
                  arc=1.5mm, auto outer arc,
                  breakable,
                  left=0.9mm, right=0.9mm,
                  boxrule=0.9pt,
                  title = {The Prompt for Consistency of Annotation (by GPT).}
                 ]
\textbf{Prompt}\\
You are an intelligent chatbot designed for evaluating the factual accuracy of generative outputs for video-based caption. Your task is to compare the provided text and determine if they are factually consistent. Here's how you can accomplish the task:\\
------\\
\#\#INSTRUCTIONS:\\
- Focus on the consistency of the text with the expected content or background. The text should correspond to the correct information and should not contain any contradictions or significant differences.\\
- The text must be consistent in the information it provides about the content.\\
- Consider synonyms or paraphrases as valid matches, but only if they maintain the consistency in the conveyed information.\\
- Evaluate the consistency of the text.\\
- DO NOT PROVIDE ANY OTHER OUTPUT TEXT OR EXPLANATION. Only provide a single evaluation score from 1 to 10. For example, your response should look like this: \{“score”: [score]\}.
\tcblower
\textbf{User:}\\
Please evaluate the following video caption:\\
Provided caption: \textbf{“\{Caption\}”}\\
DO NOT PROVIDE ANY OTHER OUTPUT TEXT OR EXPLANATION. Only provide a single evaluation score from 1 to 10. For example, your response should look like this: \{“score”: [score]\}.
\end{tcolorbox}
\end{center}

\begin{center}
\begin{tcolorbox}[colback=gray!00,%gray background
                  colframe=c2,% black frame colour
                %   width=17.2cm,% Use 8cm total width,
                  arc=1.5mm, auto outer arc,
                  breakable,
                  left=0.9mm, right=0.9mm,
                  boxrule=0.9pt,
                  title = {The Prompt for Context of Annotation (by GPT).}
                 ]
\textbf{Prompt}\\
You are an intelligent chatbot designed for evaluating the factual accuracy of generative outputs for video-based caption. Your task is to compare the provided text and determine if they are factually consistent. Here's how you can accomplish the task: \\
------ \\
\#\#INSTRUCTIONS:  \\
- Evaluate whether the text aligns with the overall context of the expected content or background. It should not provide information that is out of context or misaligned. \\
- The text must capture the main themes and sentiments relevant to the content. \\
- Consider synonyms or paraphrases as valid matches. \\
- Provide your evaluation of the contextual understanding of the text. \\
DO NOT PROVIDE ANY OTHER OUTPUT TEXT OR EXPLANATION. Only provide a single evaluation score from 1 to 10. For example, your response should look like this: \{“score”: [score]\}. 
\tcblower
\textbf{User:}\\
Please evaluate the following video caption: \\
Provided caption: \textbf{“\{Caption\}”} \\
DO NOT PROVIDE ANY OTHER OUTPUT TEXT OR EXPLANATION. Only provide a single evaluation score from 1 to 10. For example, your response should look like this: \{“score”: [score]\}.
\end{tcolorbox}
\end{center}

\begin{center}
\begin{tcolorbox}[colback=gray!00,%gray background
                  colframe=c2,% black frame colour
                %   width=17.2cm,% Use 8cm total width,
                  arc=1.5mm, auto outer arc,
                  breakable,
                  left=0.9mm, right=0.9mm,
                  boxrule=0.9pt,
                  title = {The Prompt for Correctness of Annotation (by GPT).}
                 ]
\textbf{Prompt}\\
You are an intelligent chatbot designed for evaluating the factual accuracy of generative outputs for video-based caption. Your task is to compare the provided text and determine if they are factually consistent. Here's how you can accomplish the task: \\
------ \\
\#\#INSTRUCTIONS:  \\
- Focus on the factual correctness of the text. The text should not contain any misinterpretations or misinformation. \\
- The text must be factually accurate and align with the expected content or context. \\
- Consider synonyms or paraphrases as valid matches. \\
- Evaluate the factual accuracy of the text. \\
DO NOT PROVIDE ANY OTHER OUTPUT TEXT OR EXPLANATION. Only provide a single evaluation score from 1 to 10. For example, your response should look like this: \{“score”: [score]\}. 
\tcblower
\textbf{User:}\\
Please evaluate the following video caption: \\
Provided caption: \textbf{“\{Caption\}”} \\
DO NOT PROVIDE ANY OTHER OUTPUT TEXT OR EXPLANATION. Only provide a single evaluation score from 1 to 10. For example, your response should look like this: \{“score”: [score]\}.
\end{tcolorbox}
\end{center}

\begin{center}
\begin{tcolorbox}[colback=gray!00,%gray background
                  colframe=c2,% black frame colour
                %   width=17.2cm,% Use 8cm total width,
                  arc=1.5mm, auto outer arc,
                  breakable,
                  left=0.9mm, right=0.9mm,
                  boxrule=0.9pt,
                  title = {The Prompt for Detailed Orientation of Annotation (by GPT).}
                 ]
\textbf{Prompt}\\
You are an intelligent chatbot designed for evaluating the factual accuracy of generative outputs for video-based caption. Your task is to compare the provided text and determine if they are factually consistent. Here's how you can accomplish the task: \\
------ \\
\#\#INSTRUCTIONS:  \\
- Check if the text covers all major points relevant to the content. The text should not leave out any key aspects. \\
- Evaluate whether the text includes specific details rather than just generic points. It should provide comprehensive information that is tied to specific elements of the content. \\
- Consider synonyms or paraphrases as valid matches. \\
- Provide a single evaluation score that reflects the level of detail orientation of the text, considering both completeness and specificity. \\
DO NOT PROVIDE ANY OTHER OUTPUT TEXT OR EXPLANATION. Only provide a single evaluation score from 1 to 10. For example, your response should look like this: \{“score”: [score]\}. 
\tcblower
\textbf{User:}\\
Please evaluate the following video caption: \\
Provided caption: \textbf{“\{Caption\}”} \\
DO NOT PROVIDE ANY OTHER OUTPUT TEXT OR EXPLANATION. Only provide a single evaluation score from 1 to 10. For example, your response should look like this: \{“score”: [score]\}.
\end{tcolorbox}
\end{center}

\begin{center}
\begin{tcolorbox}[colback=gray!00,%gray background
                  colframe=c2,% black frame colour
                %   width=17.2cm,% Use 8cm total width,
                  arc=1.5mm, auto outer arc,
                  breakable,
                  left=0.9mm, right=0.9mm,
                  boxrule=0.9pt,
                  title = {The Prompt for Temporality of Annotation (by GPT).}
                 ]
\textbf{Prompt}\\
You are an intelligent chatbot designed for evaluating the factual accuracy of generative outputs for video-based caption. Your task is to compare the provided text and determine if they are factually consistent. Here's how you can accomplish the task: \\
------ \\
\#\#INSTRUCTIONS:  \\
- Focus on the temporal consistency of the text. It should correctly reflect the sequence of events or details as they are presented. \\
- Consider synonyms or paraphrases as valid matches, but only if the temporal order is maintained. \\
- Evaluate the temporal accuracy of the text. \\
DO NOT PROVIDE ANY OTHER OUTPUT TEXT OR EXPLANATION. Only provide a single evaluation score from 1 to 10. For example, your response should look like this: \{“score”: [score]\}. 
\tcblower
\textbf{User:}\\
Please evaluate the following video caption: \\
Provided caption: \textbf{“\{Caption\}”} \\
DO NOT PROVIDE ANY OTHER OUTPUT TEXT OR EXPLANATION. Only provide a single evaluation score from 1 to 10. For example, your response should look like this: \{“score”: [score]\}.
\end{tcolorbox}
\end{center}

\begin{tcolorbox}[colback=gray!00,%gray background
                  colframe=black,% black frame colour
                %   width=17.2cm,% Use 8cm total width,
                  arc=1.5mm, auto outer arc,
                  breakable,
                  left=0.9mm, right=0.9mm,
                  boxrule=0.9pt, colbacktitle = black!65!black,
                  title = {An Example of Evaluation Consistency of Annotation (by GPT).}
                 ]
\textbf{Prompt:} You are an intelligent chatbot designed for evaluating the factual accuracy of generative outputs for video-based caption. Your task is to compare the provided text and determine if they are factually consistent. Here's how you can accomplish the task:\\
------\\
\#\#INSTRUCTIONS:\\
{\color{orange} - Focus on the consistency of the text with the expected content or background. The text should correspond to the correct information and should not contain any contradictions or significant differences.\\
 - The text must be consistent in the information it provides about the content.\\
 - Consider synonyms or paraphrases as valid matches, but only if they maintain the consistency in the conveyed information.\\
 - Evaluate the consistency of the text.\\}
- DO NOT PROVIDE ANY OTHER OUTPUT TEXT OR EXPLANATION. Only provide a single evaluation score from 1 to 10. For example, your response should look like this: \{“score”: [score]\}.
\tcblower
\textbf{User:}\\
Please evaluate the following video caption:\\
Provided caption: \textbf{“A silver car is running slowly on the road. A red motorcycle running beside crashed on a red car. The motorcycle is overturned. The following vehicles began to slow down. A man in a blue coat riding a motorcycle parked next to the fallen motorcycle. From the red car two people came out. The slow motion of the vedio re-played.”}\\
DO NOT PROVIDE ANY OTHER OUTPUT TEXT OR EXPLANATION. Only provide a single evaluation score from 1 to 10. For example, your response should look like this: \{“score”: [score]\}.\\
\textbf{GPT:} {\color{orange}\{“score”: 3\}}
\end{tcolorbox}

\subsubsection{Prompt for GroundTruth Generation}
\label{subsubsec:prompt-groundtruth-generation}

\begin{center}
\begin{tcolorbox}[colback=gray!00,%gray background
                  colframe=c2,% black frame colour
                %   width=17.2cm,% Use 8cm total width,
                  arc=1.5mm, auto outer arc,
                  breakable,
                  left=0.9mm, right=0.9mm,
                  boxrule=0.9pt,
                  title = {The Prompt for Groundtruth Generation (by GPT).}
                 ]
\textbf{Prompt}\\
Given five video descriptions. Combine the five video descriptions into a single, coherent description that captures the essence of the video clip. \\
Please generate the response in the form of a Python dictionary string with keys “gt”. The value of “gt” is a List(str), of which is groundtruth for this video description. 
\tcblower
\textbf{User:}\\
Video description 1: \textbf{“\{Caption1\}”} \\
Video description 2: \textbf{“\{Caption2\}”} \\
Video description 3: \textbf{“\{Caption3\}”} \\
Video description 4: \textbf{“\{Caption4\}”} \\
Video description 5: \textbf{“\{Caption5\}”} \\
DO NOT PROVIDE ANY OTHER OUTPUT TEXT OR EXPLANATION.
Only provide one Python dictionary string. For example, your response should look like this: \{“gt”: [gt]\}
\end{tcolorbox}
\end{center}

\begin{tcolorbox}[colback=gray!00,%gray background
                  colframe=black,% black frame colour
                %   width=17.2cm,% Use 8cm total width,
                  arc=1.5mm, auto outer arc,
                  breakable,
                  left=0.9mm, right=0.9mm,
                  boxrule=0.9pt, colbacktitle = black!65!black,
                  title = {An Example of Groundtruth Generation (by GPT).}
                 ]
\textbf{Prompt:} Given five video descriptions. Combine the five video descriptions into a single, coherent description that captures the essence of the video clip. \\
Please generate the response in the form of a Python dictionary string with keys “gt”. The value of “gt” is a List(str), of which is groundtruth for this video description.
\tcblower
\textbf{User:}\\
Video description 1: \textbf{“A silver car is running slowly on the road. A red motorcycle running beside crashed on a red car. The motorcycle is overturned. The following vehicles began to slow down. A man in a blue coat riding a motorcycle parked next to the fallen motorcycle. From the red car two people came out. The slow motion of the vedio re-played.”} \\
Video description 2: \textbf{“A silver business car on the side of the road is driving, followed by a motorcycle, and a red car in front of a commercial car changing lanes crashed on a motorcycle, and the man on the motorcycle fell to the ground. A white van in the following stopped. The motorcycle driver in red stood up and the following motorcycle stoped, the man in blue came down to help the blue men pick up the package. From the red car two men and one women waled to the  man in red. The crash process is re played back in slow motion.
”} \\
Video description 3: \textbf{“The vehicle on the road was running normally. A red car crashed into the motorcycle next to it when changing lanes. The motorcycle was knocked down and the two men on the motorcycle fell. A motorcycle that passed by stopped to inquire about the situation. The two men on the motorcycle stood up, and the red car stopped. From the car down two men and a woman came. The crash scene is replayed again.
”} \\
Video description 4: \textbf{“On the highway cars were driving normal. A motorcycle when driving hit a car which drove from the side lane to middle lane. The motorcycle driver fell in from of a van and several motocycles passed by. A motorcycle driver helped the man stand on the ground. A few people from the crashed car came over to check, and then slow down the crash video.
”} \\
Video description 5: \textbf{“In the daytime, a front silver hatchback in the right side of the road was running slowly. On the left side a motorcycle quickly passed, with two people. The motorcycle passed the silver car, which collided with a red car in front which was turning left, and the car was knocked to the ground and then climbed up. Rear vehicles stoped. From the rear three motorcycles came, one of which  stopped next to the one who was knocked over. From the red car a woman and three men came out, and went to the people who was hit. Slow motion played back the accident moment.
”} \\
DO NOT PROVIDE ANY OTHER OUTPUT TEXT OR EXPLANATION.
Only provide one Python dictionary string. For example, your response should look like this: \{“gt”: [gt]\}\\
\textbf{GPT:} {\color{orange}{
    \{ “gt”: [
        “A silver car and a red motorcycle are driving on the road. The red car in front of the silver car changes lanes and collides with the motorcycle, causing it to crash. The motorcycle driver falls to the ground. A white van and several motorcycles behind stop. A man in a blue coat riding a motorcycle stops to help the fallen motorcycle driver. Two men and a woman from the red car come out to check on the situation. The crash scene is then replayed in slow motion.”
    ] \}
}}
\end{tcolorbox}

\subsubsection{Prompt for Event Extraction}
\label{subsubsec:event-extraction}

\begin{center}
\begin{tcolorbox}[colback=gray!00,%gray background
                  colframe=c2,% black frame colour
                %   width=17.2cm,% Use 8cm total width,
                  arc=1.5mm, auto outer arc,
                  breakable,
                  left=0.9mm, right=0.9mm,
                  boxrule=0.9pt,
                  title = {The Prompt for Event Extraction (by GPT).}
                 ]
\textbf{Prompt}\\
Given a video description. Extract at most 10 key events from the video description paragraph. \\
Requirements: \\
- Every event is represented by a brief sentence within 10 words, with a subject, a predicate and optionally an object,avoid unnecessary appearance descriptions. \\
- Every event must be atomic, meaning that it can not be further split into multiple events. \\
- Scene cuts and camera motions are NOT events. \\
- Substitute pronouns by the nouns they refer to. \\
Please generate the response in the form of a Python dictionary string with keys “events”. The value of “events” is a List(str), of which each item is an event. 
\tcblower
\textbf{User:}\\
Video description: \textbf{“\{Caption\}”} \\
DO NOT PROVIDE ANY OTHER OUTPUT TEXT OR EXPLANATION. Only provide the Python dictionary string. For example, your response should look like this: \{“events”: [event1, event2,...]\}
\end{tcolorbox}
\end{center}

\begin{tcolorbox}[colback=gray!00,%gray background
                  colframe=black,% black frame colour
                %   width=17.2cm,% Use 8cm total width,
                  arc=1.5mm, auto outer arc,
                  breakable,
                  left=0.9mm, right=0.9mm,
                  boxrule=0.9pt, colbacktitle = black!65!black,
                  title = {An Example of Event Extraction (by GPT).}
                 ]
\textbf{Prompt:}\\
Given a video description. Extract at most 10 key events from the video description paragraph. \\
Requirements: \\
{\color{orange}{- Every event is represented by a brief sentence within 10 words, with a subject, a predicate and optionally an object,avoid unnecessary appearance descriptions. \\
- Every event must be atomic, meaning that it can not be further split into multiple events. \\
- Scene cuts and camera motions are NOT events. \\
- Substitute pronouns by the nouns they refer to. \\}}
Please generate the response in the form of a Python dictionary string with keys “events”. The value of “events” is a List(str), of which each item is an event. 
\tcblower
\textbf{User:}\\
Video description: \textbf{A silver car and a red motorcycle are driving on the road. The red car in front of the silver car changes lanes and collides with the motorcycle, causing it to crash. The motorcycle driver falls to the ground. A white van and several motorcycles behind stop. A man in a blue coat riding a motorcycle stops to help the fallen motorcycle driver. Two men and a woman from the red car come out to check on the situation. The crash scene is then replayed in slow motion.} \\
DO NOT PROVIDE ANY OTHER OUTPUT TEXT OR EXPLANATION. Only provide the Python dictionary string. For example, your response should look like this: \{“events”: [event1, event2,...]\} \\
\textbf{GPT:} {\color{orange}{
    \\\{
    “events”: [\\
        “Silver car and red motorcycle drive on road”, \\
        “Red car changes lanes and collides with motorcycle”, \\
        “Motorcycle driver falls to ground”, \\
        “White van and motorcycles behind stop”, \\
        “Man in blue coat riding motorcycle helps driver”, \\
        “Men and woman from red car check situation”, \\
        “Crash scene replayed in slow motion”
    ]\\
 \}
}}
\end{tcolorbox}

\subsubsection{Prompt for Cross Checking}

\begin{center}
\begin{tcolorbox}[colback=gray!00,%gray background
                  colframe=c2,% black frame colour
                %   width=17.2cm,% Use 8cm total width,
                  arc=1.5mm, auto outer arc,
                  breakable,
                  left=0.9mm, right=0.9mm,
                  boxrule=0.9pt,
                  title = {The Prompt for Cross Checking (by GPT).}
                 ]
\textbf{Prompt}\\
Given a video description and a list of events. For each event, classify the relationship between the video description and the event into three classes: entailment, neutral, contradiction. \\
- “entailment” means that the video description entails the event. \\
- “contradiction” means that some detail in the video description contradicts with the event. \\
- “neutral” means that the relationship is neither “entailment” or “contradiction”. \\
Output a list in Json format: [ \{“event”: “copy an event here”, “relationship”: “put class name here” \}, ... ]. 
\tcblower
\textbf{User:}\\
Video description: \textbf{“\{Caption\}”} \\
Events: \textbf{“\{Event\}”} \\
DO NOT PROVIDE ANY OTHER OUTPUT TEXT OR EXPLANATION. Only output the JSON.
Output:
\end{tcolorbox}
\end{center}

\begin{tcolorbox}[colback=gray!00,%gray background
                  colframe=black,% black frame colour
                %   width=17.2cm,% Use 8cm total width,
                  arc=1.5mm, auto outer arc,
                  breakable,
                  left=0.9mm, right=0.9mm,
                  boxrule=0.9pt, colbacktitle = black!65!black,
                  title = {An Example of Cross Checking (by GPT).}
                 ]
\textbf{Prompt:}\\
Given a video description and a list of events. For each event, classify the relationship between the video description and the event into three classes: entailment, neutral, contradiction. \\
{\color{orange}{- “entailment” means that the video description entails the event. \\
- “contradiction” means that some detail in the video description contradicts with the event. \\
- “neutral” means that the relationship is neither “entailment” or “contradiction”. \\}}
Output a list in Json format: [ \{“event”: “copy an event here”, “relationship”: “put class name here” \}, ... ]. 
\tcblower
\textbf{User:}\\
Video description: \textbf{The video clip shows a road scene in Singapore. It begins with a view of a wet road, possibly from rain. The scene then transitions to a road with moving vehicles, including cars and buses. A silver car is prominently featured in the center, moving slowly due to traffic. Motorcycles and other vehicles are seen passing by the car. On the right side of the road, there is a group of people who appear to be involved in an incident or accident. The background consists of trees and roadside barriers.} \\
Events: \textbf{\\\{
    “events”: [\\
        “Silver car and red motorcycle drive on road”, \\
        “Red car changes lanes and collides with motorcycle”, \\
        “Motorcycle driver falls to ground”, \\
        “White van and motorcycles behind stop”, \\
        “Man in blue coat riding motorcycle helps driver”, \\
        “Men and woman from red car check situation”, \\
        “Crash scene replayed in slow motion”
    ]\\
 \}} \\
DO NOT PROVIDE ANY OTHER OUTPUT TEXT OR EXPLANATION. Only output the JSON.
Output:\\
\textbf{GPT:} {\color{orange}{\\
    \{
        “event”: “Silver car and red motorcycle drive on road”,
        “relationship”: “entailment”
    \},\\
    \{
        “event”: “Red car changes lanes and collides with motorcycle”,
        “relationship”: “neutral”
    \},\\
    \{
        “event”: “Motorcycle driver falls to ground”,
        “relationship”: “entailment”
    \},\\
    \{
        “event”: “White van and motorcycles behind stop”,
        “relationship”: “neutral”
    \},\\
    \{
        “event”: “Man in blue coat riding motorcycle helps drive”,
        “relationship”: “contradiction”
    \},\\}
    \color{orange}{
    \{
        “event”: “Men and woman from red car check situation”,
        “relationship”: “entailment”
    \},\\
    \{
        “event”: “Crash scene replayed in slow motion”,
        “relationship”: “contradiction”
    \}
}}
\end{tcolorbox}

\clearpage

\subsection{Video Caption Prompts}
\label{subsec:lvlms-prompt}

To ensure consistent input and fair evaluation across all vision-language models (LVLMs), we follow the official prompt provided by each model’s repository. If an official prompt is unavailable, we adopt a standardized instruction:

\begin{center}
\begin{tcolorbox}[colback=gray!00,%gray background
                  colframe=c2,% black frame colour
                %   width=17.2cm,% Use 8cm total width,
                  arc=1.5mm, auto outer arc,
                  breakable,
                  left=0.9mm, right=0.9mm,
                  boxrule=0.9pt,
                  title = {Default Prompt.}
                 ]
Describe the video in details.
\end{tcolorbox}
\end{center}

This default prompt is applied to VideoLLaMA2, Video-LLaVA, ShareGPT4Video, Tarsier, VideoChat2, InternVL-2.5, Qwen2.5-VL, and GPT-4o. For LLaVA-NEXT-Video, we follow its official instruction as follows:

\begin{center}
\begin{tcolorbox}[colback=gray!00,%gray background
                  colframe=c2,% black frame colour
                %   width=17.2cm,% Use 8cm total width,
                  arc=1.5mm, auto outer arc,
                  breakable,
                  left=0.9mm, right=0.9mm,
                  boxrule=0.9pt,
                  title = {Prompt for LLaVA-NEXT-Video.}
                 ]
Please provide a detailed description of the video, focusing on the main subjects, their actions, and the background scenes.
\end{tcolorbox}
\end{center}

Tab.~\ref{tab:prompt-url-table} summarizes the key details of all models used in our study, including their developer, parameter size, and access method. This helps support transparent evaluation and reproducibility.

\begin{table}[h!]
    \centering
    \fontsize{8pt}{10pt}\selectfont
    \setlength{\tabcolsep}{3.5pt}
    \caption{Official resources and access methods for the LVLMs evaluated in this work.}
    \begin{tabular}{llll}
    \toprule
      \textbf{LVLMs} & \textbf{Main Creator} & \textbf{\#Parameters} & \textbf{Access} \\ \midrule
       \textbf{\href{https://github.com/DAMO-NLP-SG/VideoLLaMA2}{VideoLLaMA2}} \cite{videollama2} & Alibaba Group & 7B & Weights \\ 
       \textbf{\href{https://github.com/PKU-YuanGroup/Video-LLaVA}{Video-LLaVA}} \cite{videollava} & Peking University & 7B & Weights \\ 
       \textbf{\href{https://github.com/ShareGPT4Omni/ShareGPT4Video}{ShareGPT4Video}} \cite{sharegpt4video} & University of Science and Technology of China & 8B & Weights \\
       \textbf{\href{https://github.com/bytedance/tarsier}{Tarsier}} \cite{tarsier} & ByteDance & 7B & Weights \\ 
       \textbf{\href{https://github.com/OpenGVLab/Ask-Anything}{VideoChat2}} \cite{videochat2} & Shanghai AI Laboratory & 7B & Weights \\ 
       \textbf{\href{https://github.com/LLaVA-VL/LLaVA-NeXT}{LLaVA-NEXT-Video}} \cite{llavanextvideo} & ByteDance & 7B & Weights \\
       \textbf{\href{https://internvl.github.io/blog/2024-12-05-InternVL-2.5/}{InternVL-2.5}} \cite{InternVL} & OpenGVLab & 8B & Weights \\
       \textbf{\href{https://github.com/QwenLM/Qwen2.5-VL}{Qwen2.5-VL}} \cite{Qwen} & Alibaba Group & 7B & Weights \\
       \textbf{\href{https://openai.com/index/GPT-4/}{GPT-4o}} \cite{gpt4} & OpenAI & / & API \\
    \bottomrule
    \label{tab:prompt-url-table}
    \end{tabular}
\end{table}

Regarding frame selection, we adopt a unified setting of 8 input frames per video for all main experiments on the FIOVA benchmark. This practice, widely adopted in video understanding research~\cite{maaz2024videogpt+, wang2024emu3}, balances semantic coverage with computational efficiency and enhances the comparability and reproducibility of model results. All primary evaluation scores reported in FIOVA are based on this 8-frame setting.

To further investigate models’ temporal sensitivity under complex scenarios, we conduct a frame-length ablation study on the \textsc{FIOVA}$_\text{hard}$ subset. Specifically, we evaluate four representative models—Tarsier, InternVL-2.5, Qwen2.5-VL, and GPT-4o—using three different frame settings: 8, 16, and 32 input frames. Results are presented in Tab.~\ref{tab:frame-ablation}, which demonstrate how increasing visual input granularity affects model performance in high-semantic-load video tasks. Most models benefit from higher frame counts, particularly in Recall and F1, indicating that richer temporal context helps capture more complete event semantics.

\clearpage

\section{Detailed Experimental Results}
\label{sec:detailed-exp}

\subsection{LVLMs v.s. Humans on Traditional Metrics}
\label{subsec:results-lvlms-humans}

\begin{table}[h!]
\caption{Comparison of LVLMs and Humans on FIOVA based on traditional metrics (BLEU, METEOR, and GLEU). The background color represents the performance of the metric. The darker the green, the better the performance.}
% \vspace{-8pt}
    \label{tab:traditional-metric}
    \centering
    \fontsize{8pt}{10pt}\selectfont
        \setlength{\tabcolsep}{3pt}
\renewcommand{\arraystretch}{1.2}
\begin{tabular}{llllllll}
\toprule
\textbf{Metrics} & \textbf{LVLMs} & \textbf{Human1} & \textbf{Human2} & \textbf{Human3} & \textbf{Human4} & \textbf{Human5} & \multicolumn{1}{l}{\textbf{GT}} \\\midrule
 & \textbf{Tarsier} & \cellcolor[HTML]{AFDDBC}0.025 & \cellcolor[HTML]{ACDCBA}0.025 & \cellcolor[HTML]{B0DEBE}0.024 & \cellcolor[HTML]{B0DEBE}0.025 & \cellcolor[HTML]{B2DFC0}0.024 & \cellcolor[HTML]{63BE7B}0.043 \\
 & \textbf{VideoLLaMA2} & \cellcolor[HTML]{CBE8D5}0.018 & \cellcolor[HTML]{C6E7D1}0.019 & \cellcolor[HTML]{CAE8D4}0.018 & \cellcolor[HTML]{CCE9D5}0.018 & \cellcolor[HTML]{CCE9D5}0.018 & \cellcolor[HTML]{99D4AA}0.030 \\
 & \textbf{LLaVA-NEXT-Video} & \cellcolor[HTML]{E1F1E7}0.013 & \cellcolor[HTML]{DAEFE2}0.014 & \cellcolor[HTML]{DDF0E5}0.014 & \cellcolor[HTML]{DDF0E5}0.014 & \cellcolor[HTML]{E0F1E7}0.013 & \cellcolor[HTML]{C3E5CE}0.020 \\
 & \textbf{Video-LLaVA} & \cellcolor[HTML]{CEEAD8}0.017 & \cellcolor[HTML]{C9E8D3}0.019 & \cellcolor[HTML]{CAE8D4}0.018 & \cellcolor[HTML]{CBE8D5}0.018 & \cellcolor[HTML]{D0EBD9}0.017 & \cellcolor[HTML]{A6D9B5}0.027 \\
 & \textbf{ShareGPT4Video} & \cellcolor[HTML]{FCFCFF}0.006 & \cellcolor[HTML]{FAFCFD}0.007 & \cellcolor[HTML]{FBFCFE}0.006 & \cellcolor[HTML]{FBFCFE}0.006 & \cellcolor[HTML]{FCFCFF}0.006 & \cellcolor[HTML]{EAF5F0}0.010 \\
\multirow{-6}{*}{\textbf{BLEU ($\uparrow$)}} & \textbf{VideoChat2} & \cellcolor[HTML]{BDE3C9}0.021 & \cellcolor[HTML]{B3DFC0}0.024 & \cellcolor[HTML]{B8E1C4}0.023 & \cellcolor[HTML]{B9E1C6}0.022 & \cellcolor[HTML]{BCE2C8}0.022 & \cellcolor[HTML]{7EC992}0.037 \\\midrule

 & \textbf{Tarsier} & \cellcolor[HTML]{B9E1C5}0.232 & \cellcolor[HTML]{B9E1C5}0.232 & \cellcolor[HTML]{BEE3C9}0.229 & \cellcolor[HTML]{BCE2C8}0.230 & \cellcolor[HTML]{BAE2C6}0.231 & \cellcolor[HTML]{7FCA93}0.265 \\
 & \textbf{VideoLLaMA2} & \cellcolor[HTML]{A1D7B1}0.245 & \cellcolor[HTML]{9CD5AC}0.248 & \cellcolor[HTML]{A0D7AF}0.246 & \cellcolor[HTML]{9FD6AE}0.247 & \cellcolor[HTML]{9ED6AE}0.247 & \cellcolor[HTML]{7AC88F}0.268 \\
 & \textbf{LLaVA-NEXT-Video} & \cellcolor[HTML]{A0D7AF}0.246 & \cellcolor[HTML]{9AD5AB}0.249 & \cellcolor[HTML]{9CD5AC}0.248 & \cellcolor[HTML]{9BD5AC}0.249 & \cellcolor[HTML]{9FD7AF}0.247 & \cellcolor[HTML]{77C68C}0.270 \\
 & \textbf{Video-LLaVA} & \cellcolor[HTML]{AEDDBB}0.238 & \cellcolor[HTML]{A8DAB6}0.242 & \cellcolor[HTML]{AADBB9}0.240 & \cellcolor[HTML]{AADBB8}0.240 & \cellcolor[HTML]{ABDBB9}0.240 & \cellcolor[HTML]{8DCF9F}0.257 \\
 & \textbf{ShareGPT4Video} & \cellcolor[HTML]{F9FBFD}0.194 & \cellcolor[HTML]{F6FAFA}0.196 & \cellcolor[HTML]{F5F9F9}0.197 & \cellcolor[HTML]{F8FBFB}0.195 & \cellcolor[HTML]{FCFCFF}0.192 & \cellcolor[HTML]{D0EBD9}0.218 \\
\multirow{-6}{*}{\textbf{METEOR ($\uparrow$)}} & \textbf{VideoChat2} & \cellcolor[HTML]{8FD0A1}0.256 & \cellcolor[HTML]{88CD9B}0.260 & \cellcolor[HTML]{8DCF9F}0.257 & \cellcolor[HTML]{8BCE9E}0.258 & \cellcolor[HTML]{8CCF9E}0.258 & \cellcolor[HTML]{63BE7B}0.281 \\\midrule

 & \textbf{Tarsier} & \cellcolor[HTML]{92D1A3}0.091 & \cellcolor[HTML]{8FD0A1}0.092 & \cellcolor[HTML]{92D2A4}0.090 & \cellcolor[HTML]{92D1A4}0.091 & \cellcolor[HTML]{93D2A5}0.090 & \cellcolor[HTML]{63BE7B}0.119 \\
 & \textbf{VideoLLaMA2} & \cellcolor[HTML]{B7E0C4}0.068 & \cellcolor[HTML]{B3DFC0}0.071 & \cellcolor[HTML]{B4DFC1}0.070 & \cellcolor[HTML]{B5E0C2}0.069 & \cellcolor[HTML]{B7E0C3}0.068 & \cellcolor[HTML]{96D3A7}0.088 \\
 & \textbf{LLaVA-NEXT-Video} & \cellcolor[HTML]{D9EEE1}0.047 & \cellcolor[HTML]{D6EDDE}0.049 & \cellcolor[HTML]{D6EDDF}0.049 & \cellcolor[HTML]{D7EDE0}0.048 & \cellcolor[HTML]{DAEEE2}0.047 & \cellcolor[HTML]{C5E6CF}0.060 \\
 & \textbf{Video-LLaVA} & \cellcolor[HTML]{C3E5CD}0.061 & \cellcolor[HTML]{BFE3CA}0.063 & \cellcolor[HTML]{BFE4CA}0.063 & \cellcolor[HTML]{C1E4CC}0.062 & \cellcolor[HTML]{C3E5CE}0.061 & \cellcolor[HTML]{A9DBB7}0.077 \\
 & \textbf{ShareGPT4Video} & \cellcolor[HTML]{FBFCFE}0.027 & \cellcolor[HTML]{FAFBFD}0.028 & \cellcolor[HTML]{FAFBFD}0.027 & \cellcolor[HTML]{FBFCFE}0.027 & \cellcolor[HTML]{FCFCFF}0.026 & \cellcolor[HTML]{EEF7F3}0.034 \\
\multirow{-6}{*}{\textbf{GLEU ($\uparrow$)}} & \textbf{VideoChat2} & \cellcolor[HTML]{ABDCB9}0.075 & \cellcolor[HTML]{A6DAB5}0.078 & \cellcolor[HTML]{A7DAB6}0.078 & \cellcolor[HTML]{A9DAB7}0.077 & \cellcolor[HTML]{AADBB8}0.076 & \cellcolor[HTML]{85CC99}0.098\\
\bottomrule
\end{tabular}
% \vspace{-12pt}
\end{table}

Tab.~\ref{tab:traditional-metric} shows that comparing model-generated captions with GPT-summarized human annotations (aggregated GT) results in higher metric scores than direct comparisons with individual human annotations. This phenomenon can be attributed to the enhanced information coverage and evaluation fairness introduced by aggregated GT.

The FIOVA dataset adopts a multi-perspective annotation approach, where each video is described by five independent annotators. Due to individual differences in focus, some annotators emphasize character actions and event sequences, while others detail background elements or contextual information. GPT integration synthesizes these diverse descriptions into a comprehensive and representative GT, ensuring broader semantic coverage than any single human annotation. As illustrated in Fig.~\ref{fig:acc15}, this aggregation process captures both primary actions and contextual details, reducing information gaps and making the GT a more complete reference for evaluation.

The observed increase in metric scores can be explained by two key factors. First, the expanded alignment scope: because the aggregated GT integrates multi-perspective information, model-generated captions are more likely to align with at least one aspect of the GT. This increases lexical and semantic matches, leading to improved BLEU and METEOR scores. Second, the limitations of single annotator descriptions: individual annotations may prioritize specific elements while omitting others, resulting in lower recall when directly compared to model outputs.

By integrating multi-source annotations, the aggregated GT not only ensures fairness in evaluation but also aligns with FIOVA’s methodological principles for comprehensive and human-aligned assessment. 

\clearpage

\subsection{Human Validation of Evaluation Metrics}
\label{app:human-validation}

To substantiate our claim that FIOVA-DQ better aligns with human preferences than existing evaluation metrics, we conducted a supplementary experiment comparing human rankings with metric-based rankings.

\paragraph{Experimental Setup.}
We recruited ten human evaluators with backgrounds in computer vision. Each evaluator was shown six videos and asked to rank six model-generated captions per video from 1 (best) to 6 (worst), considering informativeness, accuracy, fluency, and coherence. 

The six videos used in this human evaluation correspond to the visual samples shown in  Fig.~\ref{fig:example--} to Fig.~\ref{fig:example-bad}. These samples were selected to cover a range of semantic structures, event densities, and model performance levels, ensuring a representative basis for assessing alignment between human preferences and automatic evaluation metrics.

This resulted in ten individual ranking matrices (S1–S10) and one average matrix (S-All).

\paragraph{Evaluation Metrics.}
We computed Spearman correlations between the human rankings and automatic rankings derived from six metrics: F1, Precision, and Recall under FIOVA-DQ, and F1, Precision, and Recall under AutoDQ. Tab.~\ref{tab:spearman-detail} presents detailed results for each evaluator.

\begin{table}[htbp]
\centering
\caption{Spearman correlations between human rankings (S1–S10 and S-All) and automatic metrics. Underlined values denote negative correlations.}
\label{tab:spearman-detail}
\small
\setlength{\tabcolsep}{3pt}
\renewcommand{\arraystretch}{1.2}
\begin{tabular}{l ccc c ccc}
\toprule
\textbf{Subject} & \multicolumn{3}{c}{\textbf{AutoCQ-based Metrics}} & & \multicolumn{3}{c}{\textbf{FIOVA-DQ-based Metrics}} \\
\cmidrule{2-4} \cmidrule{6-8} 
 & F1 & Recall & Precision & & F1 & Recall & Precision \\
\midrule
S1  & 0.114 & 0.038 & 0.343 & & 0.676 & 0.476 & 0.219 \\
S2  & 0.038 & \underline{-0.114} & 0.067 & & 0.210 & 0.038 & 0.238 \\
S3  & 0.105 & 0.010 & \underline{-0.095} & & 0.105 & \underline{-0.057} & 0.371 \\
S4  & \underline{-0.143} & \underline{-0.105} & \underline{-0.038} & & 0.390 & 0.267 & \underline{-0.238} \\
S5  & 0.190 & 0.295 & 0.210 & & 0.343 & 0.352 & \underline{-0.143} \\
S6  & 0.095 & 0.105 & 0.057 & & 0.390 & 0.390 & 0.362 \\
S7  & 0.095 & 0.095 & 0.171 & & 0.286 & \underline{-0.029} & 0.067 \\
S8  & \underline{-0.038} & \underline{-0.219} & 0.533 & & 0.248 & 0.010 & 0.124 \\
S9  & \underline{-0.010} & \underline{-0.019} & 0.219 & & 0.286 & 0.190 & 0.210 \\
S10 & 0.010 & \underline{-0.029} & 0.029 & & 0.229 & 0.219 & \underline{-0.171} \\
\midrule
\textbf{Avg. (S1–S10)} & 0.046 & 0.006 & 0.150 & & 0.316 & 0.186 & 0.104 \\
\textbf{S-All (ranking)} & 0.125 & 0.048 & 0.268 & & 0.579 & 0.359 & 0.239 \\
\bottomrule
\end{tabular}
\vspace{3pt}
\begin{minipage}{0.96\linewidth}
\small
\textit{Note:} “Avg. (S1–S10)” is the average over all subjects. “S-All” is the correlation between the average human ranking and each metric.
\end{minipage}
\end{table}

\paragraph{Discussion.}
The results demonstrate that FIOVA-DQ consistently achieves higher alignment with human preferences than AutoDQ across all three metrics. In particular, FIOVA-DQ's F1 score exhibits the strongest correlation with the average human ranking (0.579), while AutoDQ’s correlations are noticeably lower and more variable, including multiple negative correlations in Recall.

Despite individual variation among evaluators, FIOVA-DQ maintains greater overall robustness. In contrast, AutoDQ's performance fluctuates significantly, with some metrics correlating negatively with human judgment on certain samples.

Evaluator feedback supports these findings: participants reported penalizing hallucinations and omissions, and favoring concise yet accurate event descriptions. Notably, all ten evaluators ranked VideoChat2's caption for a video (Fig.~\ref{fig:example-0}) as the worst due to hallucinated content, reflecting rare unanimity across subjective judgments. This further validates the cognitive alignment achieved by FIOVA-DQ and supports its use as a reliable human-aligned metric for video caption evaluation.

\clearpage

\subsection{Correlation Analysis of Evaluation Metrics}
\label{app:metric-correlation-analysis}

\begin{table}[htbp]
\centering
\caption{Average Spearman correlations between nine evaluation metrics over six baseline models. Row and column blocks correspond to metric families. Underlined values denote negative correlations.}
\label{tab:spearman-metric-summary}
\small
\setlength{\tabcolsep}{3pt}
\renewcommand{\arraystretch}{1.2}
\begin{tabular}{l ccc c ccc c ccc}
\toprule
\multicolumn{1}{c}{} & \multicolumn{3}{c}{\textbf{Traditional Metrics}} & & \multicolumn{3}{c}{\textbf{AutoDQ-based Metrics}} & & \multicolumn{3}{c}{\textbf{FIOVA-DQ-based Metrics}} \\
\cmidrule{2-4} \cmidrule{6-8} \cmidrule{10-12}
\textbf{Metric} & BLEU & METEOR & GLEU & & F1 & Recall & Precision & & F1 & Recall & Precision \\
\midrule
\textbf{Traditional} \\
BLEU       & 1.000 & 0.560 & 0.787 & & \underline{-0.021} & 0.097 & \underline{-0.118} & & 0.059 & 0.046 & \underline{-0.128} \\
METEOR     & 0.560 & 1.000 & 0.582 & & \underline{-0.006} & 0.176 & \underline{-0.126} & & 0.126 & 0.113 & \underline{-0.135} \\
GLEU       & 0.787 & 0.582 & 1.000 & & \underline{-0.044} & 0.030 & \underline{-0.124} & & 0.002 & \underline{-0.015} & \underline{-0.131} \\
\addlinespace
\textbf{AutoDQ} \\
F1    & \underline{-0.021} & \underline{-0.006} & \underline{-0.044} & & 1.000 & 0.025 & 0.176 & & 0.074 & 0.051 & 0.145 \\
Recall& 0.097 & 0.176 & 0.030 & & 0.025 & 1.000 & \underline{-0.223} & & 0.733 & 0.644 & \underline{-0.212} \\
Precision & \underline{-0.118} & \underline{-0.126} & \underline{-0.124} & & 0.176 & \underline{-0.223} & 1.000 & & 0.019 & \underline{-0.022} & 0.840 \\
\addlinespace
\textbf{FIOVA-DQ} \\
F1   & 0.059 & 0.126 & 0.002 & & 0.074 & 0.733 & 0.019 & & 1.000 & 0.810 & 0.043 \\
Recall & 0.046 & 0.113 & \underline{-0.015} & & 0.051 & 0.644 & \underline{-0.022} & & 0.810 & 1.000 & \underline{-0.001} \\
Precision & \underline{-0.128} & \underline{-0.135} & \underline{-0.131} & & 0.145 & \underline{-0.212} & 0.840 & & 0.043 & \underline{-0.001} & 1.000 \\
\bottomrule
\end{tabular}
% \vspace{3pt}
% \begin{minipage}{0.96\linewidth}
% \small
% \textit{Note:} Rows and columns are grouped by metric type. Underlined values indicate negative correlations. FIOVA-DQ F1 is the only metric positively correlated with all others.
% \end{minipage}
\end{table}

To investigate the relationship between different evaluation metrics, we computed pairwise Spearman correlations over six baseline models on the full FIOVA dataset. The matrix includes nine metrics: three traditional text-matching metrics (BLEU, METEOR, GLEU), three from AutoDQ (F1, Recall, Precision), and three from FIOVA-DQ (F1, Recall, Precision). The averaged correlation matrix is shown in Tab.~\ref{tab:spearman-metric-summary}.

Key findings:
\begin{itemize}
    \item \textbf{Low correlation between linguistic and content-based metrics:} BLEU and GLEU exhibit negative or near-zero correlation with AutoDQ and FIOVA-DQ scores (\eg, BLEU–FIOVA-F1: 0.059, BLEU–Auto-F1: –0.021), indicating lexical overlap is insufficient to reflect semantic or event-level quality.
    \item \textbf{High internal consistency within event-based metrics:} AutoDQ Recall and FIOVA-DQ Recall show strong correlation (0.644), as do AutoDQ and FIOVA-DQ Precision (0.840), supporting the alignment of their event extraction paradigms. FIOVA-DQ F1 and Recall also correlate highly (0.810), reflecting internal robustness.
    \item \textbf{FIOVA-DQ-F1 is the most representative metric:} It is the only dimension that maintains positive correlation with all other metrics, making it a reliable and comprehensive choice for evaluating caption quality.
\end{itemize}

These results reinforce the limitations of traditional linguistic metrics and underscore the value of cognitively-aligned, event-based evaluation frameworks such as FIOVA-DQ.

\clearpage

\subsection{Results on Different Groups}
\label{subsec:results-on-8groups}

\begin{table}[h!]
\caption{Comparison of LVLMs on FIOVA based on traditional metrics (BLEU, METEOR, and GLEU). The background color represents the performance of the metric. The darker the green, the better the performance.}
    \label{tab:traditional-metric-on-8group}
    \centering
    \fontsize{8pt}{10pt}\selectfont
    \setlength{\tabcolsep}{3pt}
\renewcommand{\arraystretch}{1.2}
\resizebox{\textwidth}{!}{
\begin{tabular}{lllllllllll}
\toprule
 &  & \multicolumn{8}{c}{\textbf{Group}} &  \\ \cmidrule{3-10}
\multirow{-2}{*}{\textbf{Metrics}} & \multirow{-2}{*}{\textbf{LVLMs}} & \textbf{A} & \textbf{B} & \textbf{C} & \textbf{D} & \textbf{E} & \textbf{F} & \textbf{G} & \textbf{H} & \multirow{-2}{*}{\textbf{All}} \\ \midrule

 & \textbf{Tarsier} & \cellcolor[HTML]{63BE7B}0.058 & \cellcolor[HTML]{8DCF9F}0.044 & \cellcolor[HTML]{98D4A9}0.041 & \cellcolor[HTML]{93D2A4}0.042 & \cellcolor[HTML]{8ACE9D}0.045 & \cellcolor[HTML]{A2D8B1}0.038 & \cellcolor[HTML]{76C68C}0.052 & \cellcolor[HTML]{8FD0A1}0.043 & \cellcolor[HTML]{90D1A2}0.043 \\
 & \textbf{VIdeoLLaMA2} & \cellcolor[HTML]{C1E4CC}0.028 & \cellcolor[HTML]{B7E0C3}0.031 & \cellcolor[HTML]{B8E1C4}0.030 & \cellcolor[HTML]{B8E1C4}0.030 & \cellcolor[HTML]{BAE2C6}0.030 & \cellcolor[HTML]{C5E6D0}0.026 & \cellcolor[HTML]{CDE9D7}0.024 & \cellcolor[HTML]{CDE9D6}0.024 & \cellcolor[HTML]{B9E1C5}0.030 \\
 & \textbf{LLaVA-NEXT-Video} & \cellcolor[HTML]{D0EAD9}0.023 & \cellcolor[HTML]{D9EEE1}0.020 & \cellcolor[HTML]{D9EEE1}0.020 & \cellcolor[HTML]{D9EEE1}0.020 & \cellcolor[HTML]{D5EDDE}0.021 & \cellcolor[HTML]{DBEFE3}0.019 & \cellcolor[HTML]{CCE9D6}0.024 & \cellcolor[HTML]{EEF7F3}0.013 & \cellcolor[HTML]{D9EEE1}0.020 \\
 & \textbf{Video-LLaVA} & \cellcolor[HTML]{C6E6D0}0.026 & \cellcolor[HTML]{C0E4CC}0.028 & \cellcolor[HTML]{C2E5CD}0.027 & \cellcolor[HTML]{C1E4CC}0.028 & \cellcolor[HTML]{C7E7D1}0.026 & \cellcolor[HTML]{D1EBDA}0.022 & \cellcolor[HTML]{CBE9D5}0.024 & \cellcolor[HTML]{D8EEE0}0.020 & \cellcolor[HTML]{C3E5CE}0.027 \\
 & \textbf{ShareGPT4Video} & \cellcolor[HTML]{ECF6F1}0.014 & \cellcolor[HTML]{F6FAF9}0.011 & \cellcolor[HTML]{F6FAFA}0.011 & \cellcolor[HTML]{F7FAFB}0.010 & \cellcolor[HTML]{F7FAFB}0.010 & \cellcolor[HTML]{FCFCFF}0.008 & \cellcolor[HTML]{F9FBFC}0.010 & \cellcolor[HTML]{F7FAFB}0.010 & \cellcolor[HTML]{F6FAFA}0.010 \\
\multirow{-6}{*}{\textbf{BLEU ($\uparrow$)}} & \textbf{VideoChat2} & \cellcolor[HTML]{96D3A7}0.041 & \cellcolor[HTML]{A2D8B1}0.037 & \cellcolor[HTML]{A5D9B4}0.036 & \cellcolor[HTML]{A4D9B3}0.037 & \cellcolor[HTML]{A6D9B4}0.036 & \cellcolor[HTML]{B5DFC2}0.031 & \cellcolor[HTML]{A3D8B2}0.037 & \cellcolor[HTML]{BEE3C9}0.028 & \cellcolor[HTML]{A4D9B3}0.037 \\

\midrule

 & \textbf{Tarsier} & \cellcolor[HTML]{65BF7D}0.288 & \cellcolor[HTML]{8DCF9F}0.267 & \cellcolor[HTML]{95D3A6}0.263 & \cellcolor[HTML]{90D1A2}0.265 & \cellcolor[HTML]{91D1A3}0.265 & \cellcolor[HTML]{A3D8B3}0.255 & \cellcolor[HTML]{93D2A4}0.264 & \cellcolor[HTML]{66BF7D}0.288 & \cellcolor[HTML]{90D1A2}0.265 \\
 & \textbf{VIdeoLLaMA2} & \cellcolor[HTML]{78C78E}0.278 & \cellcolor[HTML]{85CC99}0.271 & \cellcolor[HTML]{8DCFA0}0.267 & \cellcolor[HTML]{89CE9C}0.269 & \cellcolor[HTML]{90D1A2}0.265 & \cellcolor[HTML]{9BD5AB}0.260 & \cellcolor[HTML]{A3D8B2}0.255 & \cellcolor[HTML]{99D4AA}0.260 & \cellcolor[HTML]{8BCE9D}0.268 \\
 & \textbf{LLaVA-NEXT-Video} & \cellcolor[HTML]{7BC890}0.277 & \cellcolor[HTML]{83CB96}0.272 & \cellcolor[HTML]{85CC99}0.271 & \cellcolor[HTML]{8CCF9E}0.268 & \cellcolor[HTML]{8ED0A0}0.267 & \cellcolor[HTML]{95D3A6}0.263 & \cellcolor[HTML]{92D1A4}0.264 & \cellcolor[HTML]{80CA94}0.274 & \cellcolor[HTML]{87CD9B}0.270 \\
 & \textbf{Video-LLaVA} & \cellcolor[HTML]{91D1A3}0.265 & \cellcolor[HTML]{96D3A7}0.262 & \cellcolor[HTML]{A3D8B2}0.255 & \cellcolor[HTML]{9AD4AA}0.260 & \cellcolor[HTML]{AFDDBD}0.249 & \cellcolor[HTML]{BDE3C9}0.241 & \cellcolor[HTML]{B4DFC1}0.246 & \cellcolor[HTML]{D5ECDD}0.229 & \cellcolor[HTML]{9FD7AF}0.257 \\
 & \textbf{ShareGPT4Video} & \cellcolor[HTML]{B9E1C5}0.244 & \cellcolor[HTML]{E4F3EA}0.221 & \cellcolor[HTML]{E7F4ED}0.219 & \cellcolor[HTML]{F2F8F7}0.213 & \cellcolor[HTML]{EFF7F3}0.215 & \cellcolor[HTML]{FCFCFF}0.208 & \cellcolor[HTML]{F0F7F4}0.215 & \cellcolor[HTML]{E0F1E6}0.223 & \cellcolor[HTML]{E9F5EF}0.218 \\
\multirow{-6}{*}{\textbf{METEOR ($\uparrow$)}} & \textbf{VideoChat2} & \cellcolor[HTML]{63BE7B}0.289 & \cellcolor[HTML]{68C080}0.286 & \cellcolor[HTML]{75C68B}0.279 & \cellcolor[HTML]{72C488}0.281 & \cellcolor[HTML]{7BC890}0.277 & \cellcolor[HTML]{8DCFA0}0.267 & \cellcolor[HTML]{84CC97}0.272 & \cellcolor[HTML]{89CE9C}0.269 & \cellcolor[HTML]{72C488}0.281 \\

\midrule

& \textbf{Tarsier} & \cellcolor[HTML]{63BE7B}0.139 & \cellcolor[HTML]{7FCA93}0.120 & \cellcolor[HTML]{83CB97}0.117 & \cellcolor[HTML]{81CB95}0.118 & \cellcolor[HTML]{80CA94}0.119 & \cellcolor[HTML]{89CE9C}0.113 & \cellcolor[HTML]{79C78E}0.124 & \cellcolor[HTML]{67C07E}0.137 & \cellcolor[HTML]{80CA94}0.119 \\
 & \textbf{VIdeoLLaMA2} & \cellcolor[HTML]{AEDDBB}0.086 & \cellcolor[HTML]{ABDCBA}0.088 & \cellcolor[HTML]{AADBB9}0.089 & \cellcolor[HTML]{ACDCBA}0.087 & \cellcolor[HTML]{ACDCBA}0.087 & \cellcolor[HTML]{AFDDBD}0.085 & \cellcolor[HTML]{B1DEBF}0.084 & \cellcolor[HTML]{AADBB8}0.089 & \cellcolor[HTML]{ACDCBA}0.088 \\
 & \textbf{LLaVA-NEXT-Video} & \cellcolor[HTML]{D0EBD9}0.062 & \cellcolor[HTML]{D4ECDD}0.059 & \cellcolor[HTML]{D3ECDC}0.060 & \cellcolor[HTML]{D4ECDD}0.059 & \cellcolor[HTML]{D0EAD9}0.062 & \cellcolor[HTML]{D2EBDB}0.060 & \cellcolor[HTML]{C3E5CE}0.071 & \cellcolor[HTML]{CFEAD8}0.063 & \cellcolor[HTML]{D3ECDC}0.060 \\
 & \textbf{Video-LLaVA} & \cellcolor[HTML]{BAE2C6}0.078 & \cellcolor[HTML]{BAE2C6}0.077 & \cellcolor[HTML]{BDE3C8}0.076 & \cellcolor[HTML]{BAE1C6}0.078 & \cellcolor[HTML]{BCE2C8}0.076 & \cellcolor[HTML]{C2E5CD}0.072 & \cellcolor[HTML]{BDE3C9}0.076 & \cellcolor[HTML]{CAE8D4}0.066 & \cellcolor[HTML]{BBE2C7}0.077 \\
 & \textbf{ShareGPT4Video} & \cellcolor[HTML]{EDF6F2}0.041 & \cellcolor[HTML]{F5F9F9}0.035 & \cellcolor[HTML]{F6FAFA}0.035 & \cellcolor[HTML]{F7FAFB}0.034 & \cellcolor[HTML]{F9FBFC}0.033 & \cellcolor[HTML]{FCFCFF}0.030 & \cellcolor[HTML]{F7FAFA}0.034 & \cellcolor[HTML]{F2F8F7}0.037 & \cellcolor[HTML]{F6FAFA}0.034 \\
\multirow{-6}{*}{\textbf{GLEU ($\uparrow$)}} & \textbf{VideoChat2} & \cellcolor[HTML]{93D2A4}0.106 & \cellcolor[HTML]{9DD6AD}0.098 & \cellcolor[HTML]{9DD6AD}0.098 & \cellcolor[HTML]{9ED6AE}0.098 & \cellcolor[HTML]{9DD6AD}0.098 & \cellcolor[HTML]{A4D9B4}0.093 & \cellcolor[HTML]{97D3A8}0.103 & \cellcolor[HTML]{9AD4AA}0.101 & \cellcolor[HTML]{9DD6AD}0.098 \\

\bottomrule

\end{tabular}}
\end{table}

\begin{table}[h!]
\caption{Comparison of LVLMs on FIOVA based on AutoDQ-based metrics. The background color represents the performance of the metric. The darker the green, the better the performance.}
    \label{tab:autodq-metric-on-8group}
    \centering
    \fontsize{8pt}{10pt}\selectfont
        \setlength{\tabcolsep}{3pt}
\renewcommand{\arraystretch}{1.2}
\resizebox{\textwidth}{!}{
\begin{tabular}{lllllllllll}
\toprule
 &  & \multicolumn{8}{c}{\textbf{Group}} &  \\ \cmidrule{3-10}
\multirow{-2}{*}{\textbf{Metrics}} & \multirow{-2}{*}{\textbf{LVLMs}} & \textbf{A} & \textbf{B} & \textbf{C} & \textbf{D} & \textbf{E} & \textbf{F} & \textbf{G} & \textbf{H} & \multirow{-2}{*}{\textbf{All}} \\ \midrule

& \textbf{Tarsier} & \cellcolor[HTML]{63BE7B}0.366 & \cellcolor[HTML]{72C588}0.346 & \cellcolor[HTML]{6FC386}0.350 & \cellcolor[HTML]{68C080}0.359 & \cellcolor[HTML]{6FC385}0.350 & \cellcolor[HTML]{6BC282}0.355 & \cellcolor[HTML]{7FC993}0.329 & \cellcolor[HTML]{83CB96}0.324 & \cellcolor[HTML]{6FC385}0.351 \\
 & \textbf{VIdeoLLaMA2} & \cellcolor[HTML]{72C488}0.346 & \cellcolor[HTML]{80CA94}0.328 & \cellcolor[HTML]{88CD9B}0.316 & \cellcolor[HTML]{7DC991}0.332 & \cellcolor[HTML]{82CB96}0.325 & \cellcolor[HTML]{82CB96}0.324 & \cellcolor[HTML]{92D1A3}0.304 & \cellcolor[HTML]{A0D7AF}0.285 & \cellcolor[HTML]{82CB96}0.325 \\
 & \textbf{LLaVA-NEXT-Video} & \cellcolor[HTML]{84CC98}0.322 & \cellcolor[HTML]{96D3A7}0.297 & \cellcolor[HTML]{93D2A4}0.302 & \cellcolor[HTML]{93D2A5}0.302 & \cellcolor[HTML]{92D1A3}0.304 & \cellcolor[HTML]{93D2A4}0.302 & \cellcolor[HTML]{A0D7B0}0.284 & \cellcolor[HTML]{ACDCBA}0.268 & \cellcolor[HTML]{94D2A5}0.301 \\
 & \textbf{Video-LLaVA} & \cellcolor[HTML]{92D1A3}0.304 & \cellcolor[HTML]{A1D7B0}0.283 & \cellcolor[HTML]{A2D8B1}0.282 & \cellcolor[HTML]{9ED6AE}0.287 & \cellcolor[HTML]{9ED6AE}0.288 & \cellcolor[HTML]{9AD5AB}0.292 & \cellcolor[HTML]{AEDDBC}0.265 & \cellcolor[HTML]{7DC992}0.331 & \cellcolor[HTML]{9FD7AF}0.285 \\
 & \textbf{ShareGPT4Video} & \cellcolor[HTML]{A6D9B4}0.277 & \cellcolor[HTML]{A6D9B5}0.276 & \cellcolor[HTML]{A7DAB6}0.274 & \cellcolor[HTML]{98D4A9}0.295 & \cellcolor[HTML]{9FD7AF}0.285 & \cellcolor[HTML]{A4D9B3}0.279 & \cellcolor[HTML]{90D1A2}0.306 & \cellcolor[HTML]{86CC99}0.320 & \cellcolor[HTML]{A2D8B2}0.281 \\
\multirow{-6}{*}{\textbf{F1 (AutoDQ) ($\uparrow$)}} & \textbf{VideoChat2} & \cellcolor[HTML]{89CE9C}0.315 & \cellcolor[HTML]{89CE9C}0.315 & \cellcolor[HTML]{92D1A4}0.303 & \cellcolor[HTML]{87CD9A}0.318 & \cellcolor[HTML]{94D2A5}0.301 & \cellcolor[HTML]{96D3A7}0.297 & \cellcolor[HTML]{B6E0C3}0.255 & \cellcolor[HTML]{FCFCFF}0.160 & \cellcolor[HTML]{8ED0A0}0.309 \\

\midrule

 & \textbf{Tarsier} & \cellcolor[HTML]{63BE7B}0.333 & \cellcolor[HTML]{78C78E}0.305 & \cellcolor[HTML]{8BCF9E}0.279 & \cellcolor[HTML]{8BCF9E}0.280 & \cellcolor[HTML]{96D3A7}0.265 & \cellcolor[HTML]{B3DFC0}0.226 & \cellcolor[HTML]{BEE3C9}0.212 & \cellcolor[HTML]{CCE9D6}0.193 & \cellcolor[HTML]{88CD9B}0.283 \\
 & \textbf{VIdeoLLaMA2} & \cellcolor[HTML]{86CD99}0.286 & \cellcolor[HTML]{94D2A5}0.268 & \cellcolor[HTML]{A7DAB5}0.243 & \cellcolor[HTML]{A7DAB6}0.242 & \cellcolor[HTML]{B7E0C3}0.222 & \cellcolor[HTML]{D9EEE0}0.176 & \cellcolor[HTML]{E7F4ED}0.157 & \cellcolor[HTML]{EEF7F3}0.147 & \cellcolor[HTML]{A5D9B4}0.245 \\
 & \textbf{LLaVA-NEXT-Video} & \cellcolor[HTML]{A0D7B0}0.252 & \cellcolor[HTML]{A8DAB6}0.241 & \cellcolor[HTML]{B3DFC0}0.227 & \cellcolor[HTML]{BCE2C8}0.215 & \cellcolor[HTML]{CCE9D6}0.193 & \cellcolor[HTML]{ECF6F1}0.151 & \cellcolor[HTML]{D7EDDF}0.179 & \cellcolor[HTML]{DFF0E6}0.168 & \cellcolor[HTML]{B7E0C4}0.221 \\
 & \textbf{Video-LLaVA} & \cellcolor[HTML]{BFE4CA}0.211 & \cellcolor[HTML]{B1DEBE}0.229 & \cellcolor[HTML]{C2E5CD}0.207 & \cellcolor[HTML]{C2E5CD}0.207 & \cellcolor[HTML]{D4ECDC}0.183 & \cellcolor[HTML]{ECF6F2}0.150 & \cellcolor[HTML]{EEF7F3}0.148 & \cellcolor[HTML]{D3ECDC}0.183 & \cellcolor[HTML]{C1E4CC}0.208 \\
 & \textbf{ShareGPT4Video} & \cellcolor[HTML]{B1DEBE}0.229 & \cellcolor[HTML]{BBE2C7}0.216 & \cellcolor[HTML]{C4E6CF}0.204 & \cellcolor[HTML]{CAE8D4}0.196 & \cellcolor[HTML]{D4ECDC}0.183 & \cellcolor[HTML]{EDF6F2}0.149 & \cellcolor[HTML]{FBFCFE}0.130 & \cellcolor[HTML]{F4F9F8}0.140 & \cellcolor[HTML]{C6E7D1}0.201 \\
\multirow{-6}{*}{\textbf{Recall (AutoDQ)  ($\uparrow$)}} & \textbf{VideoChat2} & \cellcolor[HTML]{75C68B}0.309 & \cellcolor[HTML]{9CD5AC}0.257 & \cellcolor[HTML]{B0DDBD}0.231 & \cellcolor[HTML]{ADDCBB}0.235 & \cellcolor[HTML]{BFE4CA}0.211 & \cellcolor[HTML]{D1EBDA}0.186 & \cellcolor[HTML]{CAE8D4}0.195 & \cellcolor[HTML]{FCFCFF}0.128 & \cellcolor[HTML]{ABDCBA}0.237 \\

\midrule

& \textbf{Tarsier} & \cellcolor[HTML]{FCFCFF}0.548 & \cellcolor[HTML]{D7EDE0}0.609 & \cellcolor[HTML]{CDE9D7}0.626 & \cellcolor[HTML]{C4E6CF}0.642 & \cellcolor[HTML]{B9E1C6}0.659 & \cellcolor[HTML]{C2E5CD}0.645 & \cellcolor[HTML]{B5DFC2}0.667 & \cellcolor[HTML]{9AD5AB}0.711 & \cellcolor[HTML]{CCE9D6}0.628 \\
 & \textbf{VIdeoLLaMA2} & \cellcolor[HTML]{B9E1C6}0.659 & \cellcolor[HTML]{B8E1C4}0.662 & \cellcolor[HTML]{ACDCBA}0.681 & \cellcolor[HTML]{ACDCBA}0.682 & \cellcolor[HTML]{A2D8B2}0.698 & \cellcolor[HTML]{91D1A2}0.727 & \cellcolor[HTML]{77C68D}0.769 & \cellcolor[HTML]{88CD9B}0.741 & \cellcolor[HTML]{ADDCBB}0.680 \\
 & \textbf{LLaVA-NEXT-Video} & \cellcolor[HTML]{E1F1E8}0.593 & \cellcolor[HTML]{B7E0C3}0.664 & \cellcolor[HTML]{B5E0C2}0.666 & \cellcolor[HTML]{AEDDBC}0.678 & \cellcolor[HTML]{9DD6AD}0.707 & \cellcolor[HTML]{99D4AA}0.712 & \cellcolor[HTML]{B3DFC0}0.669 & \cellcolor[HTML]{8FD0A1}0.730 & \cellcolor[HTML]{B1DEBE}0.674 \\
 & \textbf{Video-LLaVA} & \cellcolor[HTML]{BBE2C7}0.657 & \cellcolor[HTML]{ABDBB9}0.684 & \cellcolor[HTML]{9CD6AD}0.707 & \cellcolor[HTML]{9CD5AC}0.708 & \cellcolor[HTML]{86CC99}0.745 & \cellcolor[HTML]{63BE7B}0.802 & \cellcolor[HTML]{79C78E}0.766 & \cellcolor[HTML]{64BF7C}0.801 & \cellcolor[HTML]{9CD5AC}0.709 \\
 & \textbf{ShareGPT4Video} & \cellcolor[HTML]{A2D8B2}0.698 & \cellcolor[HTML]{95D2A6}0.720 & \cellcolor[HTML]{8FD0A1}0.730 & \cellcolor[HTML]{8CCF9E}0.735 & \cellcolor[HTML]{87CD9A}0.743 & \cellcolor[HTML]{7EC993}0.758 & \cellcolor[HTML]{7CC991}0.761 & \cellcolor[HTML]{72C488}0.779 & \cellcolor[HTML]{8FD0A1}0.731 \\
\multirow{-6}{*}{\textbf{Precision (AutoDQ) ($\uparrow$)}} & \textbf{VideoChat2} & \cellcolor[HTML]{DAEEE2}0.605 & \cellcolor[HTML]{C9E8D3}0.633 & \cellcolor[HTML]{B9E1C6}0.659 & \cellcolor[HTML]{B6E0C3}0.665 & \cellcolor[HTML]{AEDDBC}0.679 & \cellcolor[HTML]{9DD6AD}0.707 & \cellcolor[HTML]{8FD0A1}0.730 & \cellcolor[HTML]{C7E7D1}0.637 & \cellcolor[HTML]{BBE2C7}0.656 \\

\bottomrule

\end{tabular}}
\end{table}

\begin{table}[h!]
\caption{Comparison of LVLMs on FIOVA based on FIOVA-DQ. The background color represents the performance of the metric. The darker the green, the better the performance.}
    \label{tab:fiovadq-metric-on-8group}
    \centering
    \fontsize{8pt}{10pt}\selectfont
        \setlength{\tabcolsep}{3pt}
\renewcommand{\arraystretch}{1.2}
\resizebox{\textwidth}{!}{
\begin{tabular}{lllllllllll}
\toprule
 &  & \multicolumn{8}{c}{\textbf{Group}} &  \\ \cmidrule{3-10}
\multirow{-2}{*}{\textbf{Metrics}} & \multirow{-2}{*}{\textbf{LVLMs}} & \textbf{A} & \textbf{B} & \textbf{C} & \textbf{D} & \textbf{E} & \textbf{F} & \textbf{G} & \textbf{H} & \multirow{-2}{*}{\textbf{All}} \\ \midrule

 & \textbf{Tarsier} & \cellcolor[HTML]{83CB97}0.318 & \cellcolor[HTML]{7BC890}0.331 & \cellcolor[HTML]{87CD9A}0.312 & \cellcolor[HTML]{80CA94}0.324 & \cellcolor[HTML]{80CA94}0.324 & \cellcolor[HTML]{A2D8B2}0.271 & \cellcolor[HTML]{BCE2C8}0.231 & \cellcolor[HTML]{C0E4CB}0.226 & \cellcolor[HTML]{83CB96}0.320 \\
 & \textbf{VIdeoLLaMA2} & \cellcolor[HTML]{63BE7B}0.367 & \cellcolor[HTML]{7DC992}0.328 & \cellcolor[HTML]{93D2A4}0.295 & \cellcolor[HTML]{8CCF9F}0.305 & \cellcolor[HTML]{98D4A9}0.286 & \cellcolor[HTML]{B8E1C4}0.238 & \cellcolor[HTML]{CAE8D4}0.211 & \cellcolor[HTML]{E2F2E8}0.174 & \cellcolor[HTML]{8DCF9F}0.304 \\
 & \textbf{LLaVA-NEXT-Video} & \cellcolor[HTML]{97D3A8}0.288 & \cellcolor[HTML]{8DCFA0}0.303 & \cellcolor[HTML]{95D2A6}0.292 & \cellcolor[HTML]{97D3A8}0.289 & \cellcolor[HTML]{A9DBB7}0.261 & \cellcolor[HTML]{CCE9D6}0.207 & \cellcolor[HTML]{C3E5CE}0.220 & \cellcolor[HTML]{8FD0A1}0.301 & \cellcolor[HTML]{98D4A9}0.286 \\
 & \textbf{Video-LLaVA} & \cellcolor[HTML]{A3D8B2}0.270 & \cellcolor[HTML]{98D4A9}0.287 & \cellcolor[HTML]{A9DBB7}0.261 & \cellcolor[HTML]{9FD7AF}0.275 & \cellcolor[HTML]{AFDDBC}0.252 & \cellcolor[HTML]{C7E7D1}0.215 & \cellcolor[HTML]{C7E7D1}0.215 & \cellcolor[HTML]{BDE3C9}0.230 & \cellcolor[HTML]{A4D9B3}0.269 \\
 & \textbf{ShareGPT4Video} & \cellcolor[HTML]{A2D8B1}0.272 & \cellcolor[HTML]{A0D7B0}0.275 & \cellcolor[HTML]{A7DAB6}0.264 & \cellcolor[HTML]{A5D9B4}0.266 & \cellcolor[HTML]{AFDDBD}0.251 & \cellcolor[HTML]{C7E7D1}0.215 & \cellcolor[HTML]{DFF1E6}0.178 & \cellcolor[HTML]{CFEAD8}0.203 & \cellcolor[HTML]{A7DAB6}0.263 \\
\multirow{-6}{*}{\textbf{F1 (FIOVA-DQ) ($\uparrow$)}} & \textbf{VideoChat2} & \cellcolor[HTML]{77C68C}0.337 & \cellcolor[HTML]{8FD0A1}0.301 & \cellcolor[HTML]{9CD5AC}0.281 & \cellcolor[HTML]{95D3A6}0.292 & \cellcolor[HTML]{A3D8B2}0.270 & \cellcolor[HTML]{B8E1C4}0.238 & \cellcolor[HTML]{B2DEBF}0.246 & \cellcolor[HTML]{FCFCFF}0.133 & \cellcolor[HTML]{98D4A9}0.287 \\

\midrule

& \textbf{Tarsier} & \cellcolor[HTML]{9DD6AD}0.485 & \cellcolor[HTML]{89CE9C}0.567 & \cellcolor[HTML]{87CD9A}0.575 & \cellcolor[HTML]{81CA95}0.599 & \cellcolor[HTML]{7DC992}0.613 & \cellcolor[HTML]{78C78D}0.633 & \cellcolor[HTML]{7BC88F}0.623 & \cellcolor[HTML]{63BE7B}0.716 & \cellcolor[HTML]{84CC98}0.584 \\
 & \textbf{VIdeoLLaMA2} & \cellcolor[HTML]{C6E6D1}0.321 & \cellcolor[HTML]{D1EBDA}0.277 & \cellcolor[HTML]{DAEEE2}0.243 & \cellcolor[HTML]{D8EEE0}0.249 & \cellcolor[HTML]{DEF0E5}0.227 & \cellcolor[HTML]{ECF6F1}0.170 & \cellcolor[HTML]{F1F8F6}0.149 & \cellcolor[HTML]{FAFBFD}0.114 & \cellcolor[HTML]{D8EEE0}0.250 \\
 & \textbf{LLaVA-NEXT-Video} & \cellcolor[HTML]{D4ECDD}0.264 & \cellcolor[HTML]{D9EEE0}0.248 & \cellcolor[HTML]{DCEFE3}0.235 & \cellcolor[HTML]{DEF0E5}0.226 & \cellcolor[HTML]{E4F3EA}0.202 & \cellcolor[HTML]{F3F8F7}0.144 & \cellcolor[HTML]{EBF6F1}0.173 & \cellcolor[HTML]{E2F2E9}0.210 & \cellcolor[HTML]{DDF0E5}0.229 \\
 & \textbf{Video-LLaVA} & \cellcolor[HTML]{DDF0E4}0.230 & \cellcolor[HTML]{DAEFE2}0.241 & \cellcolor[HTML]{E3F2E9}0.207 & \cellcolor[HTML]{E1F1E7}0.215 & \cellcolor[HTML]{E6F3EC}0.194 & \cellcolor[HTML]{F0F7F4}0.156 & \cellcolor[HTML]{F0F7F5}0.154 & \cellcolor[HTML]{EDF6F2}0.165 & \cellcolor[HTML]{E1F1E7}0.216 \\
 & \textbf{ShareGPT4Video} & \cellcolor[HTML]{E0F1E7}0.219 & \cellcolor[HTML]{E0F1E7}0.218 & \cellcolor[HTML]{E4F3EA}0.202 & \cellcolor[HTML]{E3F2EA}0.205 & \cellcolor[HTML]{E8F4EE}0.185 & \cellcolor[HTML]{F0F7F4}0.156 & \cellcolor[HTML]{F8FAFB}0.124 & \cellcolor[HTML]{F2F8F7}0.145 & \cellcolor[HTML]{E4F2EA}0.203 \\
\multirow{-6}{*}{\textbf{Recall (FIOVA-DQ) ($\uparrow$)}} & \textbf{VideoChat2} & \cellcolor[HTML]{C9E8D3}0.308 & \cellcolor[HTML]{D4ECDC}0.266 & \cellcolor[HTML]{DCEFE4}0.233 & \cellcolor[HTML]{D9EEE1}0.246 & \cellcolor[HTML]{DFF1E6}0.220 & \cellcolor[HTML]{E8F4EE}0.186 & \cellcolor[HTML]{EBF5F0}0.175 & \cellcolor[HTML]{FCFCFF}0.104 & \cellcolor[HTML]{DAEEE1}0.243 \\

\midrule
 
 & \textbf{Tarsier} & \cellcolor[HTML]{FCFCFF}0.485 & \cellcolor[HTML]{D5EDDE}0.567 & \cellcolor[HTML]{D1EBDA}0.575 & \cellcolor[HTML]{C6E6D0}0.599 & \cellcolor[HTML]{BFE4CA}0.613 & \cellcolor[HTML]{B5E0C2}0.633 & \cellcolor[HTML]{BAE2C6}0.623 & \cellcolor[HTML]{8ED0A0}0.716 & \cellcolor[HTML]{CDE9D7}0.584 \\
 & \textbf{VIdeoLLaMA2} & \cellcolor[HTML]{B8E1C4}0.627 & \cellcolor[HTML]{B8E1C4}0.628 & \cellcolor[HTML]{B2DEBF}0.640 & \cellcolor[HTML]{AFDDBC}0.647 & \cellcolor[HTML]{A3D8B2}0.671 & \cellcolor[HTML]{94D2A5}0.702 & \cellcolor[HTML]{85CC99}0.733 & \cellcolor[HTML]{A1D8B1}0.674 & \cellcolor[HTML]{B0DDBD}0.645 \\
 & \textbf{LLaVA-NEXT-Video} & \cellcolor[HTML]{DEF0E5}0.549 & \cellcolor[HTML]{B6E0C3}0.632 & \cellcolor[HTML]{B4DFC1}0.635 & \cellcolor[HTML]{ACDCBA}0.652 & \cellcolor[HTML]{9FD7AF}0.680 & \cellcolor[HTML]{9ED6AE}0.682 & \cellcolor[HTML]{ACDCBA}0.652 & \cellcolor[HTML]{99D4A9}0.692 & \cellcolor[HTML]{B0DEBE}0.644 \\
 & \textbf{Video-LLaVA} & \cellcolor[HTML]{B7E0C3}0.630 & \cellcolor[HTML]{AFDDBD}0.646 & \cellcolor[HTML]{9DD6AD}0.684 & \cellcolor[HTML]{A0D7B0}0.677 & \cellcolor[HTML]{89CE9C}0.725 & \cellcolor[HTML]{67C07F}0.795 & \cellcolor[HTML]{7AC88F}0.755 & \cellcolor[HTML]{8ACE9C}0.724 & \cellcolor[HTML]{9FD6AE}0.680 \\
 & \textbf{ShareGPT4Video} & \cellcolor[HTML]{98D4A9}0.694 & \cellcolor[HTML]{92D2A4}0.706 & \cellcolor[HTML]{92D1A3}0.707 & \cellcolor[HTML]{8FD0A1}0.713 & \cellcolor[HTML]{85CC98}0.734 & \cellcolor[HTML]{7DC991}0.750 & \cellcolor[HTML]{75C68B}0.766 & \cellcolor[HTML]{63BE7B}0.803 & \cellcolor[HTML]{8FD0A1}0.714 \\
\multirow{-6}{*}{\textbf{Precision (FIOVA-DQ) ($\uparrow$)}} & \textbf{VideoChat2} & \cellcolor[HTML]{DCEFE3}0.553 & \cellcolor[HTML]{C9E8D3}0.591 & \cellcolor[HTML]{BBE2C7}0.622 & \cellcolor[HTML]{B4DFC1}0.636 & \cellcolor[HTML]{ABDBB9}0.654 & \cellcolor[HTML]{A2D8B1}0.674 & \cellcolor[HTML]{86CD99}0.731 & \cellcolor[HTML]{A8DAB7}0.661 & \cellcolor[HTML]{BBE2C7}0.621\\

\bottomrule

\end{tabular}}
\end{table}

\clearpage

\begin{sidewaysfigure}[ht!]
\centering
\includegraphics[width=\textwidth]{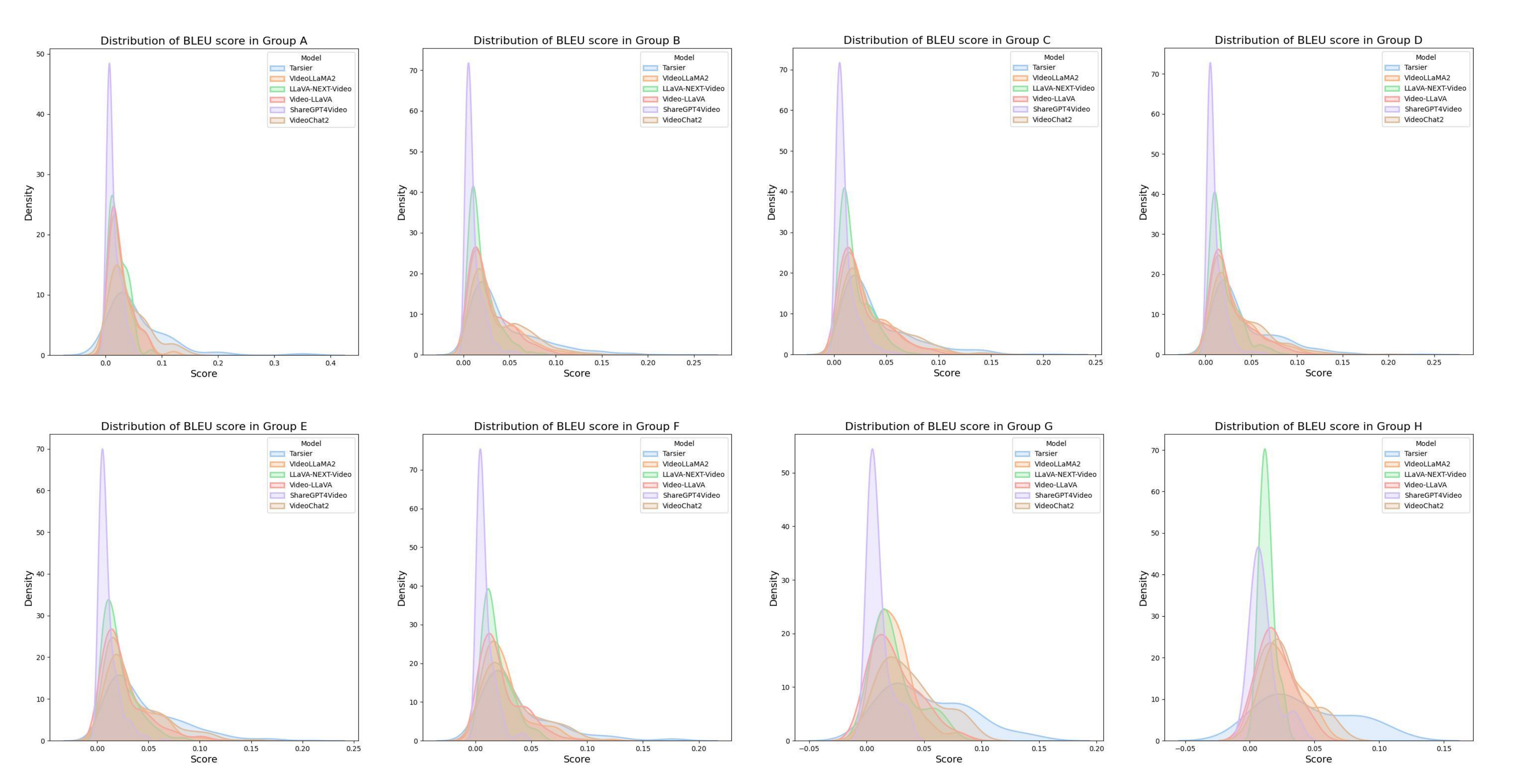}
\caption{Distribution of LVLMs scores in different groups, based on BLEU metric.}
\label{fig:BLEU}
\end{sidewaysfigure}

\clearpage

\begin{sidewaysfigure}[ht!]
\centering
\includegraphics[width=\textwidth]{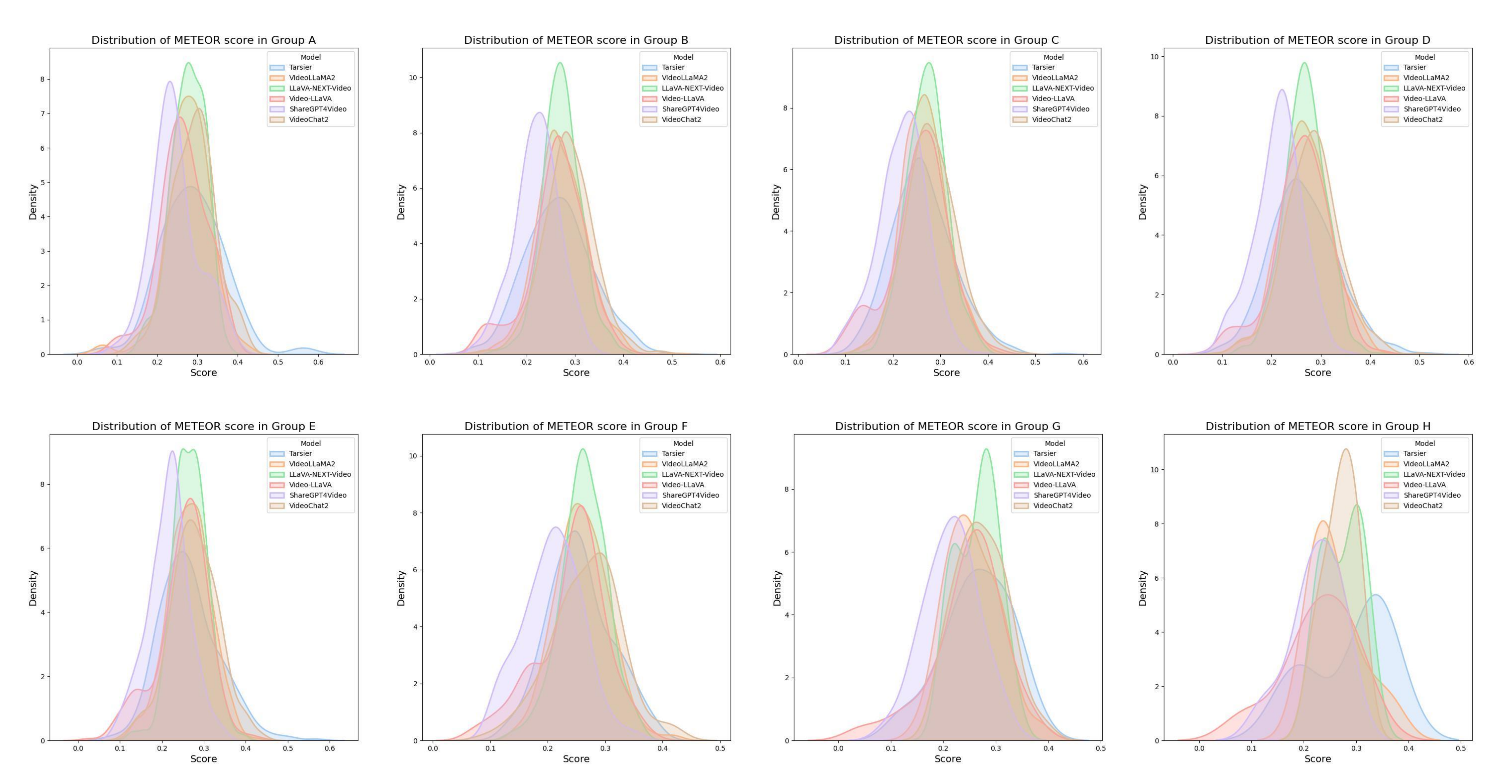}
\caption{Distribution of LVLMs scores in different groups, based on METEOR metric.}
\label{fig:METEOR}
\end{sidewaysfigure}

\clearpage

\begin{sidewaysfigure}[ht!]
\centering
\includegraphics[width=\textwidth]{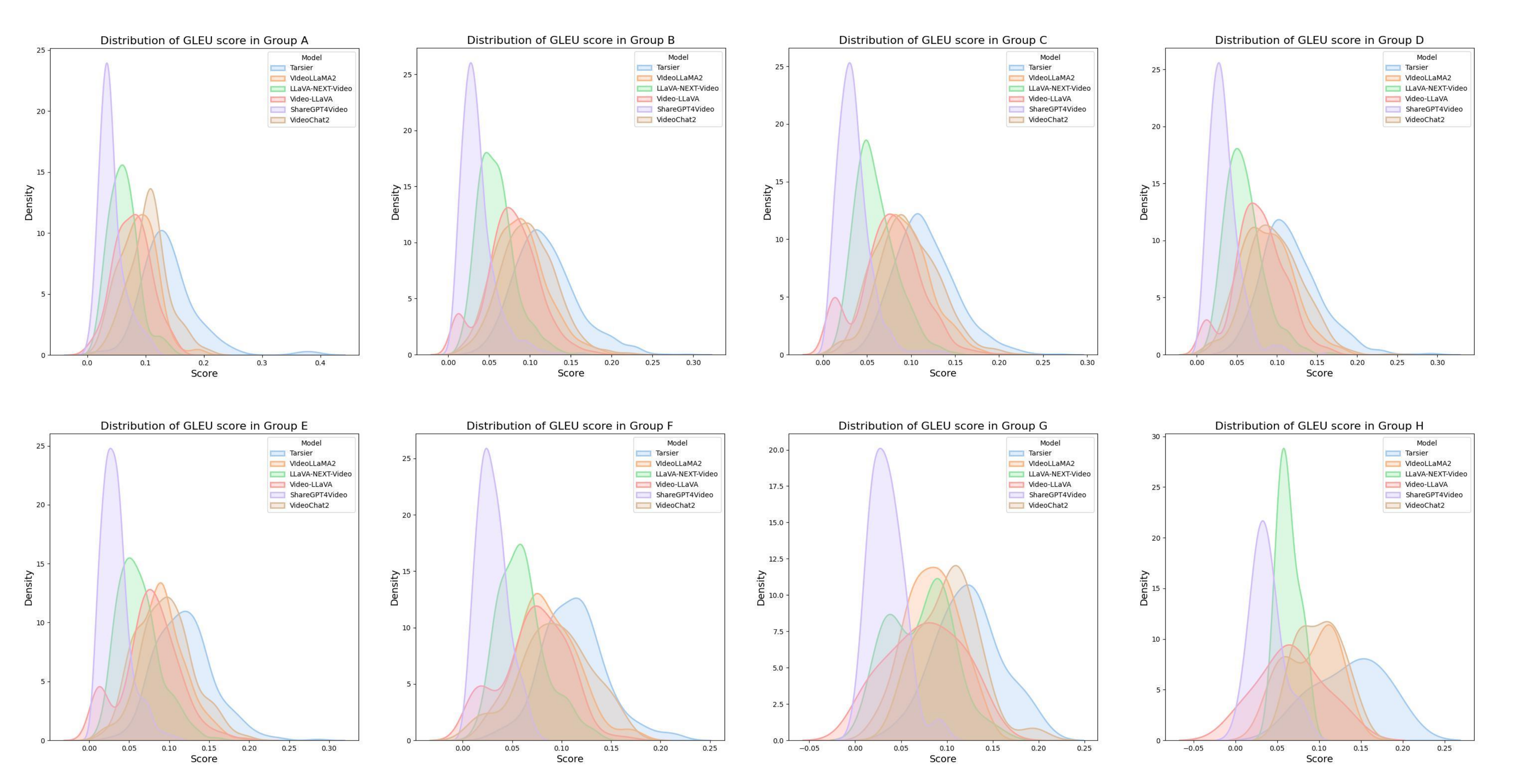}
\caption{Distribution of LVLMs scores in different groups, based on GLEU metric.}
\label{fig:GLEU}
\end{sidewaysfigure}

\clearpage

\begin{sidewaysfigure}[ht!]
\centering
\includegraphics[width=\textwidth]{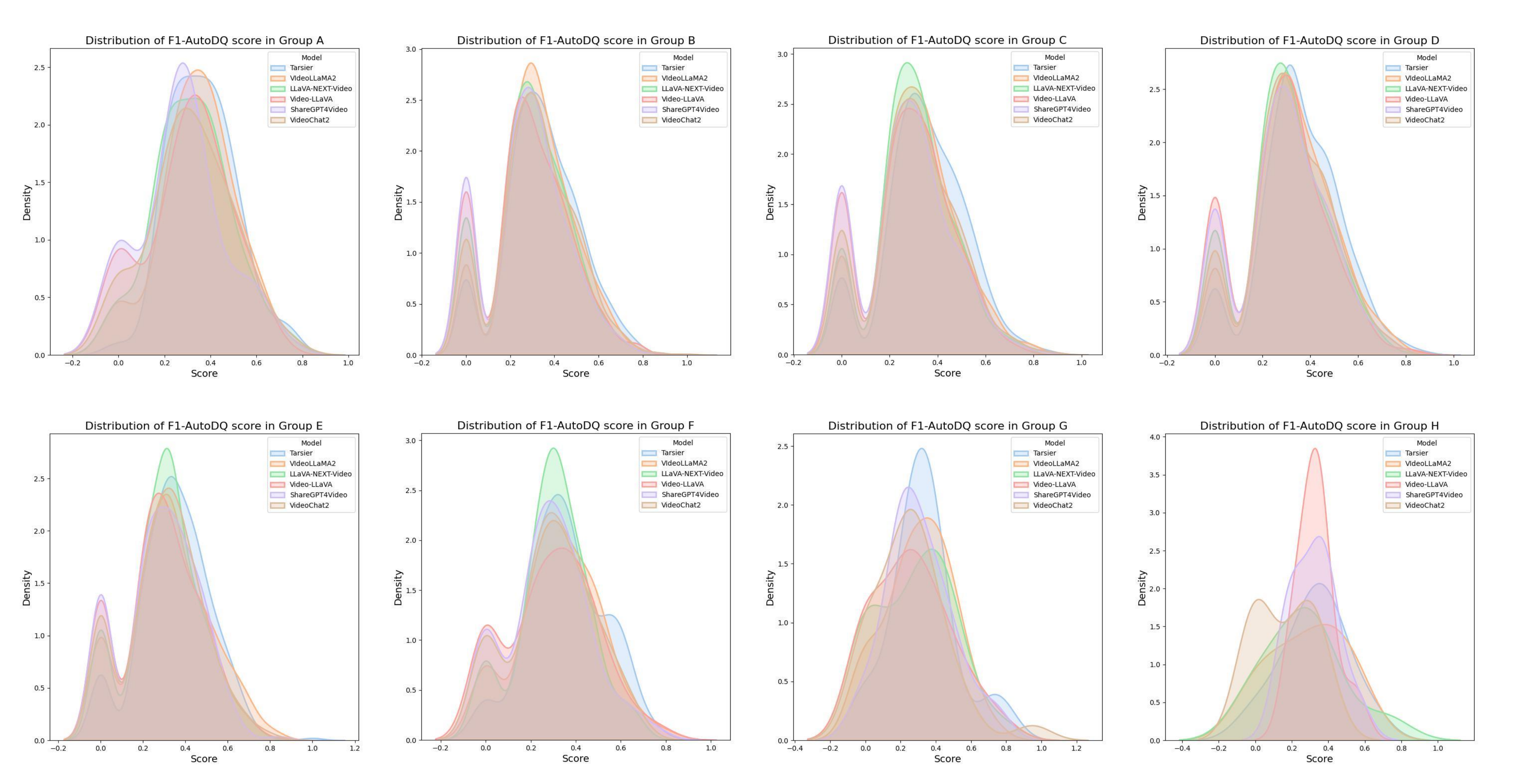}
\caption{Distribution of LVLMs scores in different groups, based on F1 (AutoDQ) metric.}
\label{fig:F1-AutoDQ}
\end{sidewaysfigure}

\clearpage

\begin{sidewaysfigure}[ht!]
\centering
\includegraphics[width=\textwidth]{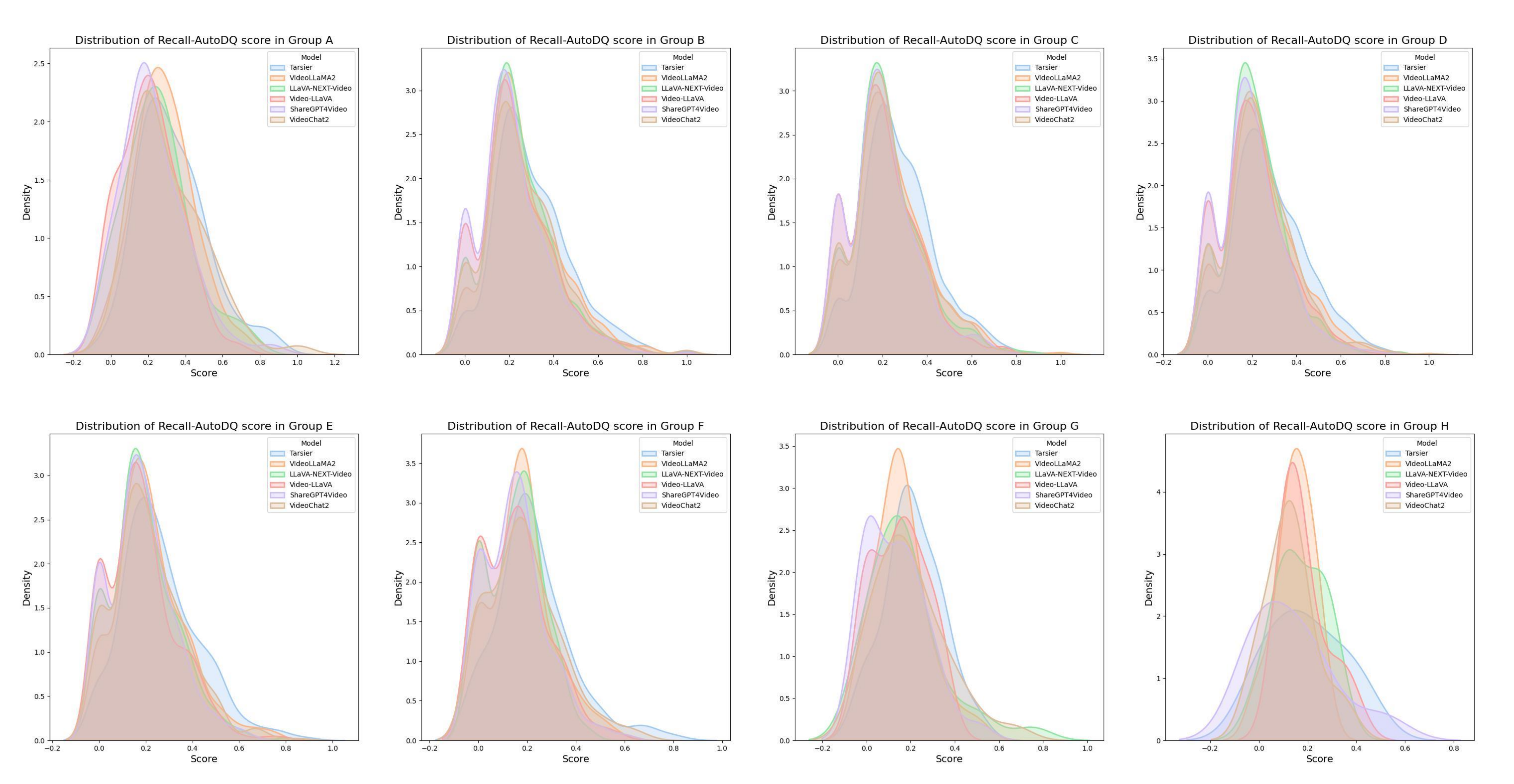}
\caption{Distribution of LVLMs scores in different groups, based on Recall (AutoDQ) metric.}
\label{fig:Recall-AutoDQ}
\end{sidewaysfigure}

\clearpage

\begin{sidewaysfigure}[ht!]
\centering
\includegraphics[width=\textwidth]{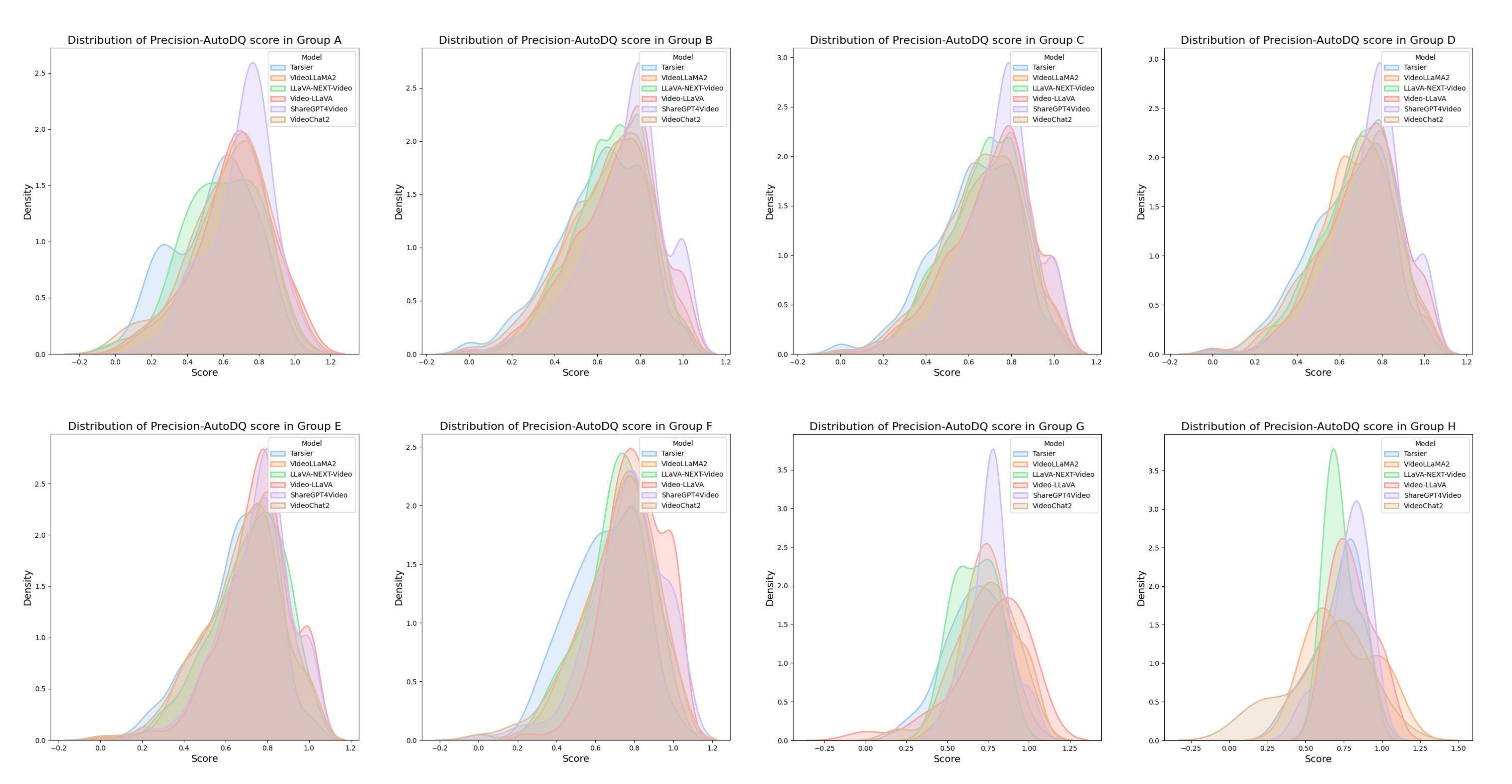}
\caption{Distribution of LVLMs scores in different groups, based on Precision (AutoDQ) metric.}
\label{fig:Precision-AutoDQ}
\end{sidewaysfigure}

\begin{sidewaysfigure}[ht!]
\centering
\includegraphics[width=\textwidth]{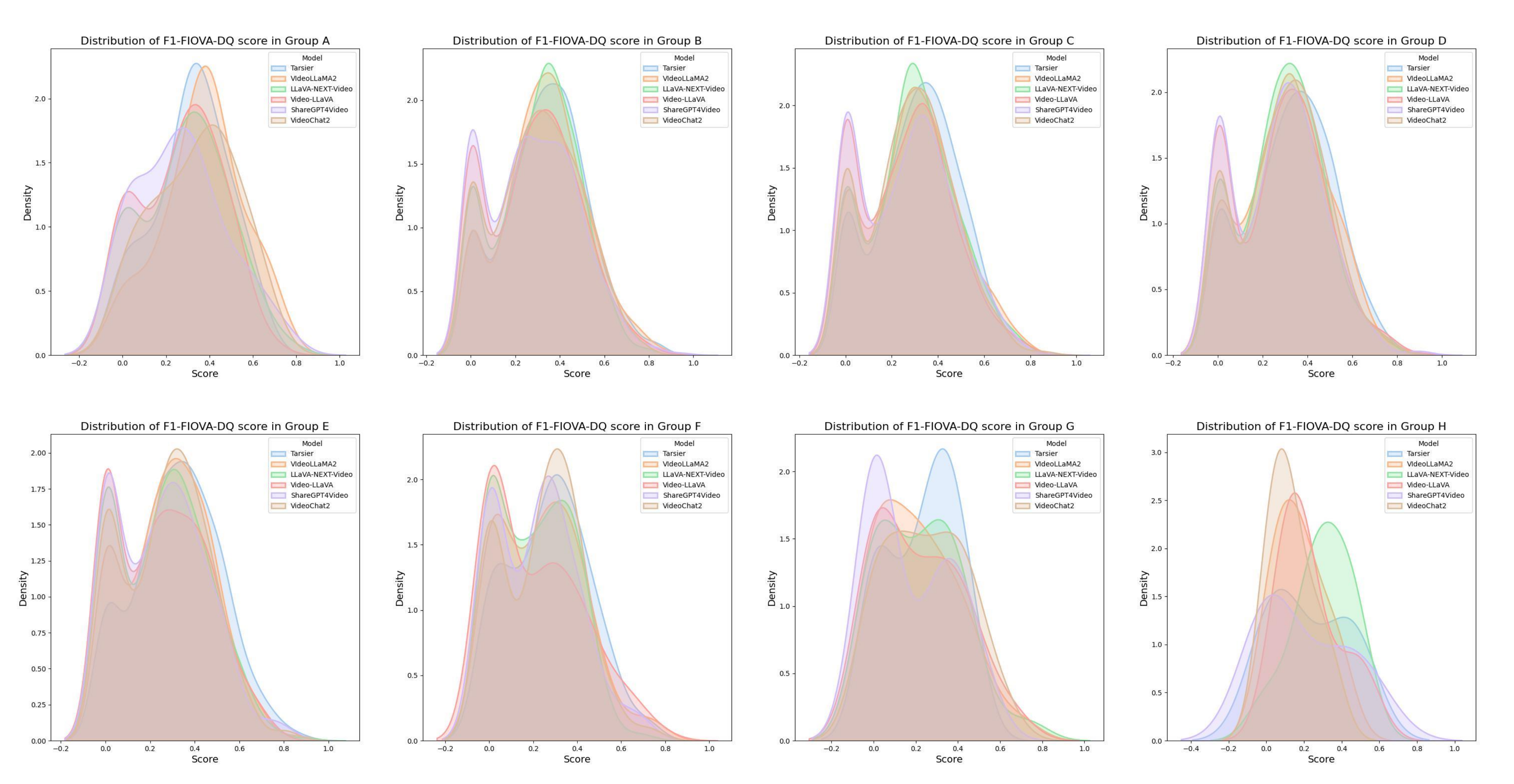}
\caption{Distribution of LVLMs scores in different groups, based on F1 (FIOVA-DQ) metric.}
\label{fig:F1-FIOVA}
\end{sidewaysfigure}

\clearpage

\begin{sidewaysfigure}[ht!]
\centering
\includegraphics[width=\textwidth]{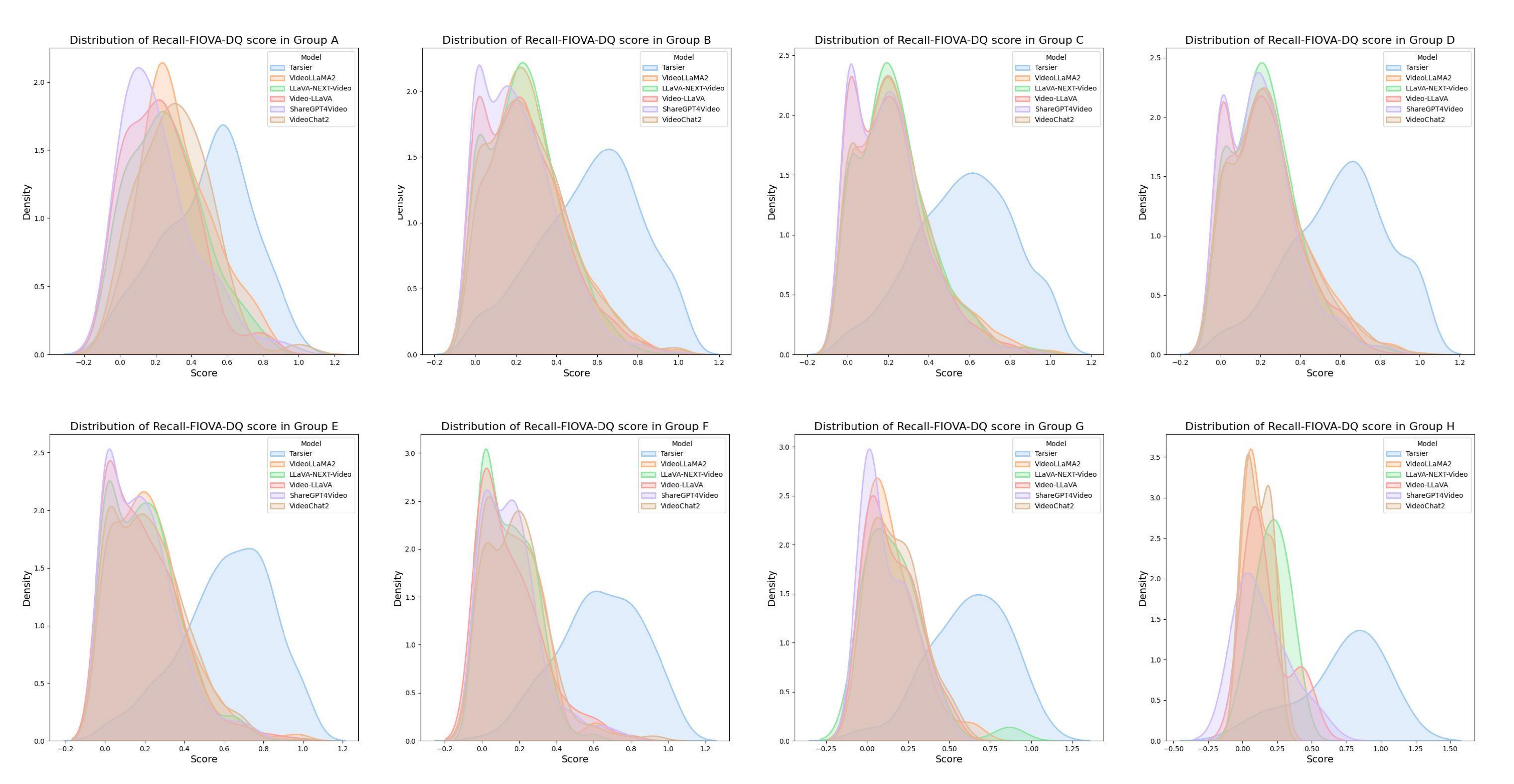}
\caption{Distribution of LVLMs scores in different groups, based on Recall (FIOVA-DQ) metric.}
\label{fig:Recall-FIOVA}
\end{sidewaysfigure}

\clearpage

\begin{sidewaysfigure}[ht!]
\centering
\includegraphics[width=\textwidth]{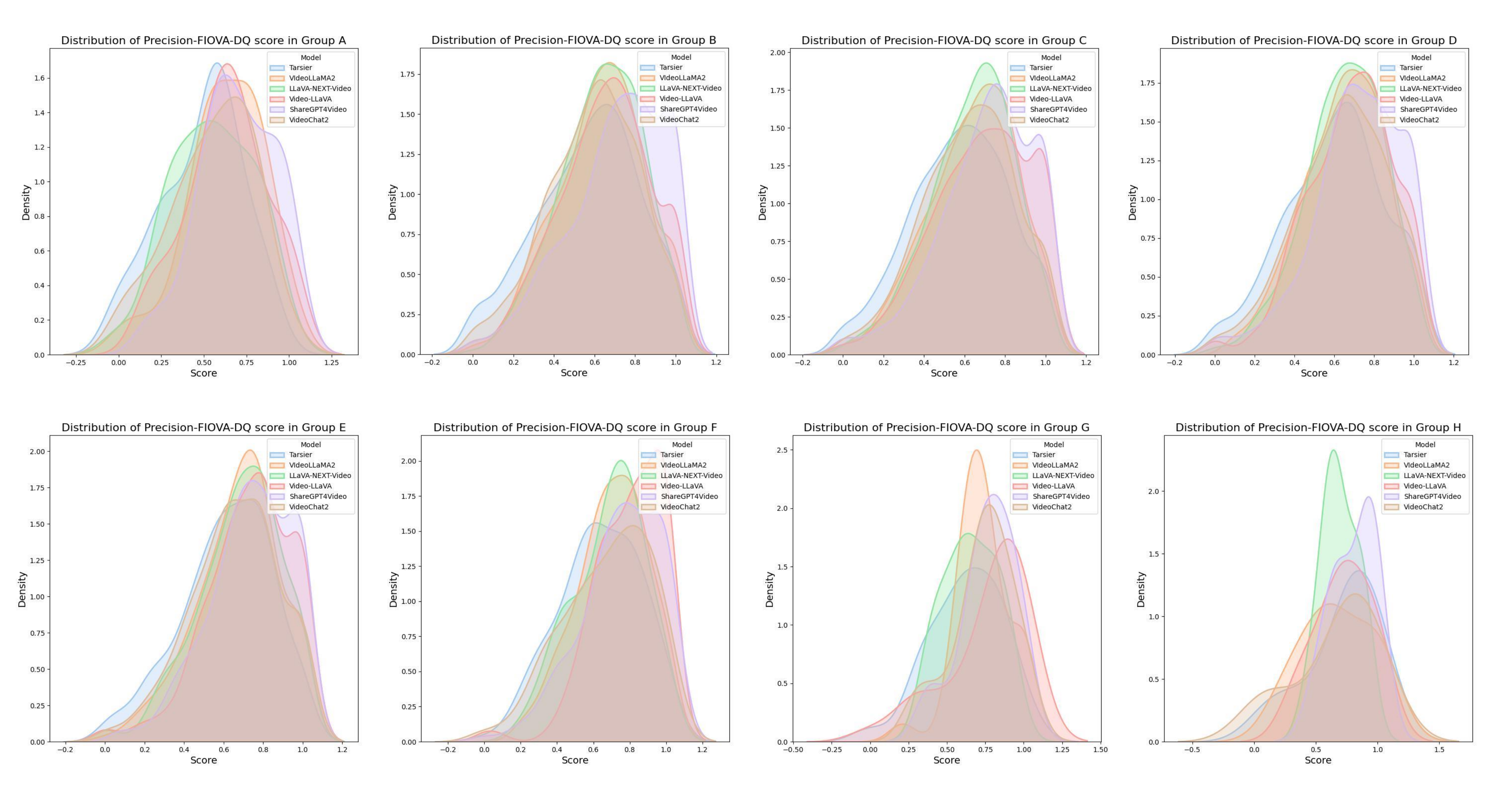}
\caption{Distribution of LVLMs scores in different groups, based on Precision (FIOVA-DQ) metric.}
\label{fig:Precision-FIOVA}
\end{sidewaysfigure}

\clearpage

\subsection{Comparison between Humans and LVLMs in Caption Length}
\label{subsec:results-length}

\begin{figure}[ht!]
\centering
\includegraphics[width=\textwidth]{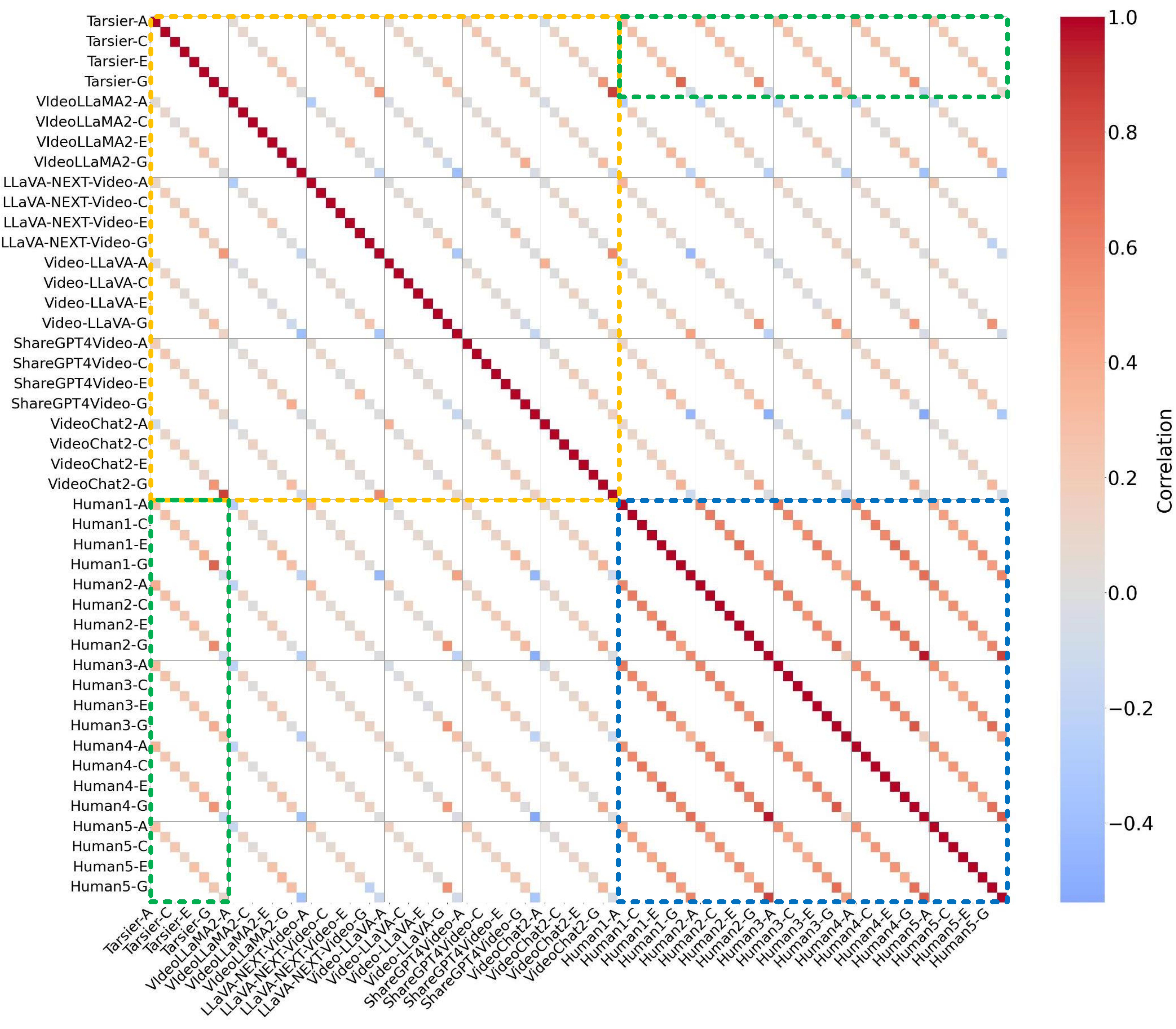}
\caption{Correlation between LVLMs and humans in video description length (based on 8 sub-groups). It can be seen that the blue dashed box represents the results of humans, and the description length is highly consistent between human annotators. The yellow dashed box shows the results of LVLMs. The description lengths between LVLMs vary greatly, especially for the descriptions of Group H, which have basically no correlation. The green dashed line is a comparison between Tarsier, the model with the best performance in multiple indicators, and humans. It can be seen that Tarsier has a higher correlation with human description length than other models.}
\label{length-corr.pdf}
\end{figure}

\clearpage

\subsection{More Specific Examples}
\label{subsec:more-examples}

\subsubsection{Error Type Categorization}

\begin{figure}[ht!]
\centering
\includegraphics[width=\textwidth]{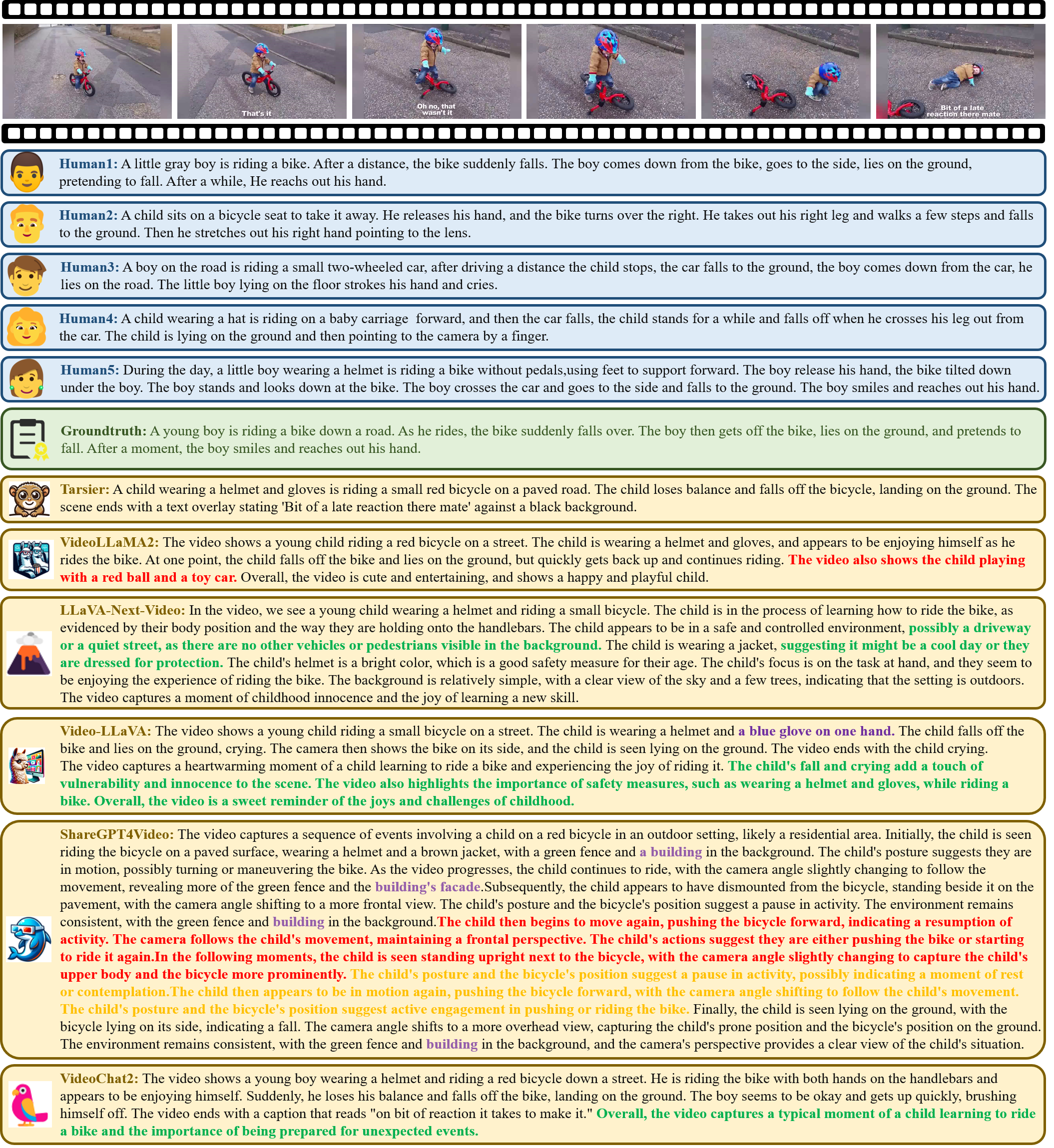}
\caption{Representative frames and descriptions in this example.}
\label{fig:example--}
\end{figure}

\begin{figure}[ht!]
\centering
\includegraphics[width=\textwidth]{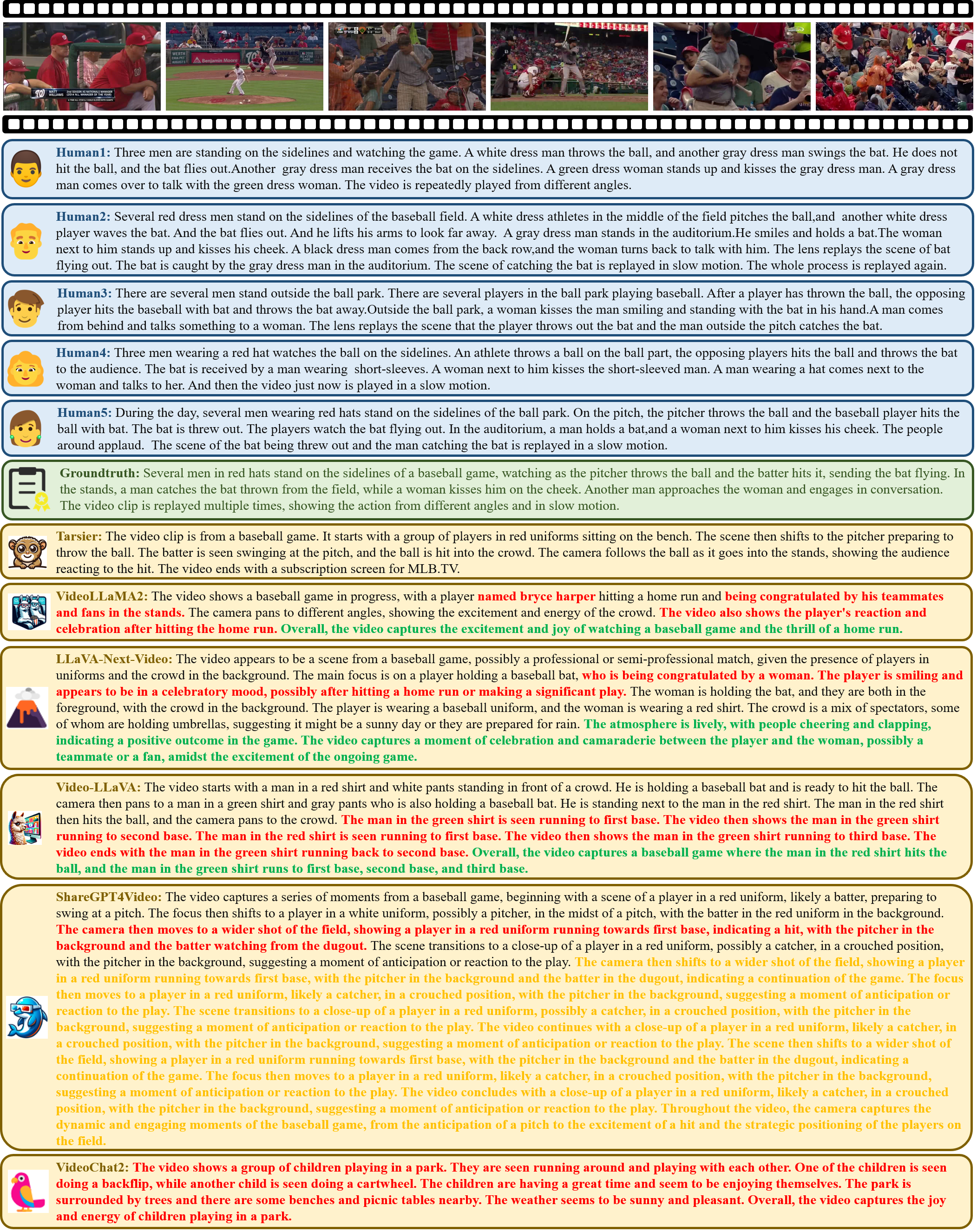}
\caption{Representative frames and descriptions in this example.
}
\label{fig:example-0}
\end{figure}

\begin{figure}[t!]
\centering
\includegraphics[width=0.83\textwidth]{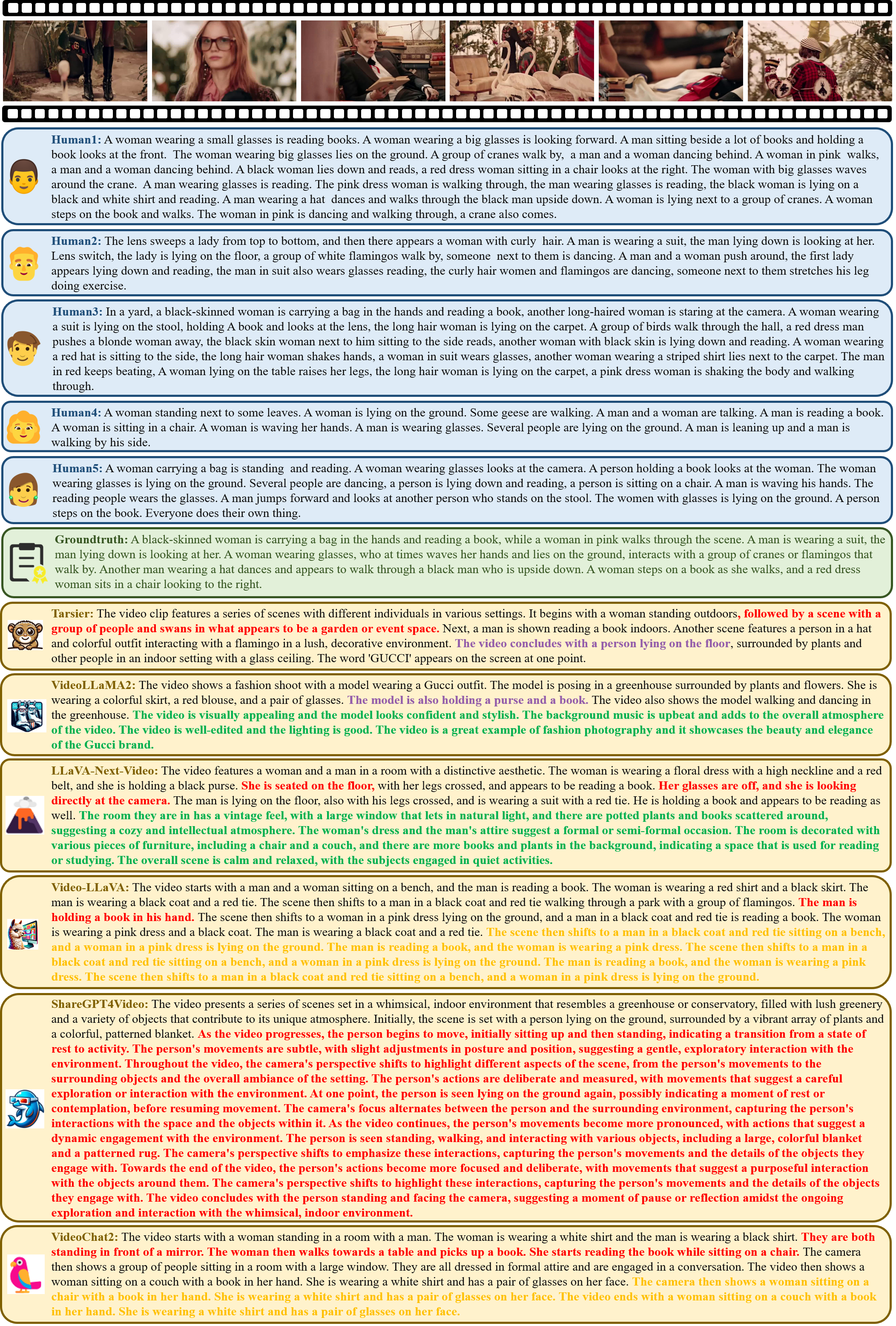}
\caption{Representative frames and descriptions in this example.
}
\label{fig:example-+}
\end{figure}

\begin{figure}[ht!]
\centering
\includegraphics[width=\textwidth]{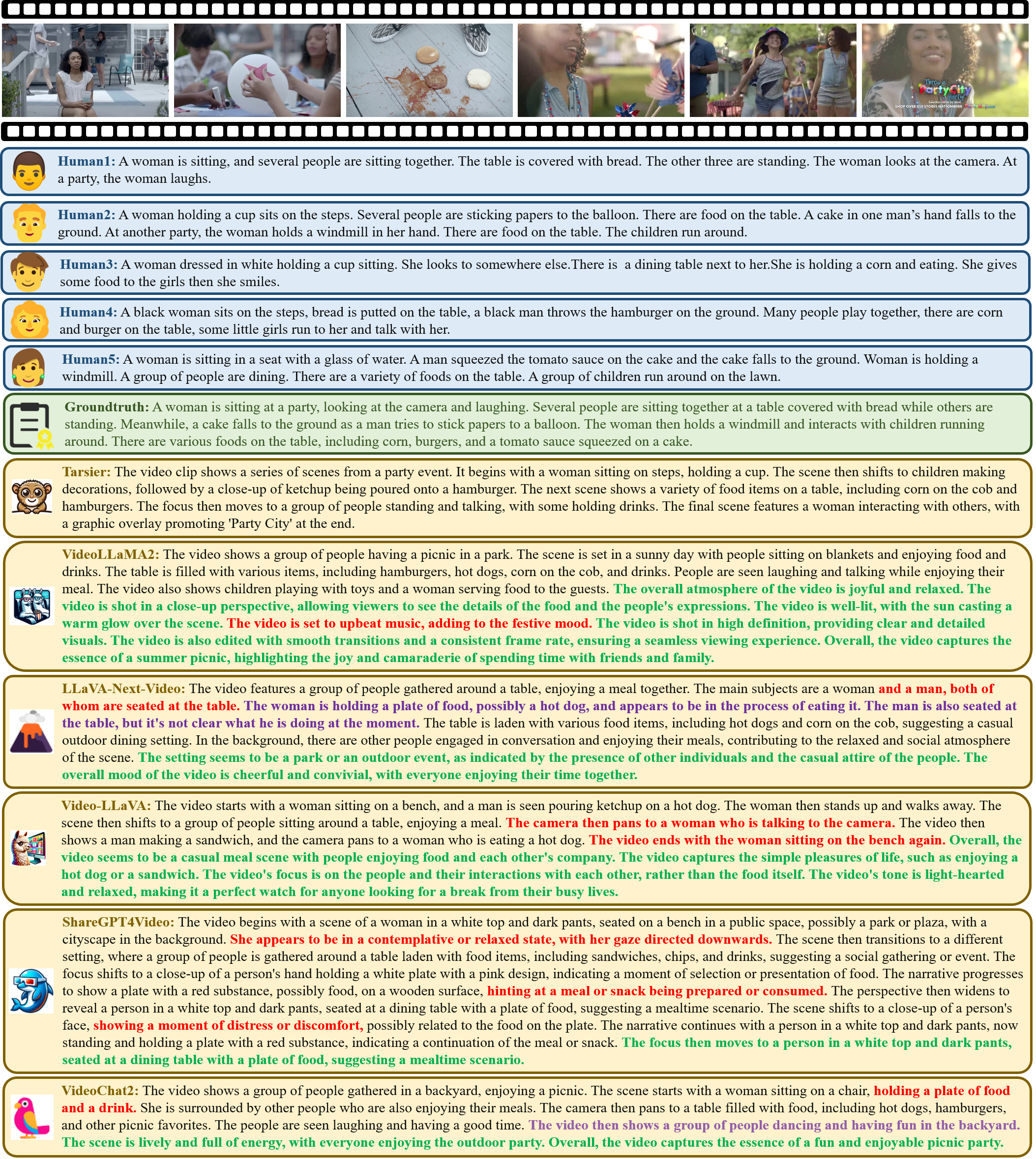}
\caption{Representative frames and descriptions in this example.}
\label{fig:example-error}
\end{figure}

\begin{figure}[ht!]
\centering
\includegraphics[width=0.85\textwidth]{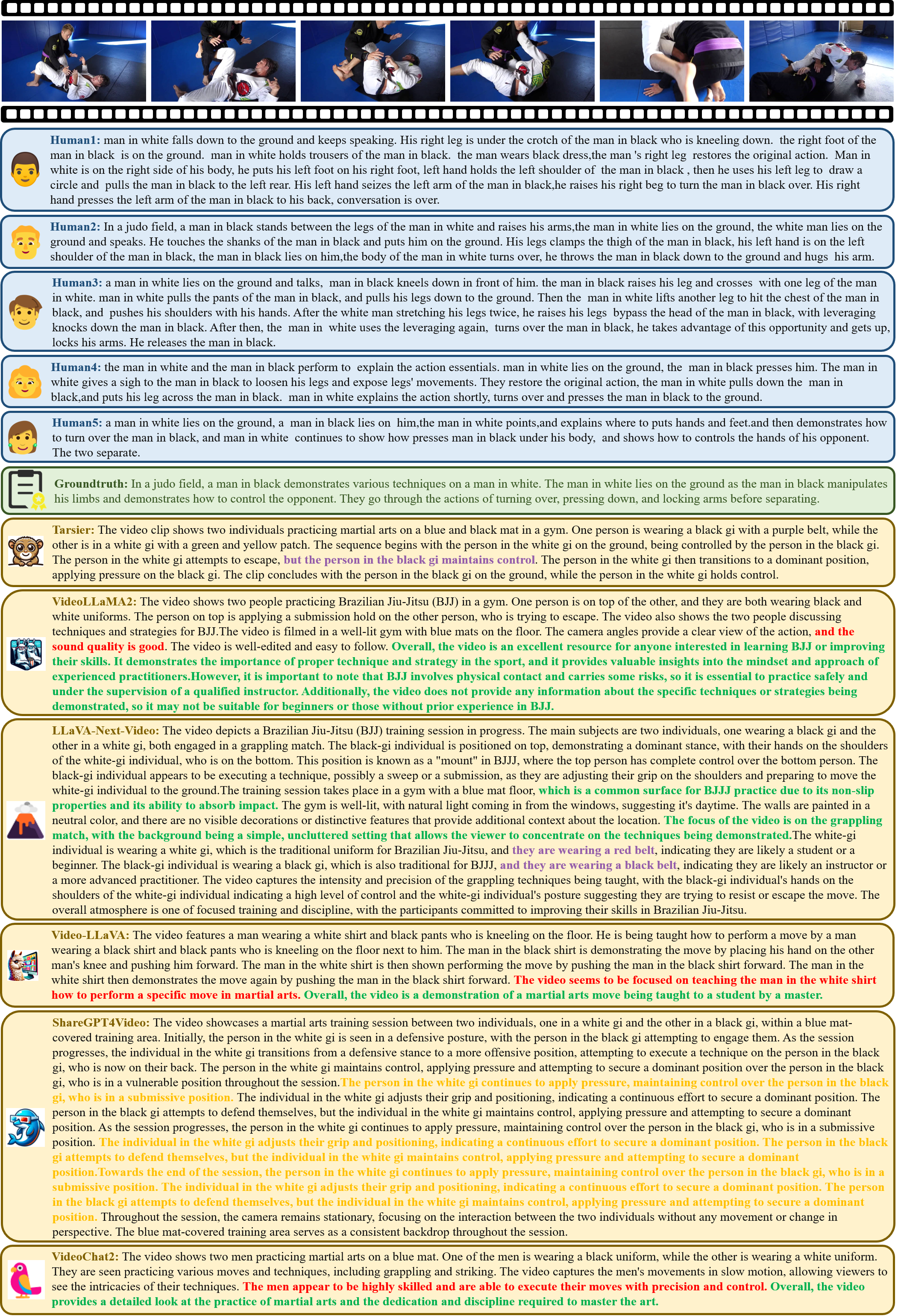}
\caption{Representative frames and descriptions in this example.
}
\label{fig:example-right}
\end{figure}

\begin{figure}[t!]
\centering
\includegraphics[width=\textwidth]{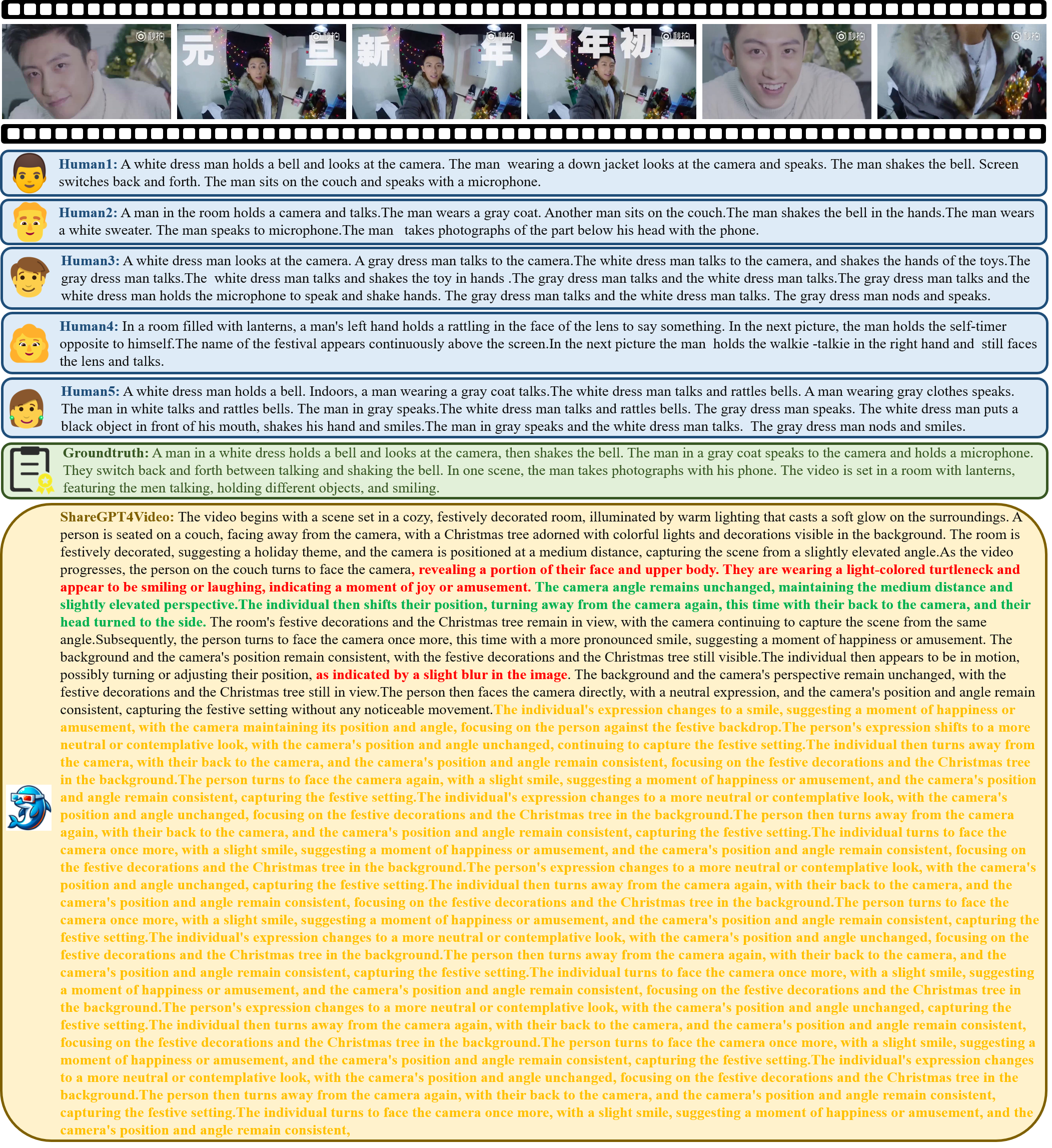}
\caption{Representative frames and descriptions in this example.}
\label{fig:example-bad}
\end{figure}

We identified five common error types through annotation of the examples (Fig.~\ref{fig:example--} to Fig.~\ref{fig:example-bad}), offering insights into the shortcomings of the model-generated descriptions:
\begin{enumerate}
    \item Omission: Critical events or objects in the video are not described by the model. Although such omissions cannot be explicitly marked in the output, we provide textual analyses to highlight these gaps for the relevant examples.
    \item Misrepresentation: The description includes content inconsistent with the video. These inaccuracies are highlighted in \textbf{purple} within the model outputs.
    \item Redundancy: Repetitions of the same event or information occur within the description. Such redundant content is marked in \textbf{yellow}.
    \item Excessive Extension: The model introduces speculative or unnecessary details beyond what is present in the video. These overextended descriptions are marked in \textbf{green}.
    \item Hallucination: The model generates content entirely absent from the video. Such fabricated details are marked in \textbf{red}.
\end{enumerate}

By categorizing these errors and visualizing them through annotations, we aim to provide a structured framework for evaluating the quality and reliability of model-generated video descriptions. This systematic error analysis highlights key areas requiring improvement, offering guidance for the development of more accurate and coherent LVLMs.

\clearpage

\label{swithd}

In Fig.~\ref{fig:example--}, human performance in video description tasks demonstrates remarkable consistency, especially in simpler scenarios where humans can effectively capture key content and provide accurate descriptions with minimal variation. In contrast, LVLMs exhibit significant limitations in these scenarios, often struggling to identify critical details and failing to match human descriptive ability. This discrepancy stems from the models' inability to fully comprehend the overall context and integrate video events with background information, which are essential for accurate and coherent descriptions.

Among the LVLMs, models like LLaVA-NEXT-Video, Video-LLaVA, and VideoChat2 frequently exhibit issues of redundancy in their outputs. ShareGPT4Video shows pronounced hallucinations and repetitive descriptions, further highlighting its challenges in maintaining precision. Tarsier, while avoiding hallucination and excessive redundancy, suffers from omissions, such as neglecting the actions occurring after the boy lies on the ground. These findings underscore the persistent gap between LVLMs and human performance, particularly in scenarios requiring detailed understanding and contextual integration.

\clearpage

In Fig.~\ref{fig:example-0}, when video content is straightforward and visually intuitive, such as a person playing baseball or a noticeable change in scenery, there is minimal difference in performance between humans and LVLMs. Both can effectively capture key visual elements and generate accurate descriptions without relying on deep contextual or cultural understanding. This similarity highlights the ability of LVLMs to process intuitive visual information efficiently.

However, in videos with frequent camera switches and more complex content, all models exhibit omissions in describing critical events. For example, ShareGPT4Video frequently includes redundant and repetitive descriptions, which detracts from its overall coherence. Furthermore, VideoChat2 demonstrates significant errors by misinterpreting the entire video as children playing, highlighting its limitations in accurately capturing contextual nuances and differentiating between events.

\clearpage

In Fig.~\ref{fig:example-+}, significant semantic variation is evident in human descriptions of this video, underscoring the interpretative challenges posed by its strong artistic elements and intricate narrative structure. This variability stems from differing human perspectives, shaped by personal experiences, cultural backgrounds, and emotional interpretations, making it difficult to achieve consistent descriptive consensus.

In contrast, LVLMs exhibit greater consistency in their descriptions but with notable shortcomings. The complexity of the video often overwhelms their ability to focus on specific scene details, leading to issues such as hallucinations, repetitive phrasing, and redundant content. These limitations are particularly evident in ShareGPT4Video, which frequently introduces irrelevant or speculative details, further reducing the coherence and accuracy of its descriptions.

The contrast between human and LVLM performance reflects their fundamental differences. Humans naturally bring diverse, subjective viewpoints to complex video descriptions, which enriches their interpretations but decreases consistency. Meanwhile, LVLMs, trained on large datasets to identify and describe widely recognizable visual elements, prioritize consistency and universality over individuality. This standardization allows LVLMs to generate predictable outputs, though at the cost of nuanced understanding in highly artistic or subjective contexts.

\clearpage

In Fig.~\ref{fig:example-error}, the models demonstrated relatively poor performance in describing this video, primarily due to the intricate nature of its content. The video depicts a woman immersed in her fantasies, where the scenes transition frequently and contain temporal discontinuities. These features introduce substantial complexity, making it challenging for LVLMs to accurately interpret and coherently describe the video. The detailed elements within the fantasies, combined with the fragmented narrative structure, further exacerbate these challenges, leading to descriptions that are often unclear and lack interpretative depth.

All LVLMs exhibited varying degrees of content omissions, failing to capture critical details of the video. Additionally, most models struggled with hallucinations and repetitive descriptions, further compromising the accuracy and coherence of their outputs. These limitations underscore the need for advancements in LVLMs to better handle videos with intricate, discontinuous narratives and rich contextual details.

\clearpage

In Fig.~\ref{fig:example-right}, the models demonstrated strong descriptive abilities for a video depicting Brazilian Jiu-Jitsu practice. The stable camera work and clear temporal relationships enabled accurate recognition and detailed description of activities and actions. LVLMs typically perform well on videos with simple, well-structured content, as such videos reduce interpretative ambiguity and allow the models to focus on key visual elements, showcasing their strengths in straightforward scenarios with minimal complexity.

Despite the overall strong performance, some issues remain. Repetitive descriptions and redundancy were observed across the models, though hallucinations were relatively rare. Interestingly, while most LVLMs successfully identified the martial arts clothing worn by the characters, Video-LLaVA recognized only the color of the clothing without distinguishing its category. These results highlight the strengths and limitations of current LVLMs in processing videos with limited complexity.

\clearpage

In Fig.~\ref{fig:example-bad}, among the six LVLMs, ShareGPT4Video demonstrates pronounced hallucination issues, characterized by the generation of extensive redundant and irrelevant content. As shown in the accompanying figure, these shortcomings significantly detract from the model's descriptive precision and coherence, highlighting its challenges in effectively filtering and prioritizing relevant visual information.

\clearpage

\subsubsection{Potential Cause Analysis}

The observed limitations in LVLM performance can be attributed to three primary factors: architectural constraints, biases in training data, and inadequacies in generation strategies. These issues collectively hinder the models' ability to produce accurate, coherent, and contextually relevant video descriptions.

\paragraph{Architectural Limitations}

Current LVLMs exhibit challenges in aligning and processing video-text data effectively:

\begin{itemize}
    \item \textbf{Cross-modal alignment issues:} LVLMs such as Tarsier and VideoLLaMA2 employ differing alignment strategies—separate visual encoders for individual frames versus shared encoders for all frames—which impact their ability to comprehensively interpret video content.
    \item \textbf{Insufficient long-sequence modeling:} Many LVLMs struggle to maintain coherence across long videos with multiple events. For instance, Video-LLaVA often prioritizes the initial scenes while neglecting later segments, resulting in incomplete descriptions.
\end{itemize}

\paragraph{Training Data Bias}

Biases and inconsistencies in training data exacerbate model limitations:

\begin{itemize}
    \item \textbf{Limited data diversity:} A lack of diverse training data restricts the models’ ability to generalize. For example, Video-LLaVA demonstrates difficulty in recognizing martial arts scenes (Fig.~\ref{fig:example-right}), likely due to insufficient representation in its training set.
    \item \textbf{Propagation of hallucinations:} Noisy or incomplete training data can lead to significant hallucinations. In Fig.~\ref{fig:example-0}, VideoChat2 misidentifies players and spectators in a baseball stadium as children, reflecting a severe misalignment between the training data and real-world scenarios.
\end{itemize}

\paragraph{Generation Strategy Issues}

Simplistic or under-constrained generation strategies further contribute to descriptive shortcomings:

\begin{itemize}
    \item \textbf{Repetition and incoherence:} Models like ShareGPT4Video exhibit repetitive descriptions, often due to reliance on basic generation techniques such as beam search.
    \item \textbf{Weak semantic constraints:} Inadequate constraints during generation processes result in hallucinated content and semantic inconsistencies, undermining descriptive accuracy and coherence.
\end{itemize}

These challenges highlight the need for advancements in architectural design, data quality, and generation techniques to improve the performance of LVLMs in complex video understanding tasks.

\clearpage

\subsubsection{Suggestions for Improvement and Optimization}

To enhance the performance and applicability of LVLMs in video description tasks, improvements can be made across three main areas: model optimization, training data enhancement, and evaluation methodology refinement.

\paragraph{Model Optimization}

Enhancing LVLM architectures is crucial for improving their descriptive accuracy and coherence:

\begin{itemize}
    \item \textbf{Detail capture and sequence modeling:} Refining attention mechanisms to focus on key events and long-term dependencies can significantly enhance the comprehensiveness of descriptions. Hierarchical attention mechanisms, such as the STC Connector in VideoLLaMA2, provide a promising direction for maintaining spatiotemporal continuity in descriptions.
    \item \textbf{Semantic alignment:} Incorporating cross-modal consistency checks, such as ensuring alignment between visual and linguistic representations, can mitigate hallucination and improve semantic coherence. The approach used by LLaVA-NeXT-Video exemplifies the importance of robust alignment strategies throughout the comprehension process.
    \item \textbf{Redundancy elimination:} Implementing mechanisms to identify and remove repetitive content during generation can enhance description fluency and reduce redundancy, ensuring more coherent outputs.
\end{itemize}

\paragraph{Training Data Optimization}

Improving the diversity and quality of training data is essential for mitigating model biases and enhancing generalization:

\begin{itemize}
    \item \textbf{Data diversity:} Expanding datasets to include a wider range of scenarios, especially complex events in long videos, can help LVLMs better adapt to diverse contexts.
    \item \textbf{Data quality:} Cleaning training corpora by removing noisy or hallucinated examples can further enhance the robustness of models. For instance, while ShareGPT4Video demonstrates improvements through high-quality data, additional refinements in its dataset could yield even better results.
\end{itemize}

\paragraph{Evaluation Methodology Refinement}

Refining evaluation methods within the FIOVA framework can provide deeper insights into model performance and limitations:

\begin{itemize}
    \item Error categorization: Incorporating fine-grained error classification, such as event-level similarity metrics in FIOVA-DQ, can more precisely identify specific weaknesses in LVLM outputs. This approach can guide targeted improvements by linking descriptive errors to their underlying causes.
\end{itemize}

By addressing these areas, the performance and applicability of LVLMs can be significantly enhanced, ensuring their capability to handle complex, multi-event video scenarios with improved accuracy and reliability.

\clearpage

\section{Limitations}
\label{sec:limitations}

While FIOVA provides a large-scale, multi-annotator benchmark for cognitively aligned video caption evaluation, several limitations remain that may affect its generalizability, fairness, and responsible deployment:

\begin{itemize}
    \item \textbf{Language Scope:} All annotations and synthesized groundtruths are in English. This constrains the benchmark’s applicability in multilingual or low-resource language settings. Future extensions could incorporate multi-language annotations to support cross-lingual and culturally diverse evaluations.

    \item \textbf{Dataset Domain Bias:} Despite covering 38 thematic categories, FIOVA may still inherit distributional bias from its video sourcing platforms. Certain communities, age groups, or underrepresented activities may be inadequately reflected, potentially skewing model performance and alignment outcomes.

    \item \textbf{Evaluation Metric Limitations:} While FIOVA-DQ incorporates cognitive salience via human weighting, it does not capture event order, causality, or discourse-level coherence. As such, it cannot fully substitute for human judgment in high-level narrative evaluation and should be complemented by qualitative assessments.

    \item \textbf{Broader Societal Impacts:} The use of human-aligned video captions in LVLMs may amplify both beneficial and adverse downstream effects. On the positive side, improved narrative understanding could support accessibility tools and education platforms. However, risks such as content hallucination, cultural misrepresentation, and misuse in surveillance or disinformation systems must be carefully considered. We encourage future research to evaluate societal risks and implement safeguards when building on FIOVA.
\end{itemize}

These limitations suggest that while FIOVA offers a robust diagnostic tool for evaluating long-video captioning, users should exercise caution when applying it to high-stakes, multilingual, or socially sensitive scenarios.


\begin{thebibliography}{10}

\bibitem{dtllm}
Xuchen Li, Xiaokun Feng, Shiyu Hu, Meiqi Wu, Dailing Zhang, Jing Zhang, and Kaiqi Huang.
\newblock Dtllm-vlt: Diverse text generation for visual language tracking based on llm.
\newblock In {\em Proceedings of the IEEE/CVF Conference on Computer Vision and Pattern Recognition}, pages 7283--7292, 2024.

\bibitem{li2024visual}
Xuchen Li, Shiyu Hu, Xiaokun Feng, Dailing Zhang, Meiqi Wu, Jing Zhang, and Kaiqi Huang.
\newblock Visual language tracking with multi-modal interaction: A robust benchmark.
\newblock {\em arXiv preprint arXiv:2409.08887}, 2024.

\bibitem{mahapatra2024impact}
Joy Mahapatra and Utpal Garain.
\newblock Impact of model size on fine-tuned llm performance in data-to-text generation: A state-of-the-art investigation.
\newblock {\em arXiv preprint arXiv:2407.14088}, 2024.

\bibitem{toolqa}
Yuchen Zhuang, Yue Yu, Kuan Wang, Haotian Sun, and Chao Zhang.
\newblock Toolqa: A dataset for llm question answering with external tools.
\newblock {\em Advances in Neural Information Processing Systems}, 36:50117--50143, 2023.

\bibitem{saito2024unsupervised}
Kuniaki Saito, Kihyuk Sohn, Chen-Yu Lee, and Yoshitaka Ushiku.
\newblock Unsupervised llm adaptation for question answering.
\newblock {\em arXiv preprint arXiv:2402.12170}, 2024.

\bibitem{gpt4}
Josh Achiam, Steven Adler, Sandhini Agarwal, Lama Ahmad, Ilge Akkaya, Florencia~Leoni Aleman, Diogo Almeida, Janko Altenschmidt, Sam Altman, Shyamal Anadkat, et~al.
\newblock Gpt-4 technical report.
\newblock {\em arXiv preprint arXiv:2303.08774}, 2023.

\bibitem{llava}
Haotian Liu, Chunyuan Li, Qingyang Wu, and Yong~Jae Lee.
\newblock Visual instruction tuning.
\newblock {\em Advances in neural information processing systems}, 36, 2024.

\bibitem{huang2024free}
Hanzhuo Huang, Yufan Feng, Cheng Shi, Lan Xu, Jingyi Yu, and Sibei Yang.
\newblock Free-bloom: Zero-shot text-to-video generator with llm director and ldm animator.
\newblock {\em Advances in Neural Information Processing Systems}, 36, 2024.

\bibitem{huang2024vtimellm}
Bin Huang, Xin Wang, Hong Chen, Zihan Song, and Wenwu Zhu.
\newblock Vtimellm: Empower llm to grasp video moments.
\newblock In {\em Proceedings of the IEEE/CVF Conference on Computer Vision and Pattern Recognition}, pages 14271--14280, 2024.

\bibitem{sotverse}
Shiyu Hu, Xin Zhao, and Kaiqi Huang.
\newblock Sotverse: A user-defined task space of single object tracking.
\newblock {\em International Journal of Computer Vision}, 132(3):872--930, 2024.

\bibitem{hu2024multi}
Shiyu Hu, Dailing Zhang, Xiaokun Feng, Xuchen Li, Xin Zhao, Kaiqi Huang, et~al.
\newblock A multi-modal global instance tracking benchmark (mgit): Better locating target in complex spatio-temporal and causal relationship.
\newblock {\em Advances in Neural Information Processing Systems}, 36, 2024.

\bibitem{hu2023global}
Shiyu Hu, Xin Zhao, Lianghua Huang, and Kaiqi Huang.
\newblock Global instance tracking: Locating target more like humans.
\newblock {\em IEEE Transactions on Pattern Analysis and Machine Intelligence (TPAMI)}, 45(1):576--592, 2023.

\bibitem{howto100m}
Antoine Miech, Dimitri Zhukov, Jean-Baptiste Alayrac, Makarand Tapaswi, Ivan Laptev, and Josef Sivic.
\newblock Howto100m: Learning a text-video embedding by watching hundred million narrated video clips.
\newblock 2019.

\bibitem{acav}
Sangho Lee, Jiwan Chung, Youngjae Yu, Gunhee Kim, Thomas Breuel, Gal Chechik, and Yale Song.
\newblock Acav100m: Automatic curation of large-scale datasets for audio-visual video representation learning.
\newblock 2021.

\bibitem{merlot}
Rowan Zellers, Ximing Lu, Jack Hessel, Youngjae Yu, Jae~Sung Park, Jize Cao, Ali Farhadi, and Yejin Choi.
\newblock Merlot: Multimodal neural script knowledge models.
\newblock 2021.

\bibitem{hdvila}
Hongwei Xue, Tiankai Hang, Yanhong Zeng, Yuchong Sun, Bei Liu, Huan Yang, Jianlong Fu, and Baining Guo.
\newblock Advancing high-resolution video-language representation with large-scale video transcriptions.
\newblock 2022.

\bibitem{panda70m}
Tsai-Shien Chen, Aliaksandr Siarohin, Willi Menapace, Ekaterina Deyneka, Hsiang-wei Chao, Byung~Eun Jeon, Yuwei Fang, Hsin-Ying Lee, Jian Ren, Ming-Hsuan Yang, et~al.
\newblock Panda-70m: Captioning 70m videos with multiple cross-modality teachers.
\newblock In {\em Proceedings of the IEEE/CVF Conference on Computer Vision and Pattern Recognition}, pages 13320--13331, 2024.

\bibitem{msvd}
David Chen and William~B Dolan.
\newblock Collecting highly parallel data for paraphrase evaluation.
\newblock 2011.

\bibitem{msrvtt}
Jun Xu, Tao Mei, Ting Yao, and Yong Rui.
\newblock Msr-vtt: A large video description dataset for bridging video and language.
\newblock 2016.

\bibitem{youcook2}
Luowei Zhou, Chenliang Xu, and Jason Corso.
\newblock Towards automatic learning of procedures from web instructional videos.
\newblock In {\em Proceedings of the AAAI Conference on Artificial Intelligence}, volume~32, 2018.

\bibitem{activitynet}
Fabian Caba~Heilbron, Victor Escorcia, Bernard Ghanem, and Juan Carlos~Niebles.
\newblock Activitynet: A large-scale video benchmark for human activity understanding.
\newblock 2015.

\bibitem{didemo}
Lisa Anne~Hendricks, Oliver Wang, Eli Shechtman, Josef Sivic, Trevor Darrell, and Bryan Russell.
\newblock Localizing moments in video with natural language.
\newblock 2017.

\bibitem{lsmdc}
Anna Rohrbach, Marcus Rohrbach, Niket Tandon, and Bernt Schiele.
\newblock A dataset for movie description.
\newblock 2015.

\bibitem{vatex}
Xin Wang, Jiawei Wu, Junkun Chen, Lei Li, Yuan-Fang Wang, and William~Yang Wang.
\newblock Vatex: A large-scale, high-quality multilingual dataset for video-and-language research.
\newblock 2019.

\bibitem{tarsier}
Jiawei Wang, Liping Yuan, and Yuchen Zhang.
\newblock Tarsier: Recipes for training and evaluating large video description models.
\newblock {\em arXiv preprint arXiv:2407.00634}, 2024.

\bibitem{chai2024auroracap}
Wenhao Chai, Enxin Song, Yilun Du, Chenlin Meng, Vashisht Madhavan, Omer Bar-Tal, Jenq-Neng Hwang, Saining Xie, and Christopher~D Manning.
\newblock Auroracap: Efficient, performant video detailed captioning and a new benchmark.
\newblock {\em arXiv preprint arXiv:2410.03051}, 2024.

\bibitem{aafaq2019video}
Nayyer Aafaq, Ajmal Mian, Wei Liu, Syed~Zulqarnain Gilani, and Mubarak Shah.
\newblock Video description: A survey of methods, datasets, and evaluation metrics.
\newblock {\em ACM Computing Surveys (CSUR)}, 52(6):1--37, 2019.

\bibitem{ramanishka2016multimodal}
Vasili Ramanishka, Abir Das, Dong~Huk Park, Subhashini Venugopalan, Lisa~Anne Hendricks, Marcus Rohrbach, and Kate Saenko.
\newblock Multimodal video description.
\newblock In {\em Proceedings of the 24th ACM international conference on Multimedia}, pages 1092--1096, 2016.

\bibitem{logothetis1996visual}
Nikos~K Logothetis and David~L Sheinberg.
\newblock Visual object recognition.
\newblock {\em Annual review of neuroscience}, 19:577--621, 1996.

\bibitem{antol2015vqa}
Stanislaw Antol, Aishwarya Agrawal, Jiasen Lu, Margaret Mitchell, Dhruv Batra, C~Lawrence Zitnick, and Devi Parikh.
\newblock Vqa: Visual question answering.
\newblock In {\em Proceedings of the IEEE international conference on computer vision}, pages 2425--2433, 2015.

\bibitem{videollama2}
Zesen Cheng, Sicong Leng, Hang Zhang, Yifei Xin, Xin Li, Guanzheng Chen, Yongxin Zhu, Wenqi Zhang, Ziyang Luo, Deli Zhao, et~al.
\newblock Videollama 2: Advancing spatial-temporal modeling and audio understanding in video-llms.
\newblock {\em arXiv preprint arXiv:2406.07476}, 2024.

\bibitem{videollava}
Bin Lin, Bin Zhu, Yang Ye, Munan Ning, Peng Jin, and Li~Yuan.
\newblock Video-llava: Learning united visual representation by alignment before projection.
\newblock {\em arXiv preprint arXiv:2311.10122}, 2023.

\bibitem{llavanextvideo}
Yuanhan Zhang, Bo~Li, haotian Liu, Yong~jae Lee, Liangke Gui, Di~Fu, Jiashi Feng, Ziwei Liu, and Chunyuan Li.
\newblock Llava-next: A strong zero-shot video understanding model, April 2024.

\bibitem{videochat2}
KunChang Li, Yinan He, Yi~Wang, Yizhuo Li, Wenhai Wang, Ping Luo, Yali Wang, Limin Wang, and Yu~Qiao.
\newblock Videochat: Chat-centric video understanding.
\newblock {\em arXiv preprint arXiv:2305.06355}, 2023.

\bibitem{sharegpt4video}
Lin Chen, Xilin Wei, Jinsong Li, Xiaoyi Dong, Pan Zhang, Yuhang Zang, Zehui Chen, Haodong Duan, Bin Lin, Zhenyu Tang, et~al.
\newblock Sharegpt4video: Improving video understanding and generation with better captions.
\newblock {\em arXiv preprint arXiv:2406.04325}, 2024.

\bibitem{InternVL}
Zhe Chen, Weiyun Wang, Yue Cao, Yangzhou Liu, Zhangwei Gao, Erfei Cui, Jinguo Zhu, Shenglong Ye, Hao Tian, Zhaoyang Liu, et~al.
\newblock Expanding performance boundaries of open-source multimodal models with model, data, and test-time scaling.
\newblock {\em arXiv preprint arXiv:2412.05271}, 2024.

\bibitem{Qwen}
Shuai Bai, Keqin Chen, Xuejing Liu, Jialin Wang, Wenbin Ge, Sibo Song, Kai Dang, Peng Wang, Shijie Wang, Jun Tang, et~al.
\newblock Qwen2. 5-vl technical report.
\newblock {\em arXiv preprint arXiv:2502.13923}, 2025.

\bibitem{maaz2023video}
Muhammad Maaz, Hanoona Rasheed, Salman Khan, and Fahad~Shahbaz Khan.
\newblock Video-chatgpt: Towards detailed video understanding via large vision and language models.
\newblock {\em arXiv preprint arXiv:2306.05424}, 2023.

\bibitem{srivastava2022beyond}
Aarohi Srivastava, Abhinav Rastogi, Abhishek Rao, Abu Awal~Md Shoeb, Abubakar Abid, Adam Fisch, Adam~R Brown, Adam Santoro, Aditya Gupta, Adri{\`a} Garriga-Alonso, et~al.
\newblock Beyond the imitation game: Quantifying and extrapolating the capabilities of language models.
\newblock {\em arXiv preprint arXiv:2206.04615}, 2022.

\bibitem{liu2020glge}
Dayiheng Liu, Yu~Yan, Yeyun Gong, Weizhen Qi, Hang Zhang, Jian Jiao, Weizhu Chen, Jie Fu, Linjun Shou, Ming Gong, et~al.
\newblock Glge: A new general language generation evaluation benchmark.
\newblock {\em arXiv preprint arXiv:2011.11928}, 2020.

\bibitem{mackie2021deep}
Iain Mackie, Jeffrey Dalton, and Andrew Yates.
\newblock How deep is your learning: the dl-hard annotated deep learning dataset.
\newblock In {\em Proceedings of the 44th International ACM SIGIR Conference on Research and Development in Information Retrieval}, pages 2335--2341, 2021.

\bibitem{huang2019got}
Lianghua Huang, Xin Zhao, and Kaiqi Huang.
\newblock Got-10k: A large high-diversity benchmark for generic object tracking in the wild.
\newblock {\em IEEE transactions on pattern analysis and machine intelligence}, 43(5):1562--1577, 2019.

\bibitem{zhao2024biodrone}
Xin Zhao, Shiyu Hu, Yipei Wang, Jing Zhang, Yiming Hu, Rongshuai Liu, Haibin Ling, Yin Li, Renshu Li, Kun Liu, and Jiadong Li.
\newblock Biodrone: A bionic drone-based single object tracking benchmark for robust vision.
\newblock {\em International Journal of Computer Vision (IJCV)}, 132:1659–1684, 2024.

\bibitem{zhao2025visual}
Xin Zhao, Shiyu Hu, and Xucheng Yin.
\newblock {\em Visual Object Tracking - An Evaluation Perspective}.
\newblock Springer, 2025.

\bibitem{tan2024koala}
Reuben Tan, Ximeng Sun, Ping Hu, Jui-hsien Wang, Hanieh Deilamsalehy, Bryan~A Plummer, Bryan Russell, and Kate Saenko.
\newblock Koala: Key frame-conditioned long video-llm.
\newblock In {\em Proceedings of the IEEE/CVF Conference on Computer Vision and Pattern Recognition}, pages 13581--13591, 2024.

\bibitem{fan2025videoagent}
Yue Fan, Xiaojian Ma, Rujie Wu, Yuntao Du, Jiaqi Li, Zhi Gao, and Qing Li.
\newblock Videoagent: A memory-augmented multimodal agent for video understanding.
\newblock In {\em European Conference on Computer Vision}, pages 75--92. Springer, 2025.

\bibitem{panagopoulou2025x}
Artemis Panagopoulou, Le~Xue, Ning Yu, Junnan Li, Dongxu Li, Shafiq Joty, Ran Xu, Silvio Savarese, Caiming Xiong, and Juan~Carlos Niebles.
\newblock X-instructblip: A framework for aligning image, 3d, audio, video to llms and its emergent cross-modal reasoning.
\newblock In {\em European Conference on Computer Vision}, pages 177--197. Springer, 2025.

\bibitem{wang2025vamos}
Shijie Wang, Qi~Zhao, Minh~Quan Do, Nakul Agarwal, Kwonjoon Lee, and Chen Sun.
\newblock Vamos: Versatile action models for video understanding.
\newblock In {\em European Conference on Computer Vision}, pages 142--160. Springer, 2025.

\bibitem{korbar2025text}
Bruno Korbar, Yongqin Xian, Alessio Tonioni, Andrew Zisserman, and Federico Tombari.
\newblock Text-conditioned resampler for long form video understanding.
\newblock In {\em European Conference on Computer Vision}, pages 271--288. Springer, 2025.

\bibitem{st-llm}
Ruyang Liu, Chen Li, Haoran Tang, Yixiao Ge, Ying Shan, and Ge~Li.
\newblock St-llm: Large language models are effective temporal learners.
\newblock In {\em European Conference on Computer Vision}, pages 1--18. Springer, 2025.

\bibitem{longvlm}
Yuetian Weng, Mingfei Han, Haoyu He, Xiaojun Chang, and Bohan Zhuang.
\newblock Longvlm: Efficient long video understanding via large language models.
\newblock In {\em European Conference on Computer Vision}, pages 453--470. Springer, 2025.

\bibitem{videoagent}
Xiaohan Wang, Yuhui Zhang, Orr Zohar, and Serena Yeung-Levy.
\newblock Videoagent: Long-form video understanding with large language model as agent.
\newblock In {\em European Conference on Computer Vision}, pages 58--76. Springer, 2025.

\bibitem{team2023gemini}
Gemini Team, Rohan Anil, Sebastian Borgeaud, Jean-Baptiste Alayrac, Jiahui Yu, Radu Soricut, Johan Schalkwyk, Andrew~M Dai, Anja Hauth, Katie Millican, et~al.
\newblock Gemini: a family of highly capable multimodal models.
\newblock {\em arXiv preprint arXiv:2312.11805}, 2023.

\bibitem{hong2024cogvlm2}
Wenyi Hong, Weihan Wang, Ming Ding, Wenmeng Yu, Qingsong Lv, Yan Wang, Yean Cheng, Shiyu Huang, Junhui Ji, Zhao Xue, et~al.
\newblock Cogvlm2: Visual language models for image and video understanding.
\newblock {\em arXiv preprint arXiv:2408.16500}, 2024.

\bibitem{wang2024lvbench}
Weihan Wang, Zehai He, Wenyi Hong, Yean Cheng, Xiaohan Zhang, Ji~Qi, Xiaotao Gu, Shiyu Huang, Bin Xu, Yuxiao Dong, et~al.
\newblock Lvbench: An extreme long video understanding benchmark.
\newblock {\em arXiv preprint arXiv:2406.08035}, 2024.

\bibitem{fu2024video}
Chaoyou Fu, Yuhan Dai, Yongdong Luo, Lei Li, Shuhuai Ren, Renrui Zhang, Zihan Wang, Chenyu Zhou, Yunhang Shen, Mengdan Zhang, et~al.
\newblock Video-mme: The first-ever comprehensive evaluation benchmark of multi-modal llms in video analysis.
\newblock {\em arXiv preprint arXiv:2405.21075}, 2024.

\bibitem{bleu}
Kishore Papineni, Salim Roukos, Todd Ward, and Wei-Jing Zhu.
\newblock Bleu: a method for automatic evaluation of machine translation.
\newblock In {\em Proceedings of the 40th annual meeting of the Association for Computational Linguistics}, pages 311--318, 2002.

\bibitem{gleu}
Andrew Mutton, Mark Dras, Stephen Wan, and Robert Dale.
\newblock Gleu: Automatic evaluation of sentence-level fluency.
\newblock In {\em Proceedings of the 45th Annual Meeting of the Association of Computational Linguistics}, pages 344--351, 2007.

\bibitem{meteor}
Satanjeev Banerjee and Alon Lavie.
\newblock Meteor: An automatic metric for mt evaluation with improved correlation with human judgments.
\newblock In {\em Proceedings of the acl workshop on intrinsic and extrinsic evaluation measures for machine translation and/or summarization}, pages 65--72, 2005.

\bibitem{cider}
Ramakrishna Vedantam, C~Lawrence~Zitnick, and Devi Parikh.
\newblock Cider: Consensus-based image description evaluation.
\newblock In {\em Proceedings of the IEEE conference on computer vision and pattern recognition}, pages 4566--4575, 2015.

\bibitem{tewel2022zero}
Yoad Tewel, Yoav Shalev, Roy Nadler, Idan Schwartz, and Lior Wolf.
\newblock Zero-shot video captioning with evolving pseudo-tokens.
\newblock {\em arXiv preprint arXiv:2207.11100}, 2022.

\bibitem{wang2019learning}
Xin Wang, Jiawei Wu, Da~Zhang, Yu~Su, and William~Yang Wang.
\newblock Learning to compose topic-aware mixture of experts for zero-shot video captioning.
\newblock In {\em Proceedings of the AAAI conference on artificial intelligence}, volume~33, pages 8965--8972, 2019.

\bibitem{zhou2024actionhub}
Jiaming Zhou, Junwei Liang, Kun-Yu Lin, Jinrui Yang, and Wei-Shi Zheng.
\newblock Actionhub: a large-scale action video description dataset for zero-shot action recognition.
\newblock {\em arXiv preprint arXiv:2401.11654}, 2024.

\bibitem{sottana2023evaluation}
Andrea Sottana, Bin Liang, Kai Zou, and Zheng Yuan.
\newblock Evaluation metrics in the era of gpt-4: reliably evaluating large language models on sequence to sequence tasks.
\newblock {\em arXiv preprint arXiv:2310.13800}, 2023.

\bibitem{maaz2024videogpt+}
Muhammad Maaz, Hanoona Rasheed, Salman Khan, and Fahad Khan.
\newblock Videogpt+: Integrating image and video encoders for enhanced video understanding.
\newblock {\em arXiv preprint arXiv:2406.09418}, 2024.

\bibitem{wang2024emu3}
Xinlong Wang, Xiaosong Zhang, Zhengxiong Luo, Quan Sun, Yufeng Cui, Jinsheng Wang, Fan Zhang, Yueze Wang, Zhen Li, Qiying Yu, et~al.
\newblock Emu3: Next-token prediction is all you need.
\newblock {\em arXiv preprint arXiv:2409.18869}, 2024.

\end{thebibliography}
\end{document}